\RequirePackage{fix-cm}
\documentclass[smallextended]{svjour3}       
\smartqed  
\usepackage{amsmath,amssymb,amsfonts}
\usepackage{algorithm} 
\usepackage{algorithmic}
\usepackage{graphicx}
\usepackage{textcomp}
\usepackage[utf8]{inputenc}
\usepackage[square,numbers]{natbib}
\journalname{Soft Computing}
\begin{document}
	
	\title{Improving Predictive Uncertainty Estimation using Dropout - Hamiltonian Monte Carlo}
	
	\titlerunning{Improving Predictive Uncertainty Estimation ...}        
	
	\author{Hern\'andez, Sergio \and
		Vergara, Diego. \and
		Valdenegro-Toro, Mat\'ias \and
		Jorquera, Felipe.
	}
	
	
	\institute{S. Hern\'andez, D. Vergara, F. Jorquera \at
		Centro de Innovaci\'on en Ingenier\'ia Aplicada.\\
		Universidad Cat\'olica del Maule. Chile \\
		\email{shernandez@ucm.cl}           
		\and
		M. Valdenegro-Toro \at
        Robotics Innovation Center\\
		German Research Center for Artificial Intelligence\\
        Bremen, Germany.
	}
	
	\date{Received: date / Accepted: date}

	\maketitle
	
	\begin{abstract}
			Estimating predictive uncertainty is crucial for many computer vision tasks, from image classification to autonomous driving systems. Hamiltonian Monte Carlo (HMC) is an sampling method for performing Bayesian inference. On the other hand, Dropout regularization has been proposed as an approximate model averaging technique that tends to improve generalization in large scale models such as deep neural networks. Although, HMC provides convergence guarantees for most standard Bayesian models, it does not handle discrete parameters arising from Dropout regularization.  In this paper, we present a robust methodology for improving predictive uncertainty in classification problems, based on Dropout and Hamiltonian Monte Carlo. Even though Dropout induces a non-smooth energy function with no such convergence guarantees, the resulting discretization of the Hamiltonian proves empirical success. The proposed method allows to effectively estimate the predictive accuracy and to provide better generalization for difficult test examples.
	\end{abstract}  
	\section{Introduction}
	
	Artificial Intelligence systems have a wide variety of applications and in some cases are part of complex systems whose operation involves making delicate and dangerous decisions\cite{leibig2017leveraging}. Uncertainty in knowledge representation and reasoning has been studied since the fall of symbolic expert systems \cite{Pearl:1988:PRI:52121}. Therefore, several research efforts focused on efficient methods for estimating model uncertainty and capturing the variability inherent to real world situations \cite{bishop2007pattern}.
	
	Inference using uncertainty estimates is crucial for many important tasks, such as image classification, detecting noisy examples (adversarial examples) and analyzing failure cases in decision-making systems \cite{prince2012computer}. These problems can benefit on predictive uncertainty for achieving good performance, requiring well calibrated probabilities. In such cases, Bayesian inference provides posterior predictive distributions that can be used to reduce over-confidence of the model outputs \cite{gelman2014bayesian}.
	
	 Bayesian neural networks consider model weights as random variables and have been proposed as a method to estimate predictive uncertainty \cite{neal2012bayesian}, however inference is computationally intractable for deep neural networks. More recently, \cite{gal2016dropout} proposed Monte Carlo Dropout as a method to obtain uncertainty estimates from deep learning architectures. Dropout has been previously proposed as a regularization technique for deep neural networks (see \cite{srivastava2014dropout}) and the relationship with Bayesian model uncertainty being justified as a variational approximation. The key idea is to train a deep learning model and perform inference using Monte Carlo and dropout in order to obtain an approximate posterior distribution.  
	 
	 Hamiltonian Monte Carlo (HMC) is a Markov Chain Monte Carlo (MCMC) method for obtaining a sequence of random samples while maintaining asymptotic consistency with respect to a target distribution \cite{neal2011mcmc}. HMC provides a mechanism for defining proposals with high acceptance rate, enabling more efficient exploration of the state space than standard random-walk proposals. In addition, another property of HMC is the feasibility to support high dimensional models arising from deep neural architectures \cite{beskos2013optimal}. 
	 
    
     Although HMC was originally proposed for training Bayesian neural networks, the method is not well suited for discrete parameters arising from dropout layers. In this work, we propose a methodology named Dropout - Hamiltonian Monte Carlo (D-HMC).  The proposed approach is tested with the MNIST digit recognition  \cite{lecun1998gradient}  and predictive uncertainty is compared to other sampling approaches such as Stochastic Gradient HMC (SGHMC) \cite{chen2014stochastic} and Stochastic Gradient Langevin Dynamics (SGLD) \cite{welling2011bayesian}. 
     
Transfer learning is a popular approach for domain adaption, when a large amount of data is used to pre-train a deep neural network and only a smaller amount of data is available for the new task \cite{Li2018}. The new model is prone to over-fitting, therefore a careful choice of the hyper-parameters used for transfer learning must be carried out in order to preserve the performance of the original task \cite{micelibarone2017}. Conversely, we develop a novel methodology to represent predictive uncertainty when using convolutional neural networks for transfer learning for age recognition \cite{eidinger2014age}.           
	\section{Related Work}	
	
	HMC relies on a numerical approximation (an integrator) of a continuous dynamical system. The method is designed in a way that the main properties of the dynamical system are preserved.  However,  the approximation error is highly sensitive to critical user-specified hyper-parameters, such as the number of steps and the step size. Hoffman \textit{et. al.}  introduced the No-U-Turn sampler (NUTS) \cite{hoffman2014no}, as a method to determine the appropriate number of  steps that the algorithm needs in order to converge to the target distribution. This method  uses a recursive algorithm to build a set of likely candidate points, stopping automatically when it starts to double back and retrace its steps. In practice, simple heuristics can also be used to establish the number of steps and the step size, based on the assumption that the posterior distribution is Gaussian with diagonal covariance \cite{hoffman2017learning}.
	
	Other highly successful implementations focus on tackling large data sets using ideas from stochastic optimization such as Stochastic Gradient Descent (SGD). The SGHMC and SGLD techniques support data batches and introduces a novel integrator using Langevin dynamics which takes into account the extra-noise induced in the gradient \cite{chen2014stochastic,welling2011bayesian}. Another technique, the Metropolis Adjusted Langevin Algorithm (MALA) \cite{roberts1998optimal,roberts2002langevin}, uses a Metropolis-Hastings (Metropolis) correction scheme to accept or reject proposals. Also, the Riemann Manifold Hamiltonian Monte Carlo method provides better adaption to the problem geometry for strongly correlated densities (RMHMC) \cite{girolami2011riemann, wang2013adaptive}. Finally, proximal algorithms and convex optimization techniques have also been studied in the context of MCMC \cite{pereyra2016proximal}, and HMC with log-concave or non-differentiable (non-smooth) energy functions \cite{chaari2016hamiltonian}.
%
%
%
%
	 
	\section{Methods}
	
	Different architectures have been proposed for image classification using deep neural networks. However, most models have a fully connected layer that compares the network output $\mathbf x$ with the sample training label $y$. Conversely, an image dataset $\mathbf{D}=\{\mathbf{d_1},\ldots,\mathbf{d_N}\}$ can be represented by a set of tuples $d=(\mathbf x,y)$ that contains the image features $\mathbf x \in \mathbb R^D$ and the labels $y=\{1,\ldots,K\}$. The fully connected layer consists of an activation function such as the softmax, which encodes the image features into a vector of class probabilities $[\phi_1(\theta),\ldots,\phi_k(\theta)]$, whose elements can be written as:
	
	\begin{align}
	\phi_i(\theta)=\frac{\operatorname{exp}(\mathbf x^T \mathbf{w_i} + b_i)}{\sum\limits_{k=1}^K \operatorname{exp}(\mathbf x^T \mathbf{w_k} + b_k)}
	\label{eq:softmax}
	\end{align}
	
	where $\theta=\{(\mathbf w_1,\mathbf b_1),\ldots,(\mathbf w_K,\mathbf b_K)\}$,  $\mathbf w_i \in \mathbb R^D$ is a vector, $b_i$ is an scalar, $D$ represents the dimensionality of the feature space $\mathbf x$ and $K$ the number of classes. 
	
	\subsection{Hamiltonian Monte Carlo}
	 Now, we consider the unknown parameter $\theta$ as a random variable. In a Bayesian setting, we want to sample from the posterior distribution $p(\theta \vert \mathbf{D})$ given by:
	 
	 \begin{align}
	 	p(\theta \vert \mathbf{D}) = \frac{p(\mathbf{D} \vert \theta)  p(\theta)}{p(\mathbf{D})} &= \frac{p(\mathbf{D} \vert \theta)  p(\theta)}{\int p(\mathbf{D} \vert \theta) p(\theta) \, d \theta}\\
	 	& \propto p(\mathbf{D} \vert \theta)  p(\theta)
	 \end{align}
	 
	The target distribution $p(\theta \vert \mathbf{D})$ is known up to a normalization constant. Standard Markov Chain Monte Carlo methods such as the Metropolis-Hastings algorithm can be used to obtain the required posterior probabilities \cite{gelman2014bayesian}. However due to the complexity of the target distribution (e.g. high-dimensional and possibly correlated image features),the algorithm would perform an inefficient exploration of the posterior. Instead, HMC improves the exploration by simulating Hamiltonian dynamics. A Hamiltonian function $H$ is composed as a potential energy $U$ and a kinetic energy $K$. These terms are constructed as follows:
		
	\begin{align}
	H(\theta,r) = U(\theta) + K(r)
	\end{align}
   
	where $r$ is called the momentum and is considered as an auxiliary variable. A positive-definite mass matrix $M$ is also introduced as follow: 

	\begin{align}
	U(\theta) &= - \sum_{d \in \mathbf{D}}^{}  \log p(d \vert \theta) - \log p(\theta) \\
	K(r) & = \frac{1}{2}r^{T}M^{-1}r
	\end{align}
	
	 For image classification problems, the density $p(d \vert \theta) = \prod_{i=1}^K \phi_i(\theta)^{[y=k]}$ takes the form of a categorical distribution over the $K$ different classes and the prior $p(\theta) \sim \mathcal N(0,\alpha \mathbf{I}_D)$ takes the form of a multivariate Gaussian distribution. Now, in order to sample from $p(\theta \vert \mathbf{D})$, the method simulates Hamiltonian dynamics while leaving the joint distribution $(\theta, r)$ invariant, such that:
	
	\begin{align}
	p(\theta \vert \mathbf{D}) &\propto \exp (-U(\theta))\\
	\pi(\theta,r) &\propto \exp( -H(\theta, r))
	\end{align}
	
	 The first step proposes a new value for the momentum variable from a Gaussian distribution. Then, a Metropolis update using Hamiltonian dynamics is used to propose a new state. The state evolves in a fictitious continuous time $t$, and the partial derivatives of the Hamiltonian can be seen in Equation \ref{eq:din0}.
	
	\begin{align}
	d\theta &= M^{-1}r \,dt \\
	dr & = -\nabla U(\theta)\,dt
	\label{eq:din0}
	\end{align}
	
	In order to simulate continuous dynamics, the leapfrog integrator can be used to discretize time. Using a small step size $\epsilon$, Equation \ref{eq:din0} becomes:  
	 
	\begin{align}
	r_{i}^{(t + \frac{\epsilon}{2})} &= r_{i}^{(t)} - \frac{\epsilon}{2} \nabla U(\theta^{(t)})\\
	\theta_{i}^{(t + \epsilon)} &= \theta_{i}^{(t)} + \epsilon r^{(t + \frac{\epsilon}{2})}M_{i}^{-1}\\
	r_{i}^{(t + \epsilon)} & = r_{i}^{(t + \frac{\epsilon}{2})} - \frac{\epsilon}{2} \nabla U( \theta^{(t + \epsilon)})
	\end{align}
	
	 After $i=1,\ldots,m$ iterations of the leapfrog integrator with finite $\epsilon$, the joint proposal becomes $(\theta^{(t+1)},r^{(t+1)}) = (\theta_m,r_m)$. Subsequently, a Metropolis update is used to accept the proposal with probability $\rho>  \operatorname{unif}(0, 1)$, such that:
	 
	\begin{equation}
	\centering
	\begin{split}
	\rho = &\min\{1,\exp(-H(\theta^{(t+1)},r^{(t+1)})) + H(\theta^{(t)},r^{(t)})\}
	\end{split}
	\end{equation}

%

		
	\subsubsection{Properties of Hamiltonian dynamics}
		
	An integrator is an algorithm that numerically approximates an evolution of the exact solution of a differential equation. Some important properties of Hamiltonian dynamics are important for building MCMC updates \cite{neal2011mcmc}: 
	
	\begin{itemize}
		\item Reversibility: The dynamics are time-reversible.
		\item Volume Preservation: Hamiltonian dynamics are volume preserving.
		\item Conservation of the Hamiltonian: $H$ is constant as $\theta$ and $r$ vary.
	\end{itemize}
	
	The leapfrog method in HMC satisfies the criteria of volume conservation and reversibility over time. However, the total energy is only conserved approximately, in this way a bias is introduced in the joint density $\pi(\theta,r)$. Conversely, the Metropolis update is used to satisfy the detailed balance condition.
	
		\subsubsection{Limitations of HMC}
	
	One of the main limitations of HMC is the lack of support for discrete parameters. The difficulty in extending HMC to a discrete parameter space stems from the fact that the construction of proposals relies on the numerical solution of a differential equation. In other hand, approximating the likelihood of a discrete parameter by a continuous density is not always possible  \cite{nishimura2017discontinuous}. Moreover, when any discontinuity is introduced into the energy function ($U(\theta)$), the first-order discretization does not ensure that a Metropolis correction maintains the stationary distribution invariant. Therefore, the standard implementation of HMC does not guarantee convergence when the parameter of interest has a discontinuous density. This is because integrators are designed for differential equations with smooth derivatives over time.

	\subsection{Dropout / DropConnect}	
	
	Dropout and DropConnect arise in the context regularization of deep neural networks and provide a way to combine exponentially many different architectures \cite{srivastava2014dropout, wan2013regularization}. Dropout/DropConnect can be described as follows:
	
	\begin{itemize}
		\item Regular (binary) DropConnect adds noise to global network weights, by setting a randomly selected subset of weights to zero computed by element-wise multiplication of a binary mask $\Gamma$:
		\begin{align}
		\phi_{c,i}(\theta,\Gamma)=\frac{\operatorname{exp}(\mathbf x^T (\Gamma \odot \mathbf{w_i}) + b_i)}{\sum\limits_{k=1}^K \operatorname{exp}(\mathbf x^T (\Gamma \odot \mathbf{w_k}) + b_k)}
		\end{align}
		\item Instead, binary Dropout randomly selects local inputs:
		\begin{align}
		\phi_{d,i}(\theta,\Gamma)=\frac{\operatorname{exp}( (\Gamma \odot \mathbf x^T) \mathbf{w_i} + b_i)}{\sum\limits_{k=1}^K \operatorname{exp}( (\Gamma \odot \mathbf x^T) \mathbf{w_k} + b_k)}
		\end{align}
	\end{itemize}

	where $\Gamma \sim  \mathcal B (1-p)$ is a Bernoulli distributed random mask and $p$ is the probability that the weight/input being dropped. 

	Due to its great effectiveness in various types of neural networks, the Dropout learning scheme has been the subject of research in recent years. Thus, it has been shown that Dropout is similar to bagging and other related ensemble methods \cite{warde2013empirical}. Since all models are averaged in a efficient and fast approximation with weights scaling, Dropout can be seen as an approximation to the geometric mean in the space of all possible models, this approximation is less expensive than the arithmetic mean in bagging methods \cite{baldi2013understanding}.

	\section{Dropout - Hamiltonian Monte Carlo}
	Stochastic gradient descent optimization has been extensively used for training neural networks. Instead of using the gradient of the full likelihood, stochastic gradients are used to update model parameters. In the context of Bayesian models, SGLD combines noisy gradient updates with Langevin dynamics in order to generate proposals from data subsets $B_i \subset \mathbf{D}$ such that:
	\begin{align}
	\tilde{U}(\theta) = - \frac{N}{n}\sum_{d \in B_i}^{}  \log p(d \vert \theta,\Gamma) - \log p(\theta)
	\end{align} 
	where $i=1,\ldots,N$, $N$ is the number of data subsets and $n=\vert B_i \vert$ the size of the mini-batch and $p$ the dropout rate.
	 
	Chen \textit{et.al.} introduce a new momentum variable $\nu$ as an alternative discretization for HMC \cite{chen2014stochastic}. Similar to SGLD, the choice of the step size must balance between efficient sampling and high acceptance rates (speed and accuracy). Equation \ref{eq:din2} shows the SGHMC update.
	\begin{subequations}
	\label{eq:din2}
	\begin{align}
	\theta^{(t+1)} &=\theta^{(t)} + \nu^{(t)}\\
	\nu^{(t+1)} &= (1 - \alpha) \nu^{(t)}  + \epsilon \nabla \tilde{U}(\theta^{(t+1)}) + \xi^{(t)}\\
	\xi^{(t)} &= \mathcal{N}(0, 2[\alpha-\beta]\epsilon)
	\end{align}
	\end{subequations}
	
where $\mathcal{N}$ denotes a multivariate normal density, $v$ denotes the momentum variable in SGHMC, $\epsilon$ denotes the step size, while $\alpha$ and $\beta$ are tuning constants. 
	
Integrating Dropout into SGHMC can be seen as adding a regularization term. Since HMC cannot perform inference on the discrete parameters, the gradient can be computed by either marginalization or by means of a local reparameterization \cite{Kingma2015}. Therefore, on each iteration a multivariate Bernoulli mask $\Gamma \sim \mathcal{B}(1-p)$ with probability $1-p$ is applied to the $\theta$ parameter and then to the gradient of the energy function $\tilde{U}(\theta)$. Conversely, the density $p(d \vert \theta,\Gamma)= \prod_{i=1}^K \phi_{d,i}(\theta,\Gamma)^{[y=k]}$ takes the form of a modified categorical distribution. Each one of the input components is randomly deleted  and the remaining elements are scaled up by $1/p$. The noisy gradient updates generate proposals from a perturbed target distribution \cite{bardenet2014towards}. The proposed method is described in algorithm \ref{alg:dsghmc}.

	\begin{algorithm}
		\caption{D-SGHMC}
		\label{alg:dsghmc}
		\begin{algorithmic} 
			\REQUIRE $\theta_{\ast},M,p,\alpha,\beta,\epsilon$.
			\STATE $v_{\ast} \sim \mathcal{N}(0,M)$
			\STATE  $(\theta_{0}, v_{0}) = (\theta_{\ast}, v_{\ast})$
			\FOR {$t = 1, 2, \ldots$} 
			\STATE \textit{Sample dropout mask $\Gamma$}
			\STATE $\Gamma \sim \mathcal{B}(1-p) $
			\STATE \textit{Transform mini-batch $B_t$ using binary mask}
			\STATE $B_t^\prime \gets 1/p \,(\Gamma \odot B_t)$
			\STATE  $(\theta^{0}, v^{0}) = (\theta_{t-1}, v_{t-1})$
			\FOR {$i = 1, 2, \ldots,L$} 
			\STATE $\theta^{i} \gets \theta^{i-1} + v^{i-1}$
			\STATE $v^{i} \gets (1 - \alpha) v^{i-1}  +\epsilon \nabla \tilde{U}( \theta^{i}) + \mathcal{N}(0,2[\alpha-\beta])\epsilon)$
			\ENDFOR
			\STATE  $(\theta_{t}, v_{t}) = (\theta^{L}, v^{L})$
			\ENDFOR
			\RETURN \textit{Fully constructed Markov Chain}
		\end{algorithmic}
	\end{algorithm}

	SGHMC introduces a second-order term that reduces the discrepancy between the  discretization noise and  the stationary distribution, so it is no longer necessary to make a Metropolis correction. However, when dropout is incorporated, the energy function is no longer differentiable. In this context, the energy function of HMC requires a specially tailored discretization. Nishimura \cite{nishimura2017discontinuous} finds a solution that preserves the critical properties of the Hamiltonian dynamics, through soft approximations, where the dynamics can be analytically integrated near the discontinuity in a way that preserves the total energy. \cite{pakman2013auxiliary} have also shown that it is possible to build a discretization where the integrator maintains the irreversibility of the Markov chain and preserves energy, but volume preservation is no longer guaranteed \cite{afshar2015reflection}. Although dropout involves a discontinuous gradient, it is still possible to evaluate the gradient using automatic differentiation \cite{baldi2014dropout,Griewank2008},  where the discontinuous parameters are taken to a continuous space \cite{carpenter2017stan}.
	
	\section{Experiments}
	
	The SGHMC, SGLD and D-SGHMC algorithms were implemented in Edward \cite{tran2017deep},  a Python library for posterior probabilistic modeling, inference, and criticism. We perform inference in two highly cited computer vision problems. These data sets were not selected with the goal of improving state-of-the-art results, but to evaluate whether we can incorporate uncertainty estimates on difficult examples.
	
	In order to compare predictive uncertainty estimates obtained with the state-of-the-art sampling methods with the proposed method, the settings for the hyper-parameters are shown in Table \ref{tab:hyper-setting}. 
	
	\begin{table}[h!]
		\centering
		\begin{tabular}{|l|c|c|c|c|c|}\hline
			Method& SGHMC & D-SGHMC & SGLD \\\hline
			Step Size ($\epsilon$)	&	\multicolumn{3}{c|}{$0.0001$}\\ \hline
			Friction Constant ($\alpha$)	& \multicolumn{2}{c|}{$1.0$}& --\\ \hline
			Mini-Batch Size &	\multicolumn{3}{c|}{$100$}\\ \hline
			Epochs	& \multicolumn{3}{c|}{$100$}\\ \hline
			Warmup	& \multicolumn{3}{c|}{$500$}\\ \hline
			Iterations	& \multicolumn{3}{c|}{Epochs $\times$ Num. Batches $+$ Warmup  }\\ \hline
			Prediction Samples	&\multicolumn{3}{c|}{$30$}\\ \hline
			Dropout probability ($p$)&--	&\multicolumn{1}{|c|}{$0.1$ to $0.9$}& --\\ \hline
		\end{tabular}	
		\caption{Hyper-Parameters Setting}
		\label{tab:hyper-setting}
	\end{table}
	
	Training data is split in mini-batches of size $n=100$ and whitening is performed on each one of these batches.  On the other hand, predictive distribution is estimated using $30$ Monte Carlo samples over the Test dataset
	
	\subsection{Digit Recognition on MNIST}
	
\begin{figure}
	\centering
	\includegraphics[ width=1\linewidth]{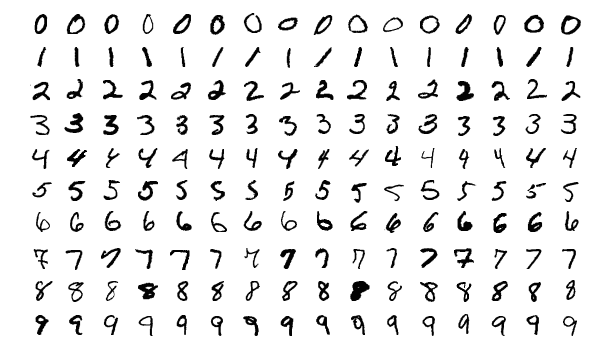}
	\caption{Examples from the MNIST database.}
	\label{fig:mnist}
	\end{figure}

In the first experiment, the predictive uncertainty on the the MNIST dataset is evaluated \cite{lecun1998gradient}. The database contains $60.000$ training images and $10.000$ test images, normalized to $28\times28$ pixels ($784$ features) and stored in gray scale. Figure \ref{fig:mnist} shows some examples of the database.

A total number  of $5$ independent chains are used for each method on MNIST. The Dropout rate is varied as $p \in [0.1, 0.5, 0.9]$, and the predictive results are compared with  SGHMC and SGLD using the hyper-parameter setting established in the Table \ref{tab:hyper-setting}. The results can be seen in Table \ref{tab:mnist}.

	\begin{table}
		\centering
		\begin{tabular}{|l|c|}\hline
			Method& Total Accuracy ($\%$)\\\hline
			SGHMC	&	$90.94 \pm 0.28 $\\
			SGLD&	$88.06 \pm 0.18$\\
			D-SGHMC ($p=0.9$)	& $91.66 \pm 0.10$\\
			D-SGHMC ($p=0.5$)    & $91.72 \pm 0.11$ \\
			D-SGHMC ($p=0.1$)    & $88.26 \pm 0.08$\\\hline
		\end{tabular}	
		\caption{Test Accuracy for MNIST data Set. $5$ independent chains.}
		\label{tab:mnist}
	\end{table}

Compared to other sampling techniques, the results show that D-SGHMC obtains a lower error when the dropout probability is set to $0.5$, which is consistent to the state-of-the-art results in terms of linear classification methods \cite{lecun1998gradient}.

\begin{figure} 
	\centering
	\includegraphics[width=1\linewidth]{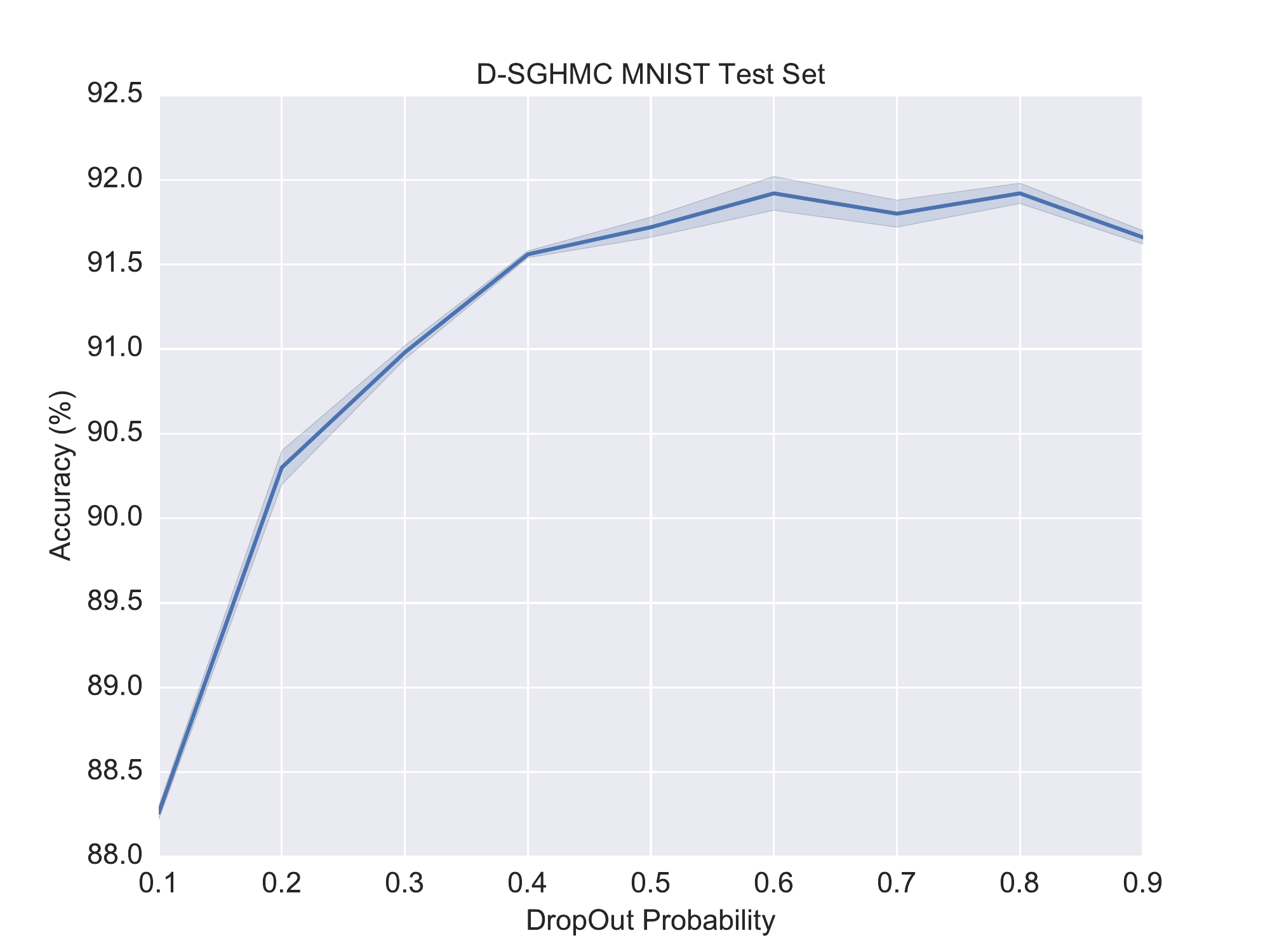}
	\caption{D-SGHMC Sensitivity analysis for $p$ hyper-parameter. $5$ independent chains on MNIST.}
	\label{fig:sensitivity}
\end{figure}

The role of the Dropout probability ($p$) for D-SGHMC is also studied. Figure \ref{fig:sensitivity}  show accuracy results with the execution of $5$ independent chains. It is possible to establish that for the studied data set a dropout probability between on $0.6$ and $0.8$ generates lower error rates. On the other hand, it can also be seen that probabilities lower than $0.4$ rapidly decreases the variable's influence on the classification results.

\begin{figure}
	\begin{minipage}[t]{0.46\linewidth}
		\centering
		\includegraphics[width=\linewidth]{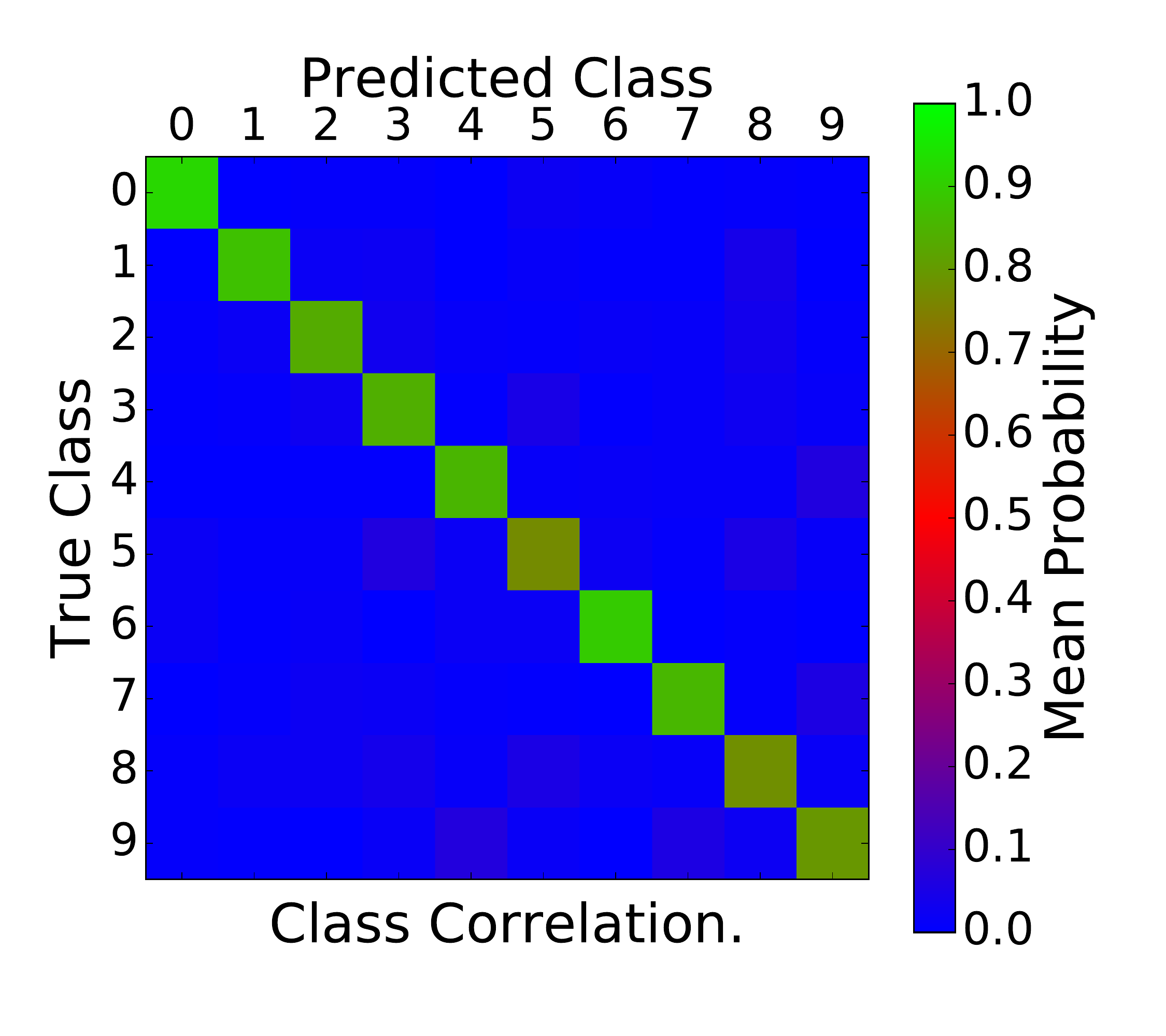}
		(a) SGHMC
	\end{minipage}
	\begin{minipage}[t]{0.46\linewidth}
		\centering
		\includegraphics[width=\linewidth]{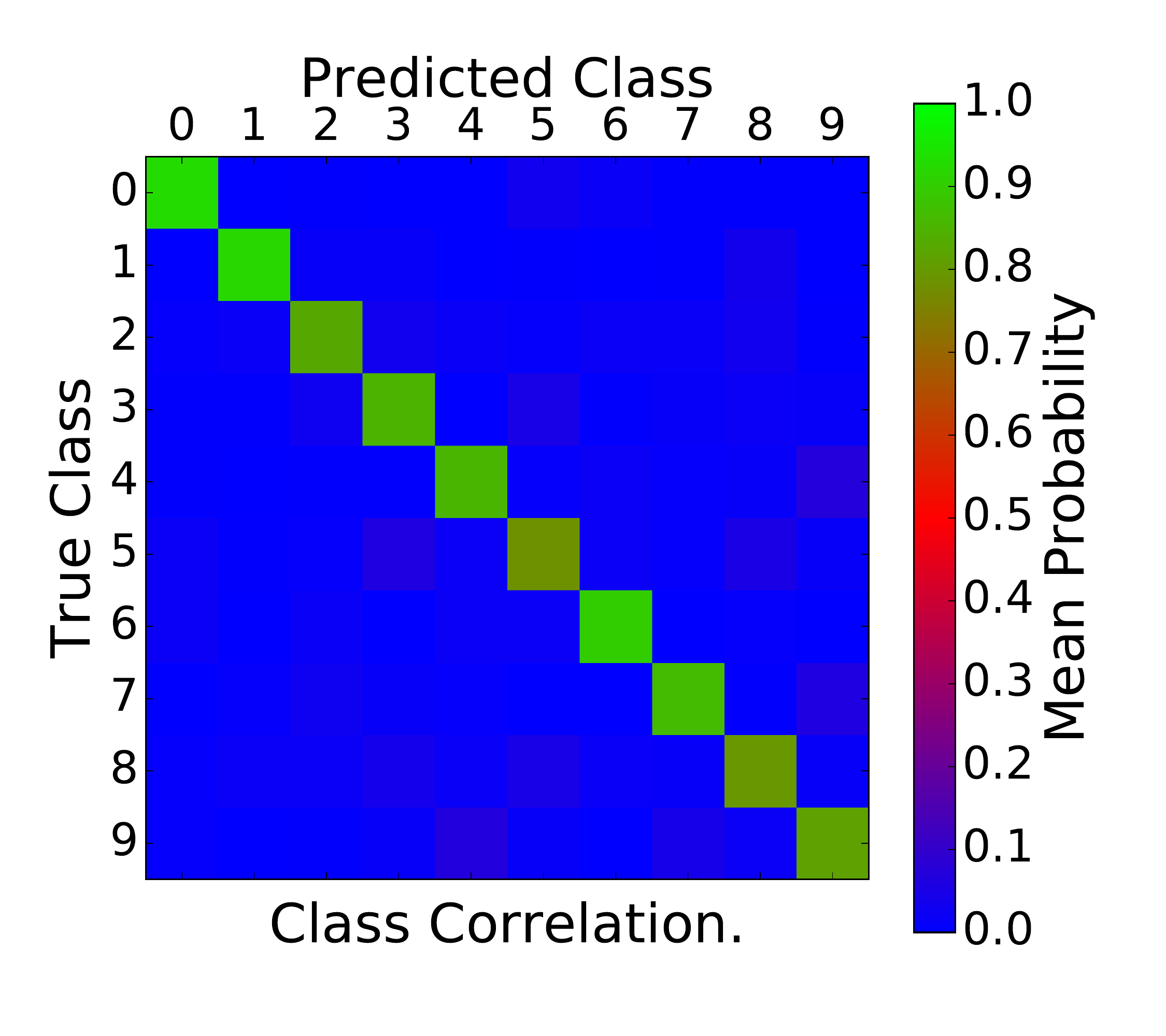}
		(b) SGLD
	\end{minipage}
	\begin{minipage}[t]{0.46\linewidth}
		\centering
		\includegraphics[width=\linewidth]{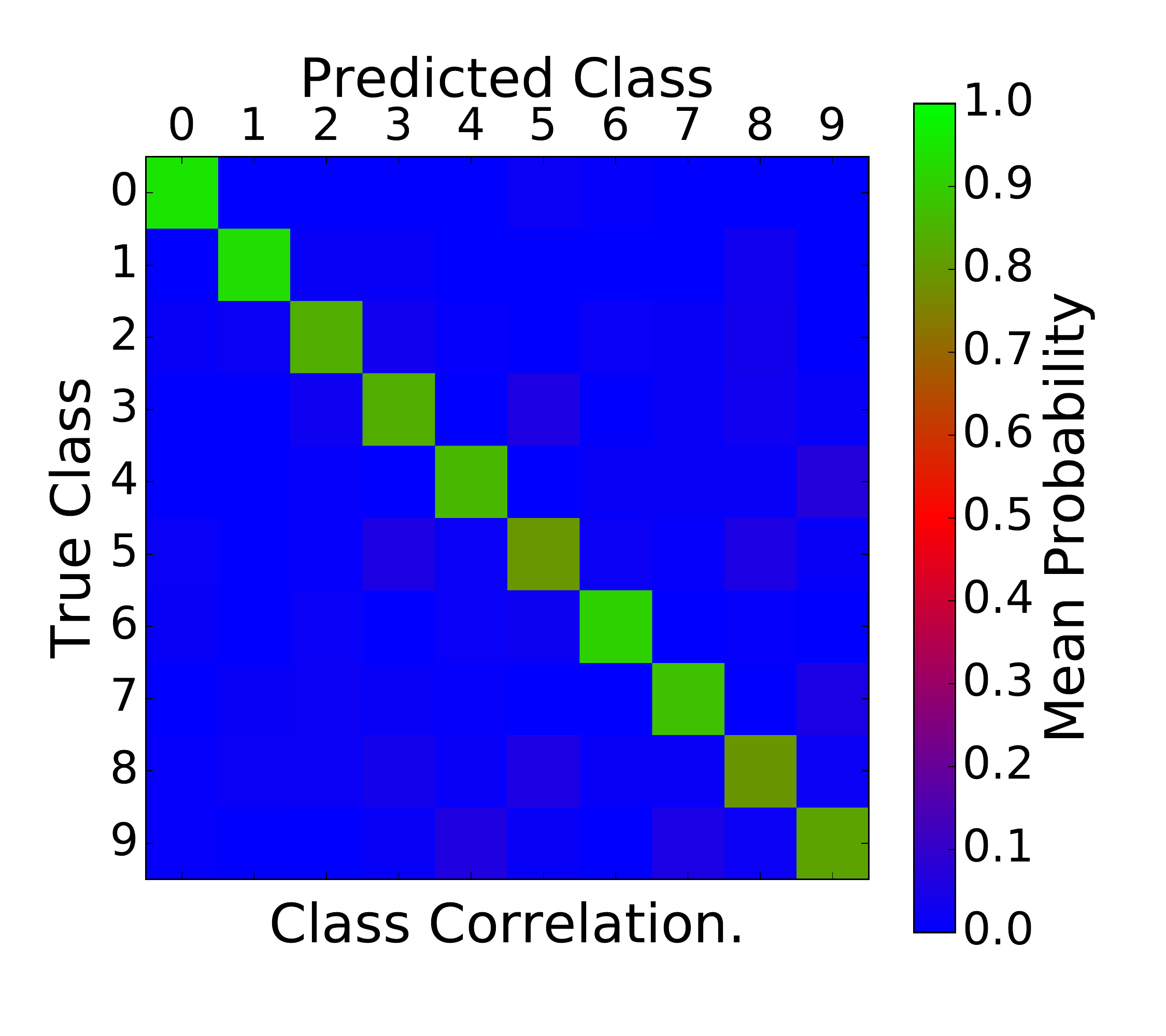}
		(c) D-SGHMC ($p = 0.9$)
	\end{minipage}
	\begin{minipage}[t]{0.46\linewidth}
		\centering
		\includegraphics[width=\linewidth]{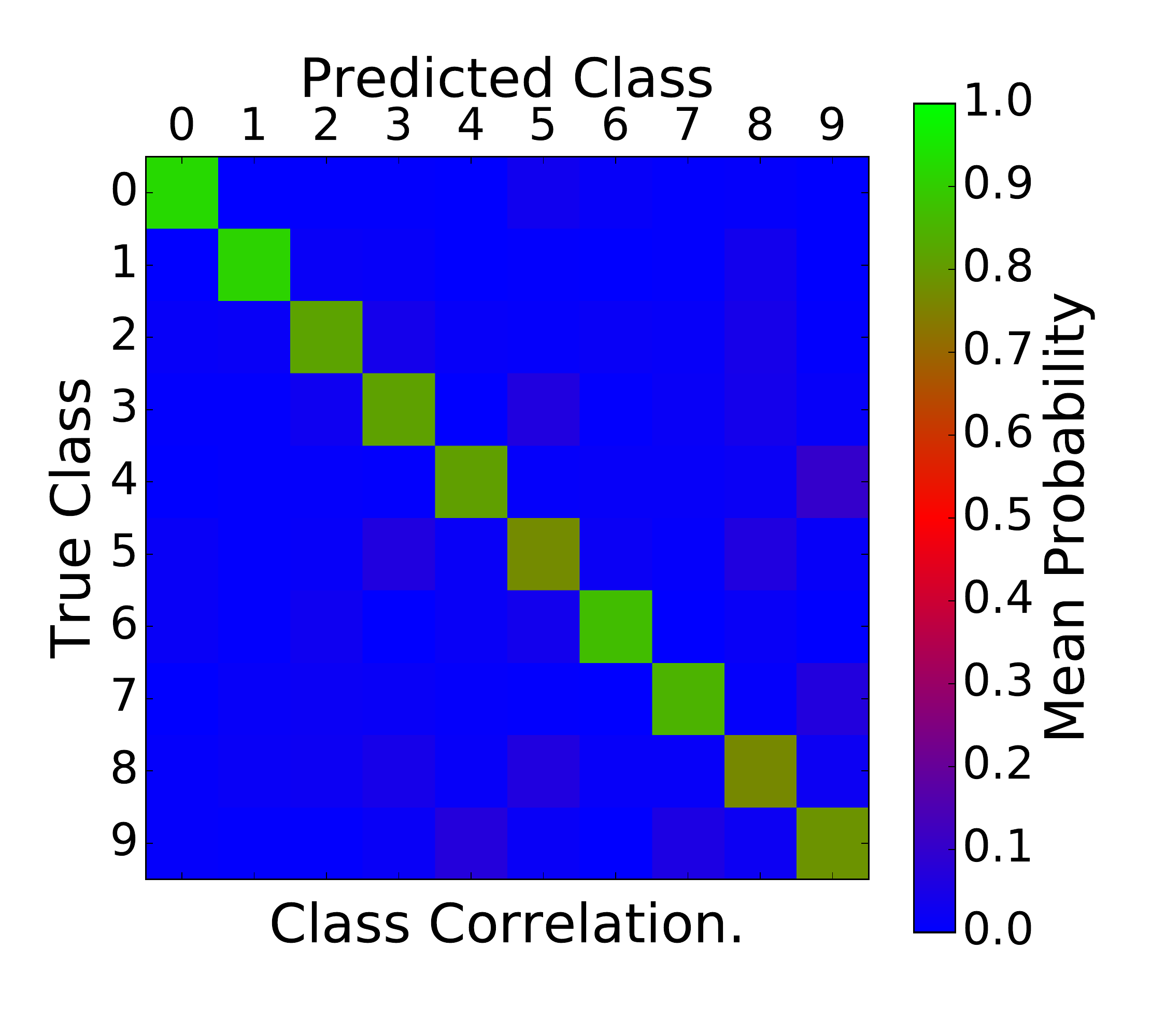}
		(d) D-SGHMC ($p = 0.6$)
	\end{minipage}
	\centering
	\begin{minipage}[t]{0.46\linewidth}
		\centering
		\includegraphics[width=\linewidth]{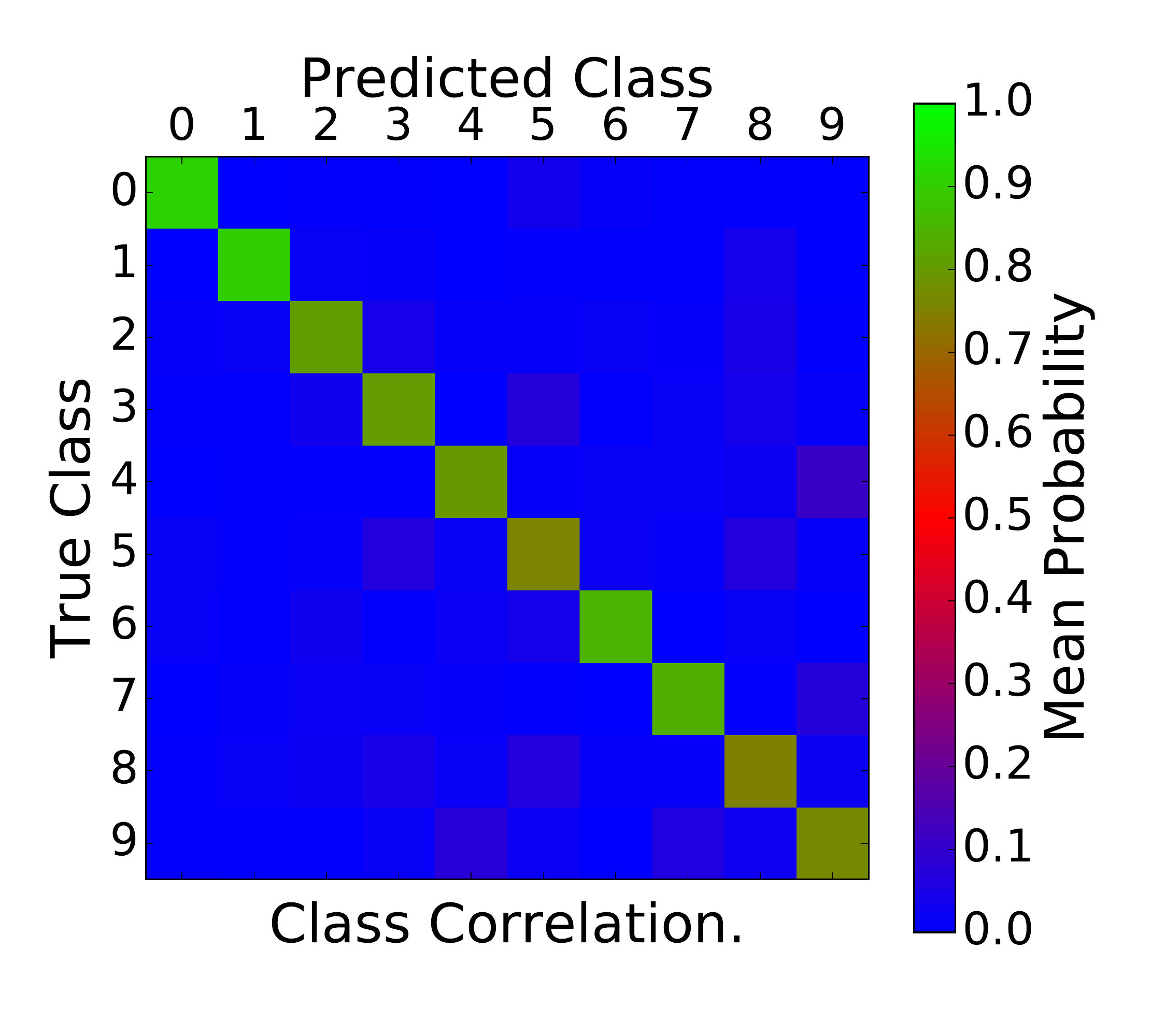}
		(e) D-SGHMC ($p = 0.5$)
	\end{minipage}%
	\begin{minipage}[t]{0.46\linewidth}
		\centering
		\includegraphics[width=\linewidth]{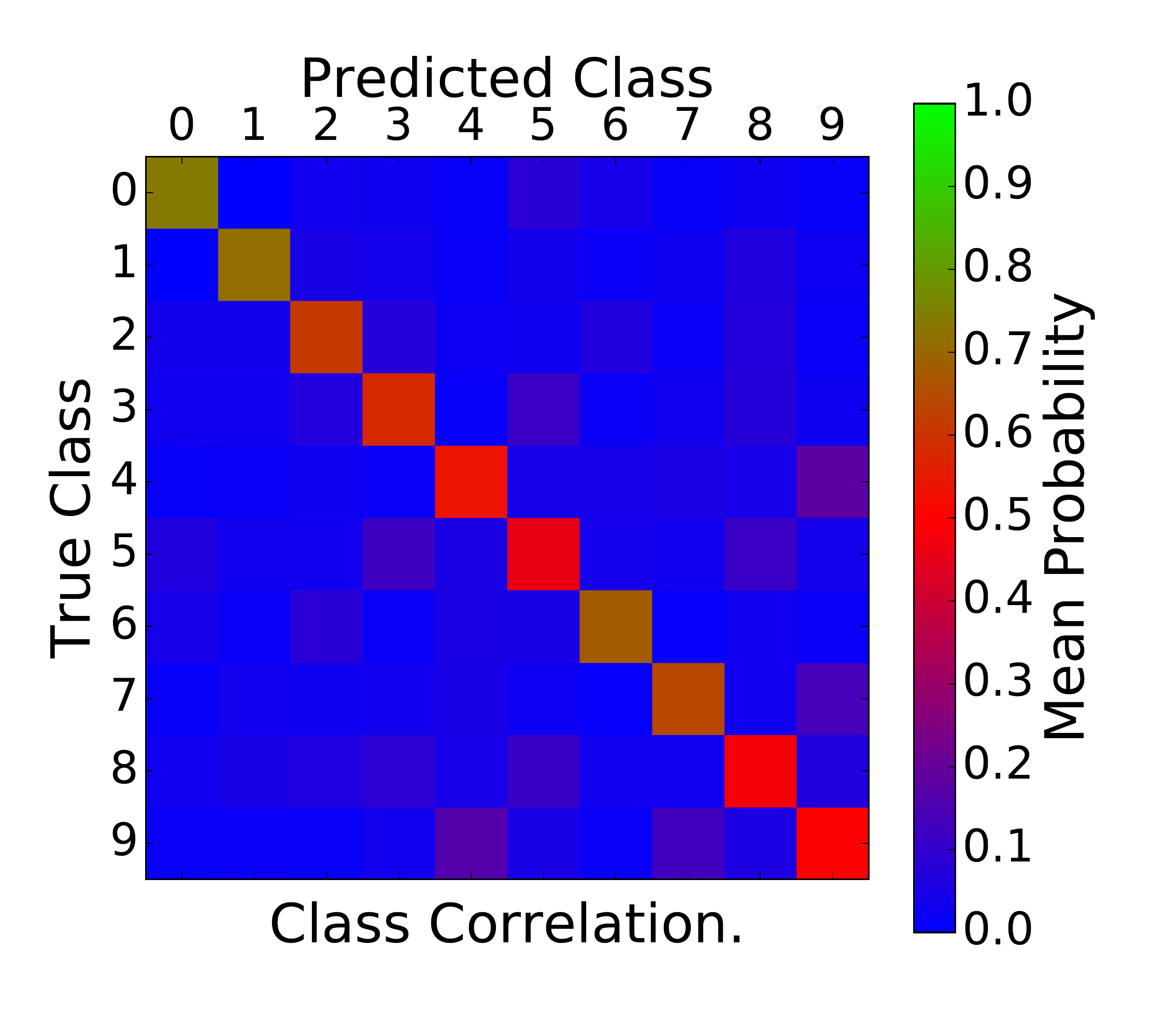}
		(f) D-SGHMC ($p = 0.1$)
	\end{minipage}%
	\caption{Matrix Correlation for error rates on MNIST. }
	\label{fig:mnist_prob_class}
\end{figure}

Figure \ref{fig:mnist_prob_class} shows the correlation matrix between the true class and the predicted class. Higher class uncertainty is achieved when classifying digits $9$ and $8$ of MNIST, which could be classified as digits $4$ and $5$ respectively.

\begin{figure}[htb]
	\begin{minipage}[t]{0.49\linewidth}
		\centering
		\includegraphics[width=\textwidth]{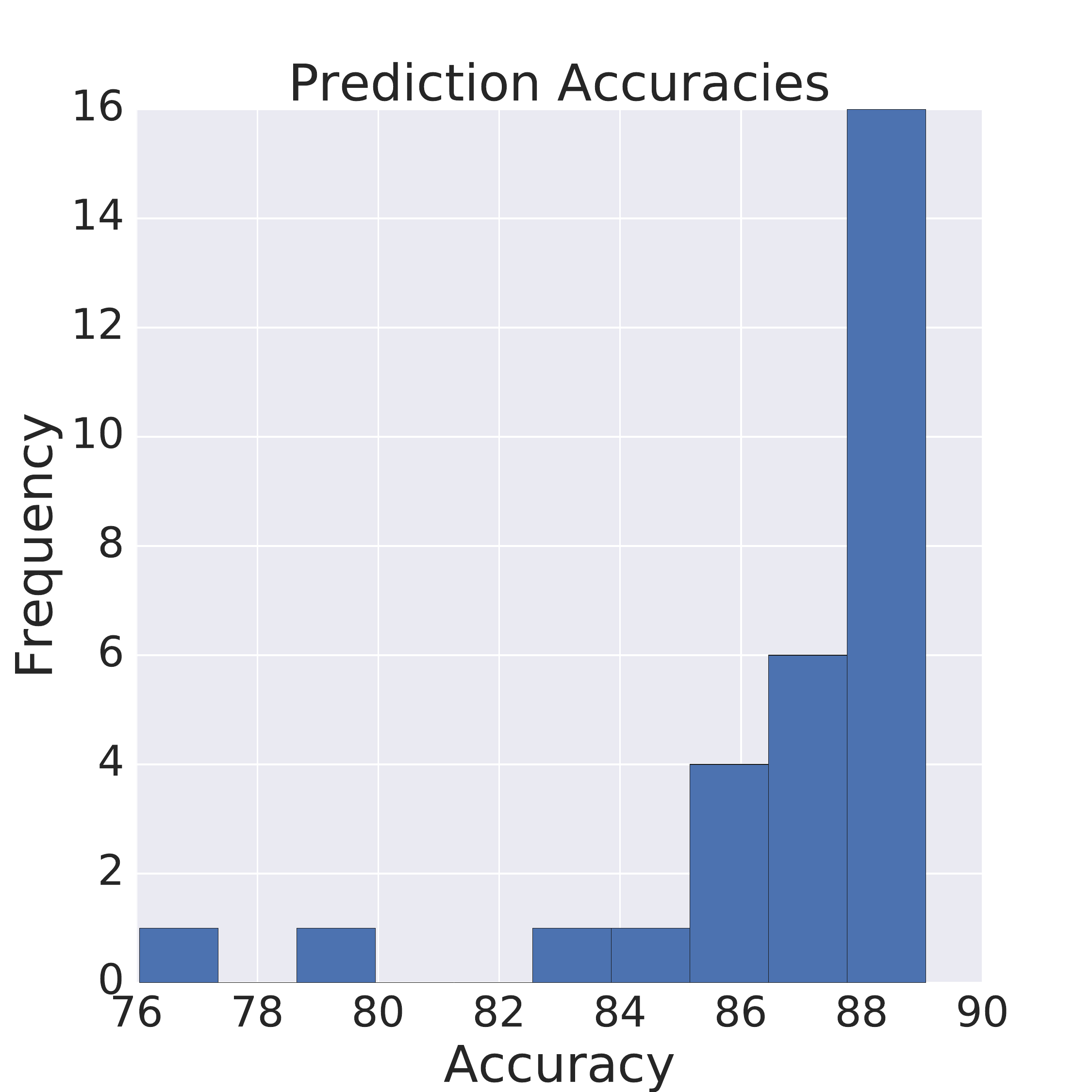}
		(a) D-SGHMC ($p = 0.6$)\\
		Predictive accuracy: $92.1 \%$
	\end{minipage}
	\begin{minipage}[t]{0.49\linewidth}
		\centering
		\includegraphics[width=\textwidth]{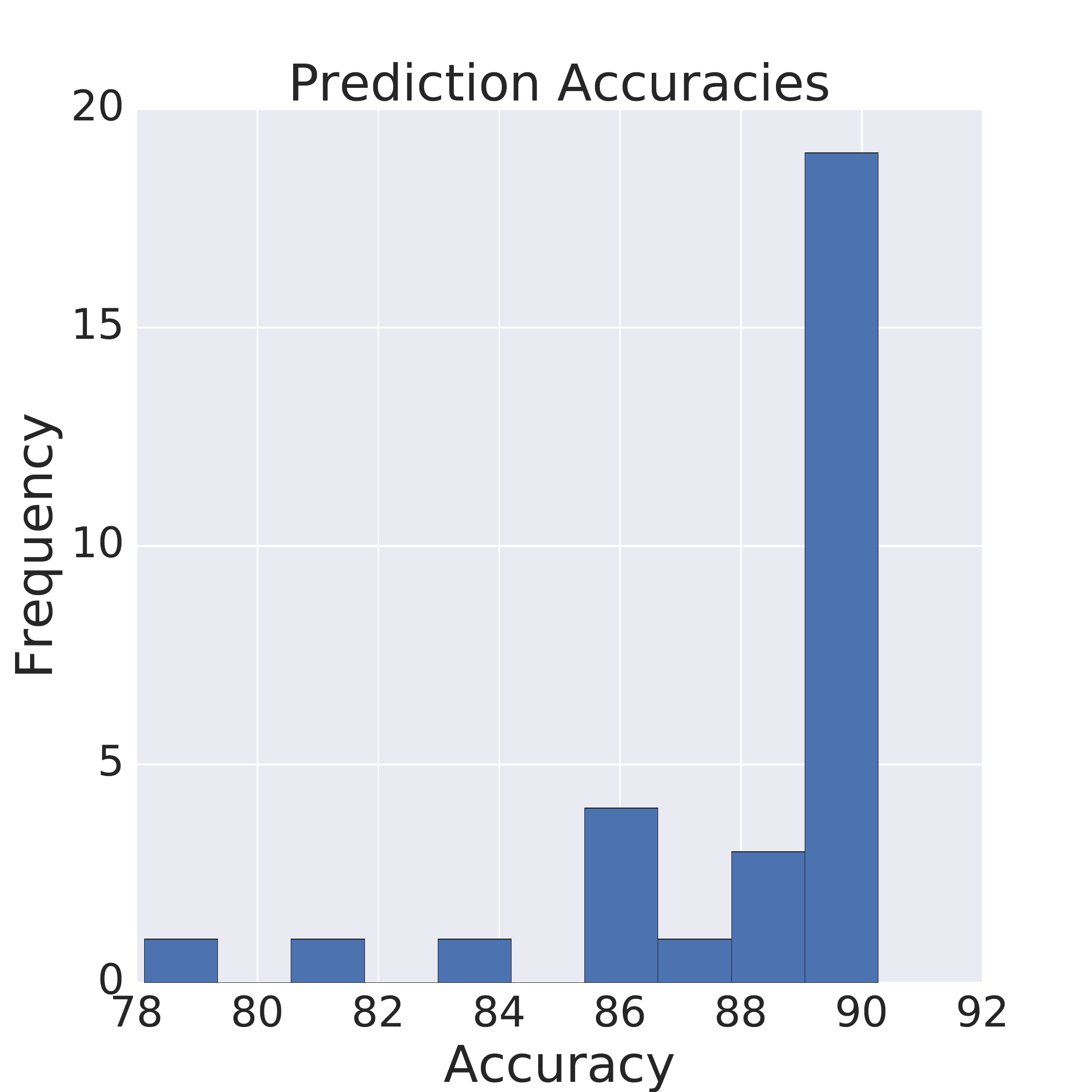}
		(b) D-SGHMC ($p = 0.8$)\\
		Predictive accuracy: $91.9 \%$
	\end{minipage}

    \begin{minipage}[t]{0.49\linewidth}
    	\centering
    	\includegraphics[width=\textwidth]{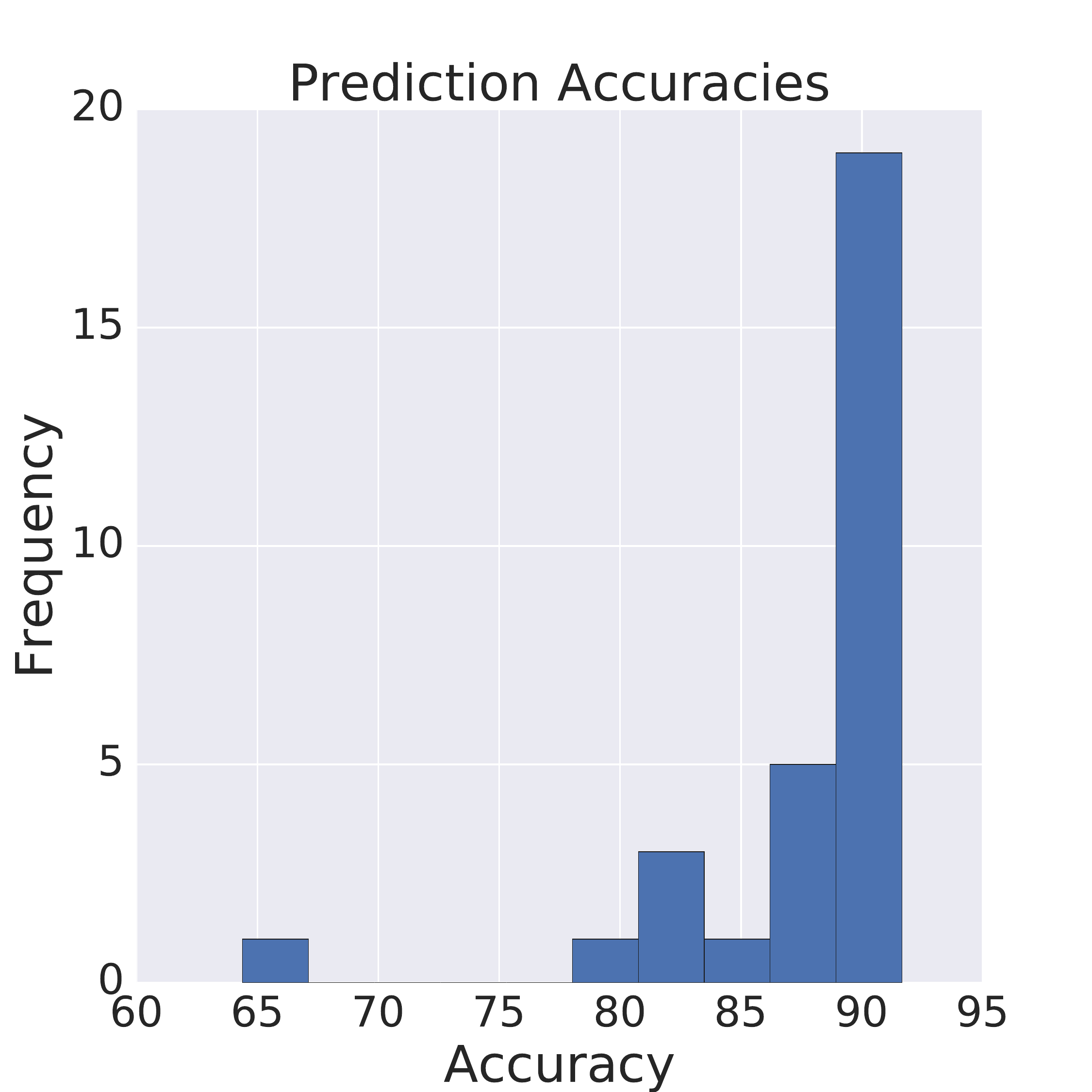}
    	(c) SGHMC\\
    	Predictive accuracy: $90.8 \%$
    \end{minipage}
    \begin{minipage}[t]{0.49\linewidth}
    	\centering
    	\includegraphics[width=\textwidth]{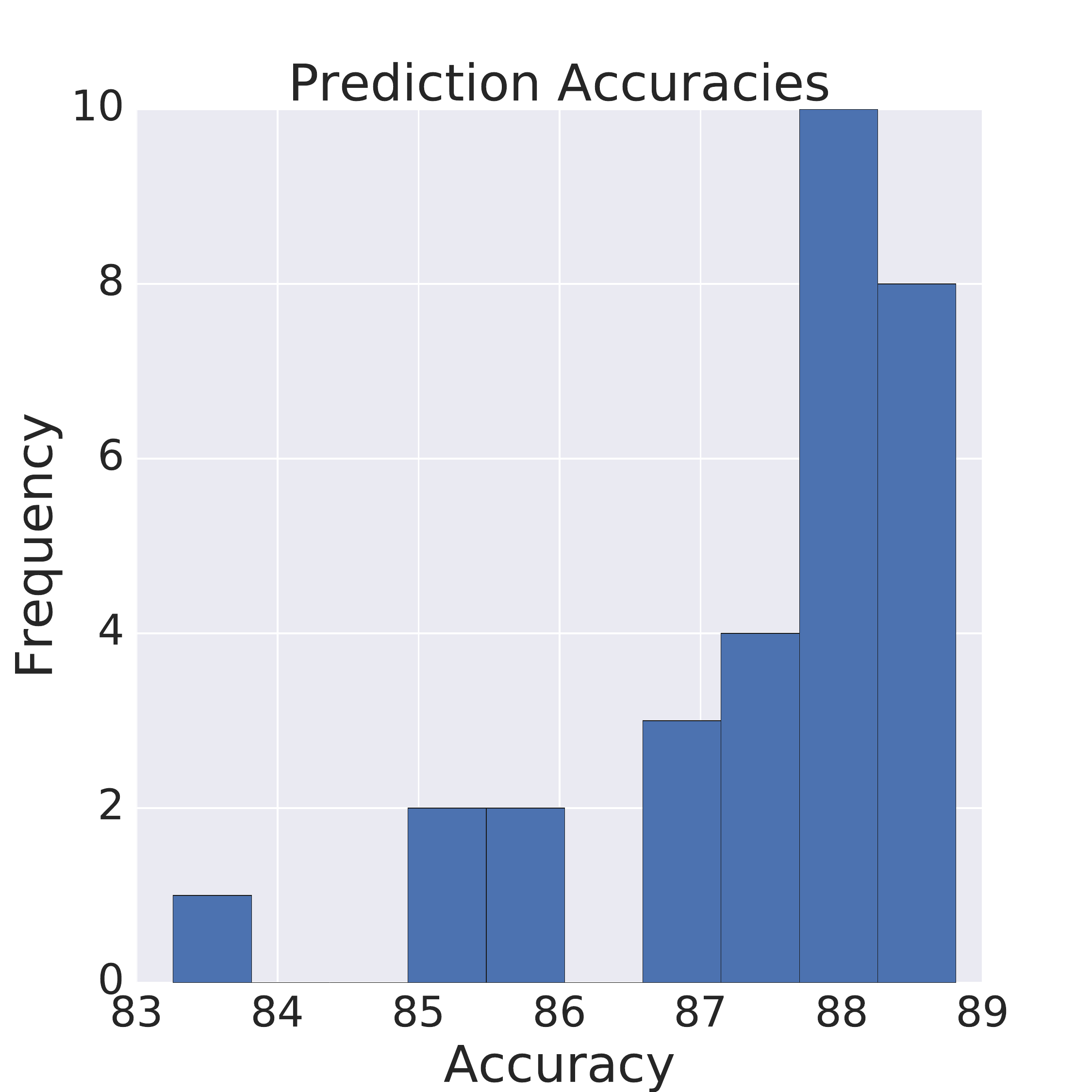}
    	(d) SGLD\\
    	Predictive accuracy: $87.9 \%$
    \end{minipage}
	\caption{Histogram of predictive uncertainty.}
	\label{fig:frecc_acc}
\end{figure}

\clearpage 
Predictive accuracy is now evaluated by comparing the different methods, where the expected predictive accuracy is computed using a Monte Carlo approximation. D-SGHMC (Fig. \ref{fig:frecc_acc}.a and \ref{fig:frecc_acc}.b) achieves higher predictive accuracy when compared to SGHMC (Fig. \ref{fig:frecc_acc}.c) and SGLD (Fig. \ref{fig:frecc_acc}.d).

\subsubsection{Confusing classes}

One of the biggest challenges of the MNIST data set is to separate highly confusing digits such as '$4$' and '$9$'. This problem has been addressed earlier in feature selection challenges \cite{guyon2005result}.

\begin{figure}
	\begin{minipage}[t]{\textwidth}
	\begin{minipage}[t]{0.46\linewidth}
		\centering
		\includegraphics[width=\linewidth]{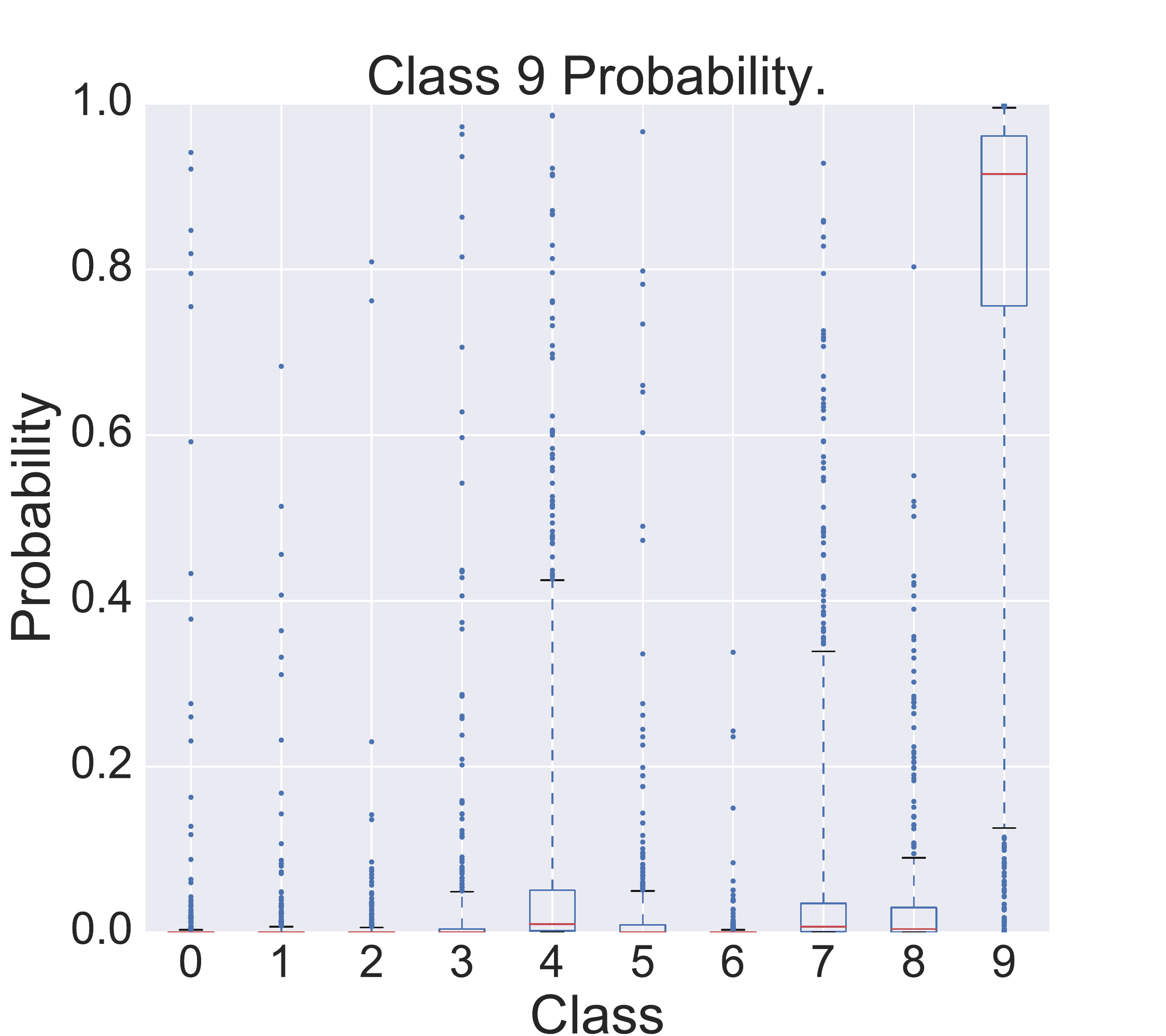}
		(a) SGHMC\\
	Predictive class accuracy: $87 \%$
	\end{minipage}
	\begin{minipage}[t]{0.46\linewidth}
		\centering
		\includegraphics[width=\linewidth]{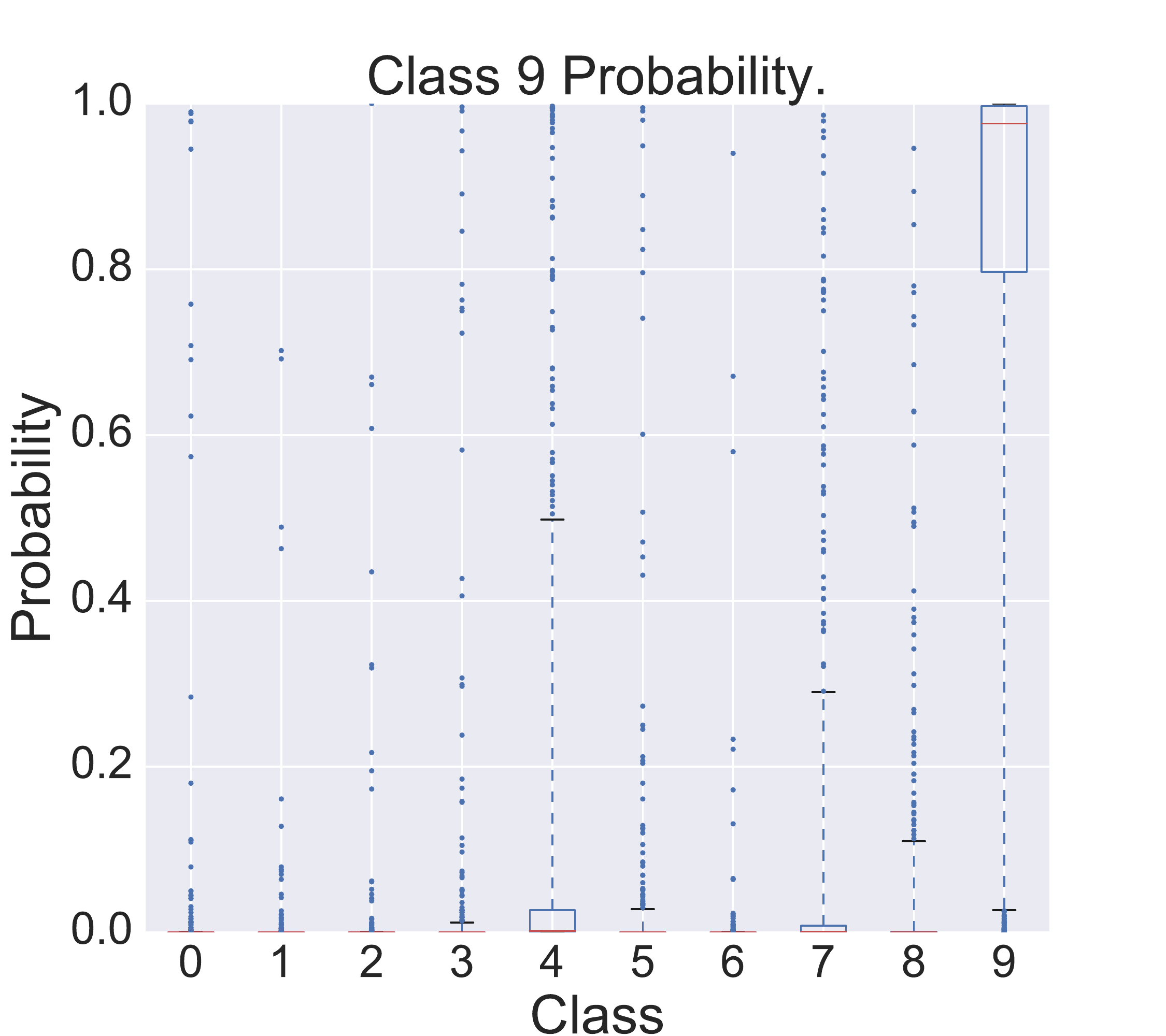}
		(b) SGLD\\
	Predictive class accuracy:  $84 \%$
	\end{minipage}
	\begin{minipage}[t]{0.46\linewidth}
		\centering
		\includegraphics[width=\linewidth]{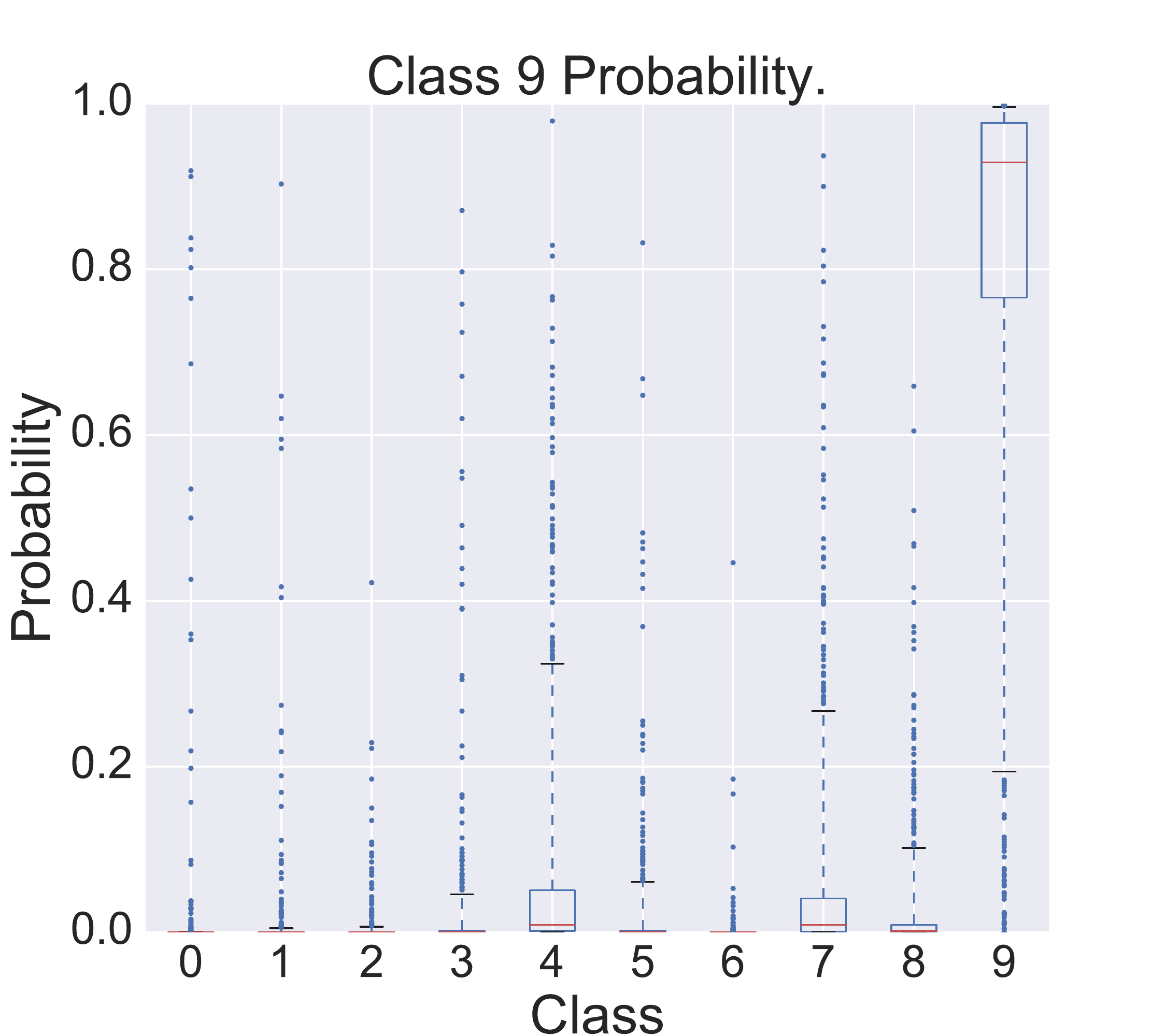}
		(c) D-SGHMC ($p = 0.9$)\\
	Predictive class accuracy:  $90 \%$
	\end{minipage}
	\begin{minipage}[t]{0.46\linewidth}
		\centering
		\includegraphics[width=\linewidth]{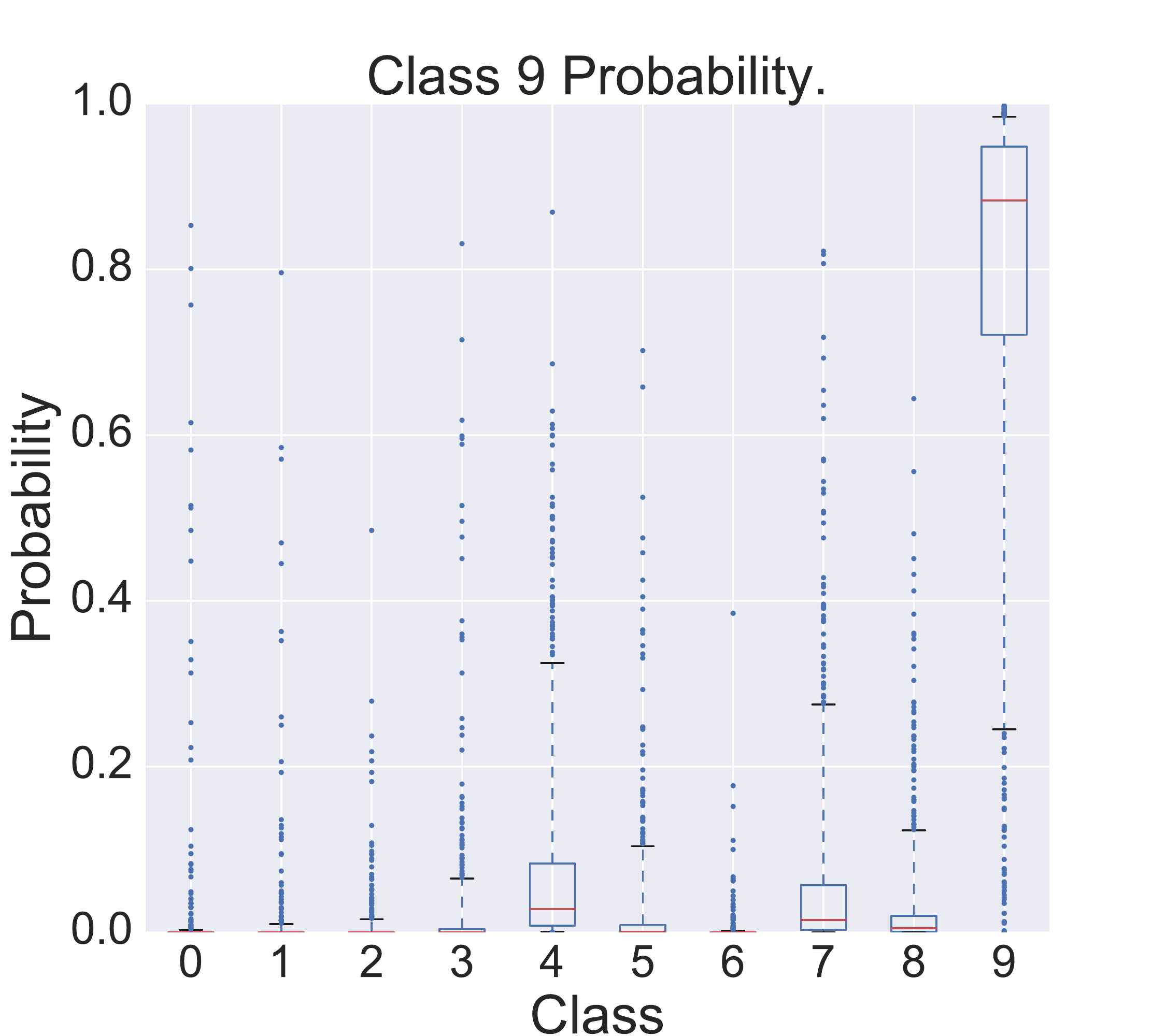}
		(d) D-SGHMC ($p = 0.6$)\\
	Predictive class accuracy:  $92 \%$
	\end{minipage}
	\centering
	\begin{minipage}[t]{0.46\linewidth}
		\centering
		\includegraphics[width=\linewidth]{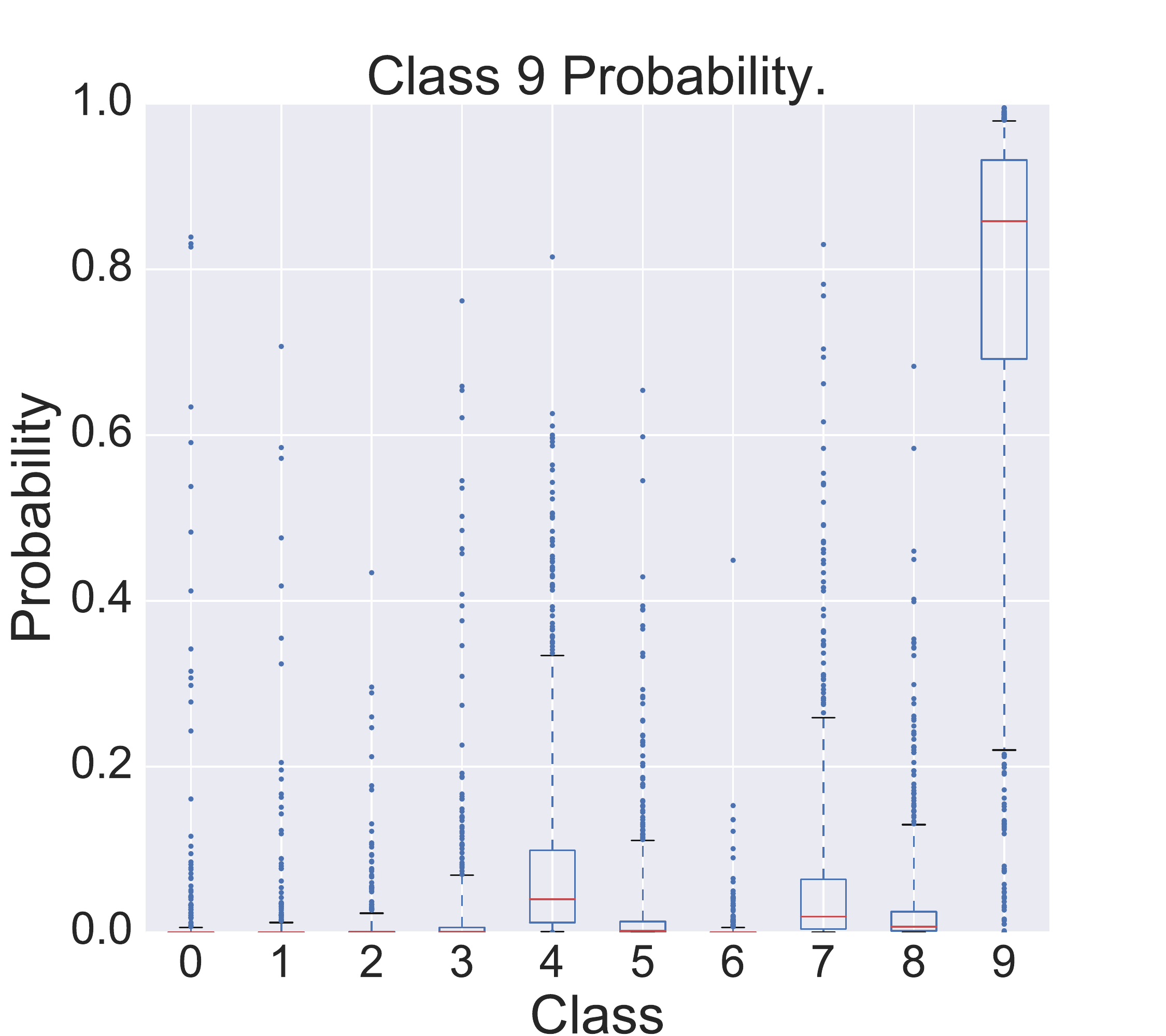}
		(e) D-SGHMC ($p = 0.5$)\\
	Predictive class accuracy:  $91 \%$
	\end{minipage}%
		\begin{minipage}[t]{0.46\linewidth}
		\centering
		\includegraphics[width=\linewidth]{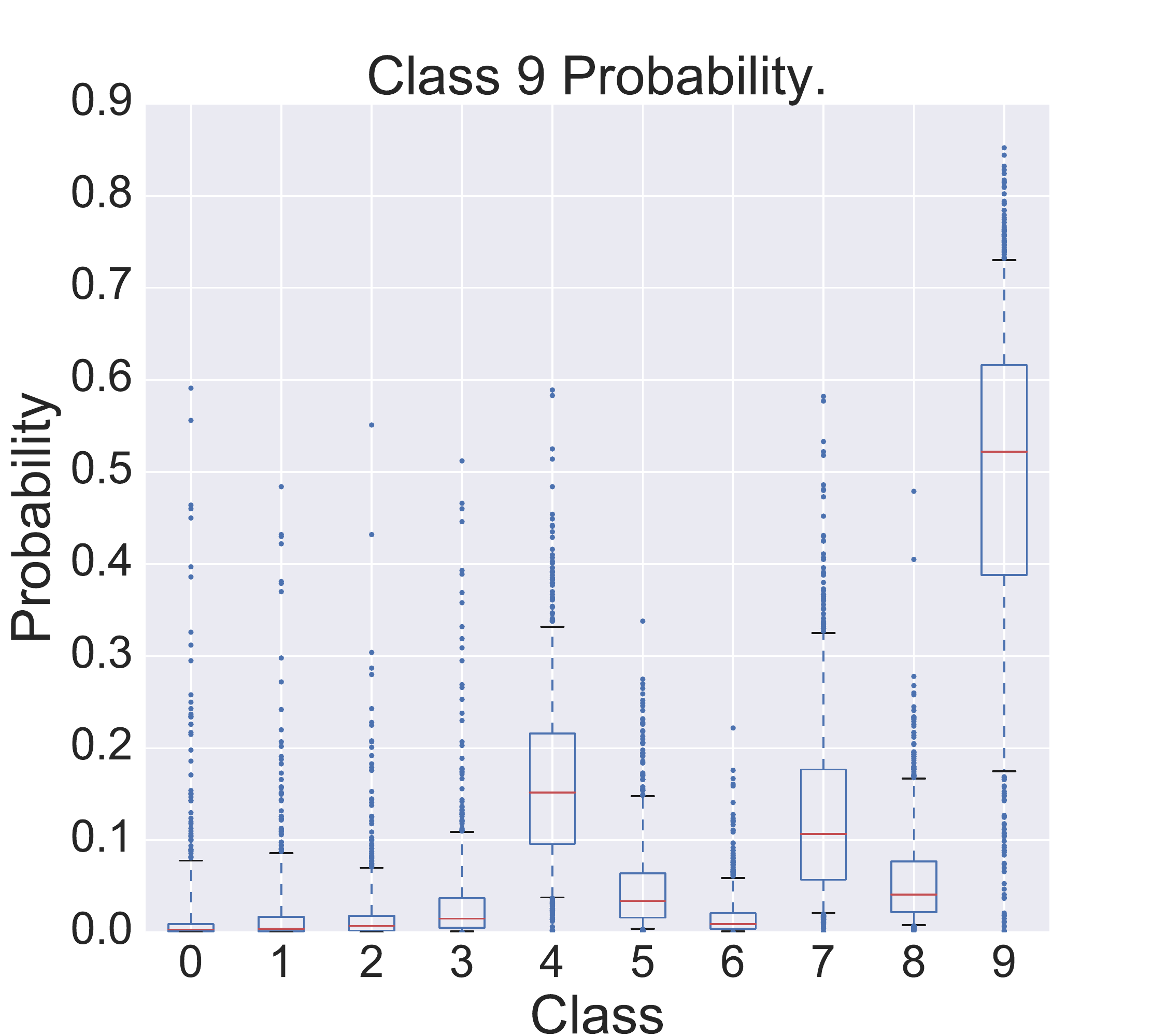}
		(f) D-SGHMC ($p = 0.1$)\\
	Predictive class accuracy:  $86 \%$
	\end{minipage}%

	\caption{Box plot of the error rates for class $9$.}
	\label{fig:prob_c9}
\end{minipage}
\end{figure}

We analyze the behavior between total uncertainty and predictive accuracy for one of these digits in each method. We can observe that there is less over-confidence and at the same time the classification results improve. Figure \ref{fig:prob_c9} shows the mean expected probabilities for digit 9 in the test set.

As the Dropout rate decreases, the uncertainty around similar classes becomes more visible. However, low Dropout probabilities does not guarantee better classification results or improved uncertainty estimates. As shown in the Figure \ref{fig:prob_c9}, models generated with low Dropout rates aggressively reduce the number of variables used for prediction.

\subsubsection{Confusing Example (digit 9)}

\begin{figure}[htb]
	\begin{minipage}[t]{0.46\linewidth}
		\centering
		\includegraphics[width=1.1\textwidth]{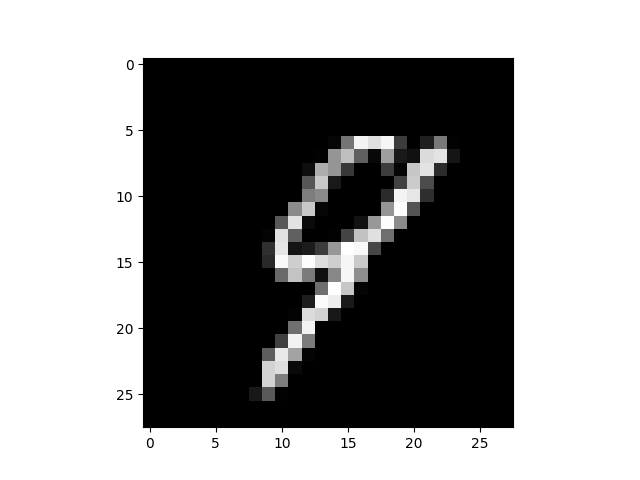}
		(a) Example digit $9$.
	\end{minipage}
	\begin{minipage}[t]{0.46\linewidth}
		\centering
		\includegraphics[width=1\textwidth]{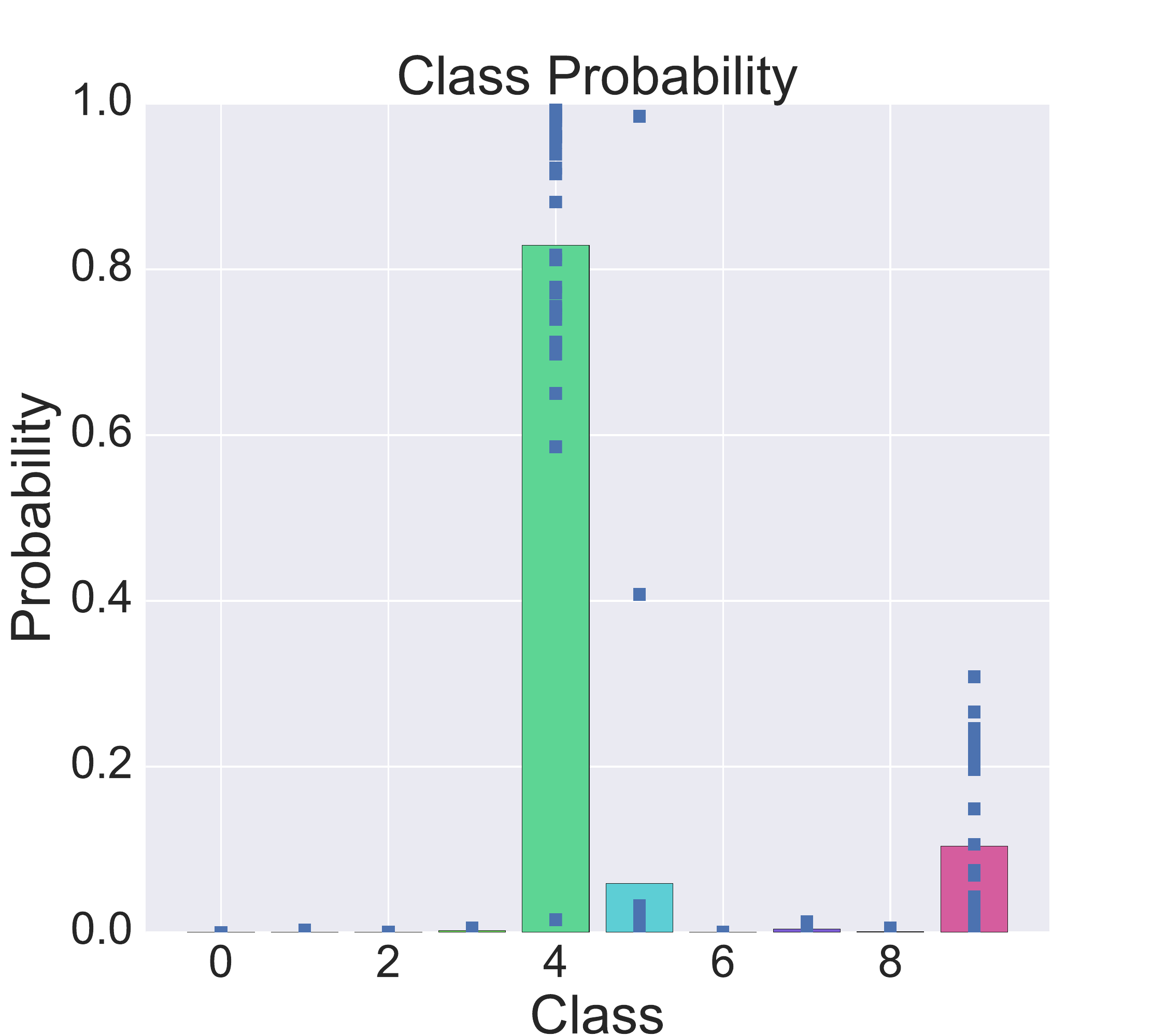}
		(b) SGHMC
	\end{minipage}
	\begin{minipage}[t]{0.46\linewidth}
		\centering
		\includegraphics[width=1\textwidth]{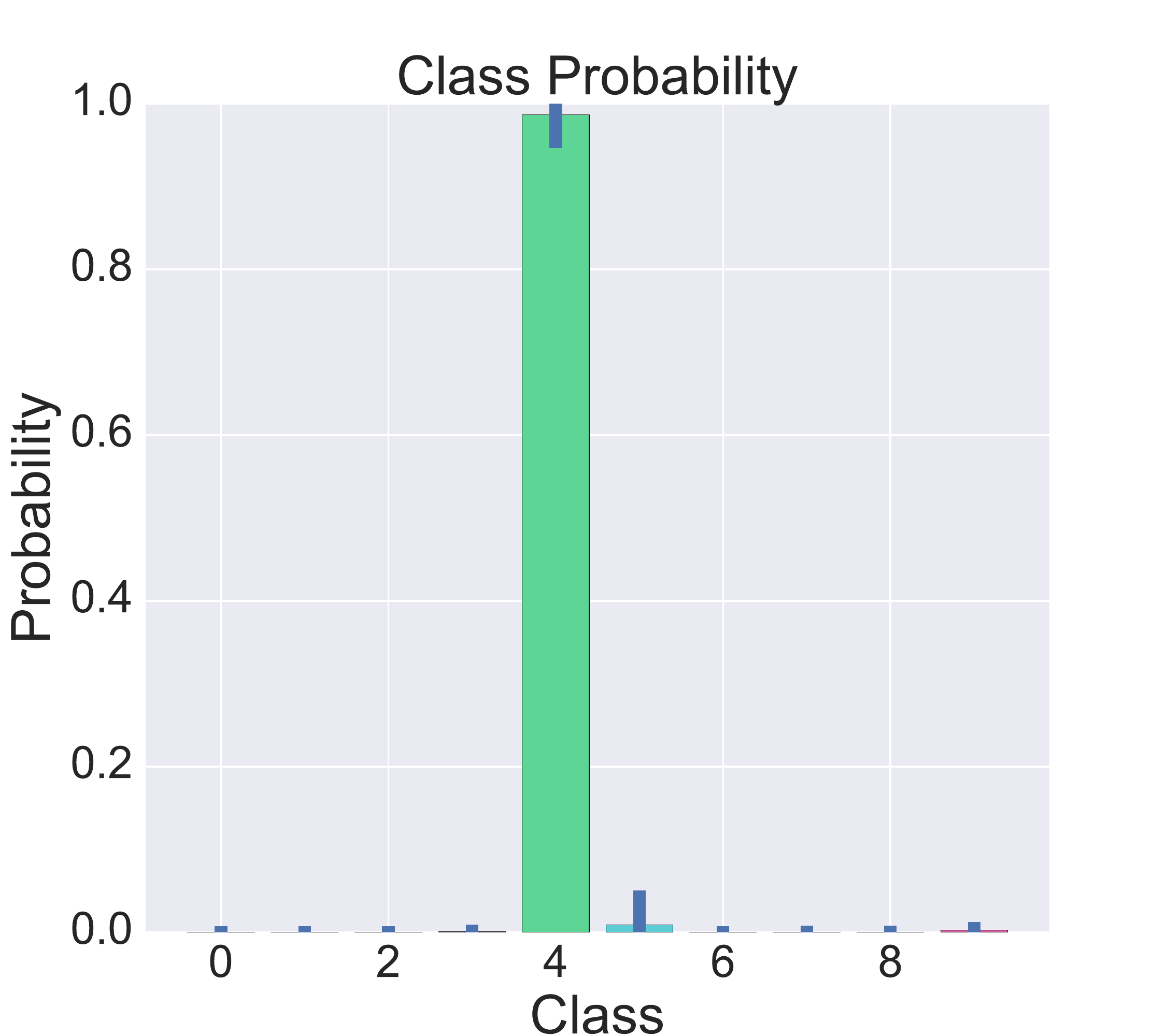}
		(c) SGLD
	\end{minipage}
	\begin{minipage}[t]{0.46\linewidth}
		\centering
		\includegraphics[width=1\textwidth]{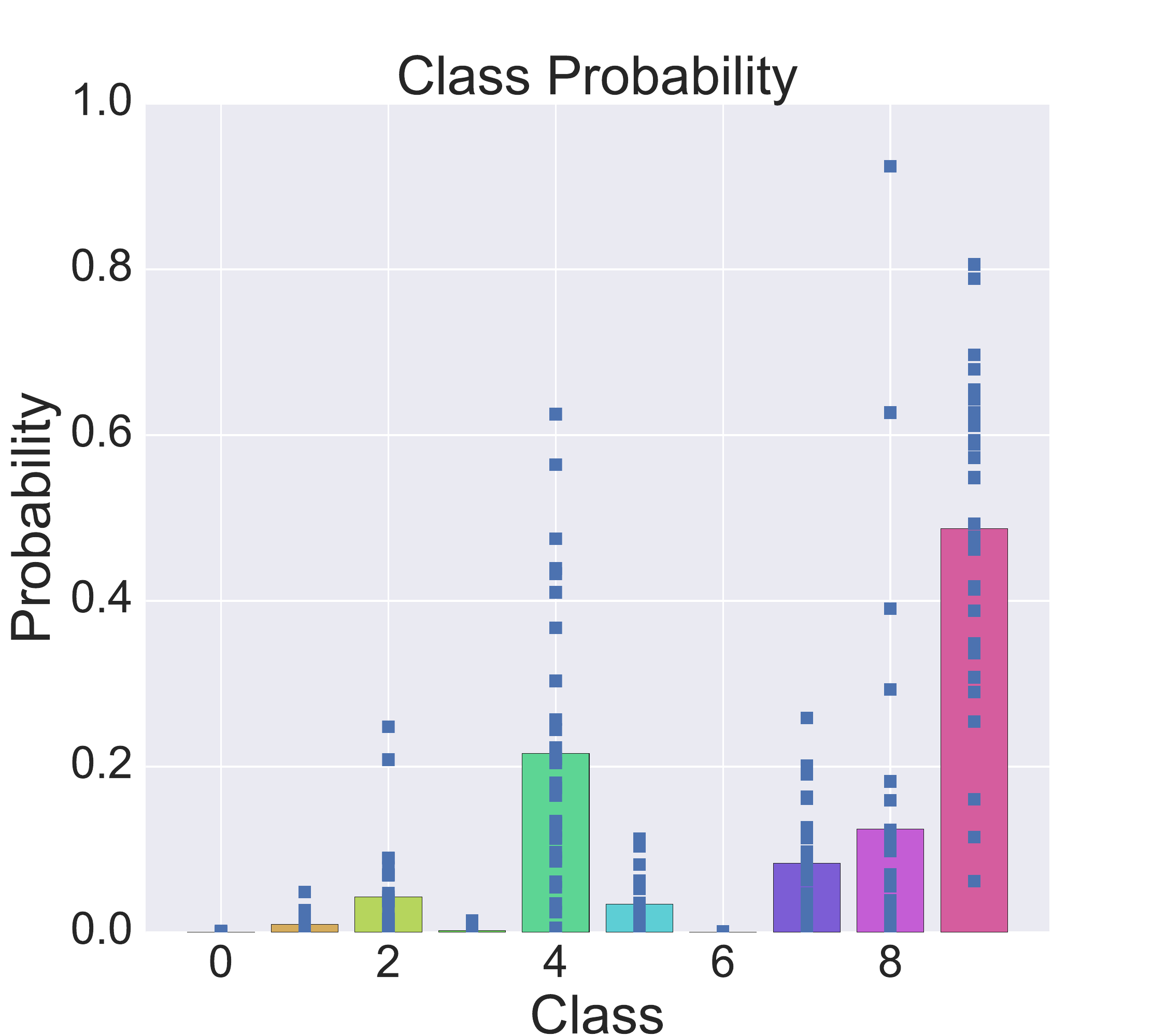}
		(d) D-SGHMC ($p = 0.9$)
	\end{minipage}
	\centering
	\begin{minipage}[t]{0.46\linewidth}
		\centering
		\includegraphics[width=1\textwidth]{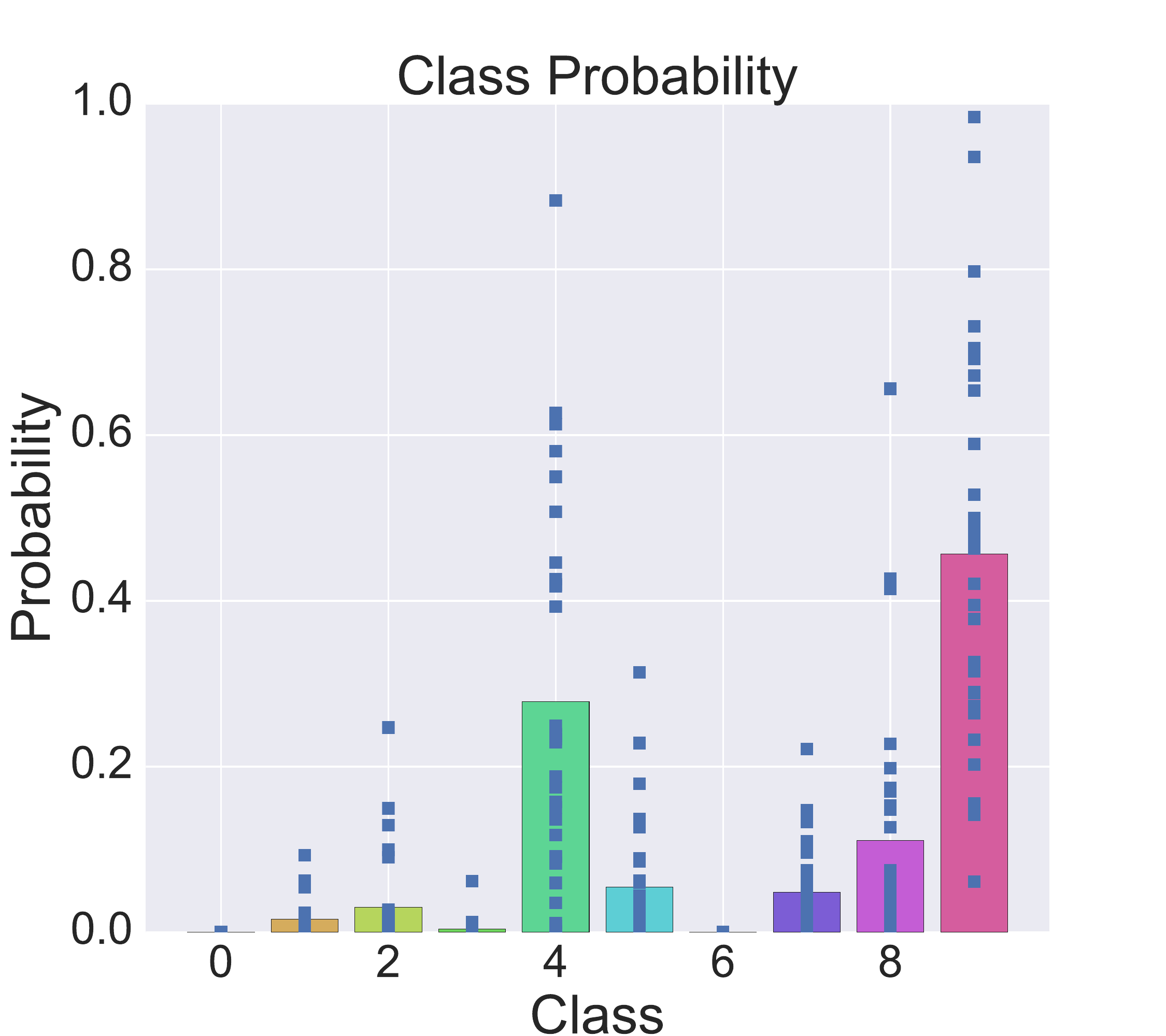}
		(e) D-SGHMC ($p = 0.7$)
	\end{minipage}
	\begin{minipage}[t]{0.46\linewidth}
	\centering
	\includegraphics[width=1\textwidth]{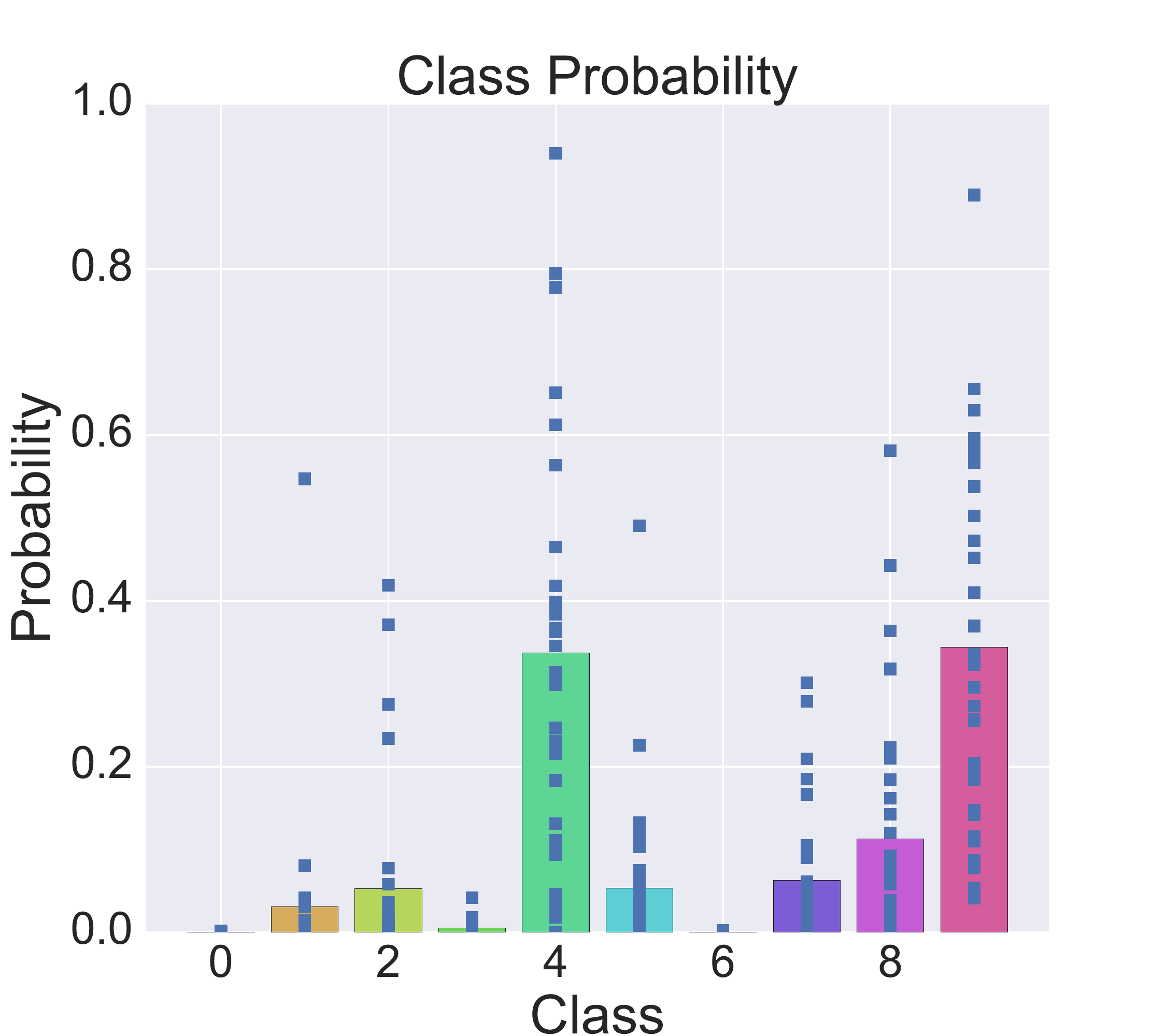}
	(f) D-SGHMC ($p = 0.5$)
    \end{minipage}
	\caption{Bar plot of error rates for confusing example $9$. }
	\label{fig:ex9}
\end{figure}

In the first example (Fig. \ref{fig:ex9}.a), both state-of-the-art SGHMC (Fig. \ref{fig:ex9}.b) and SGLD methods (Fig. \ref{fig:ex9}.c) maintain a high confidence. 

In the example we can see how the proposed method decreases the over-confidence generated by SGHMC  and SGLD  and allows to classify the example effectively, producing higher uncertainty (Fig. \ref{fig:ex9}.d, \ref{fig:ex9}.e and \ref{fig:ex9}.f).

\subsubsection{Confusing Example (digit 8)}
	
\begin{figure}[htb]
	\begin{minipage}[t]{0.46\linewidth}
		\centering
		\includegraphics[ width=1.1\textwidth]{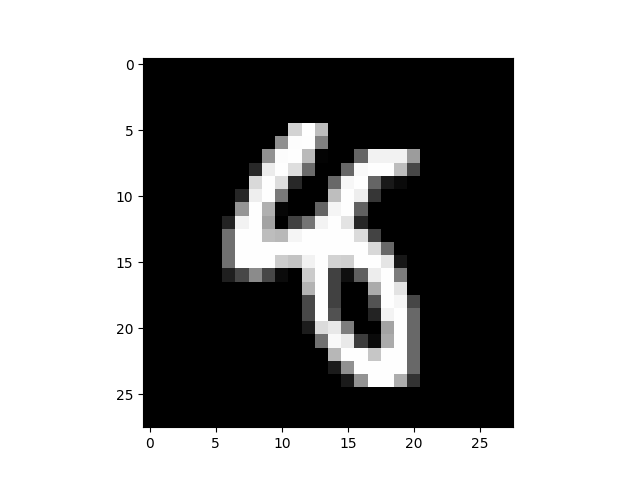}
		(a) Example digit $8$.
	\end{minipage}
	\begin{minipage}[t]{0.46\linewidth}
		\centering
		\includegraphics[width=1\textwidth]{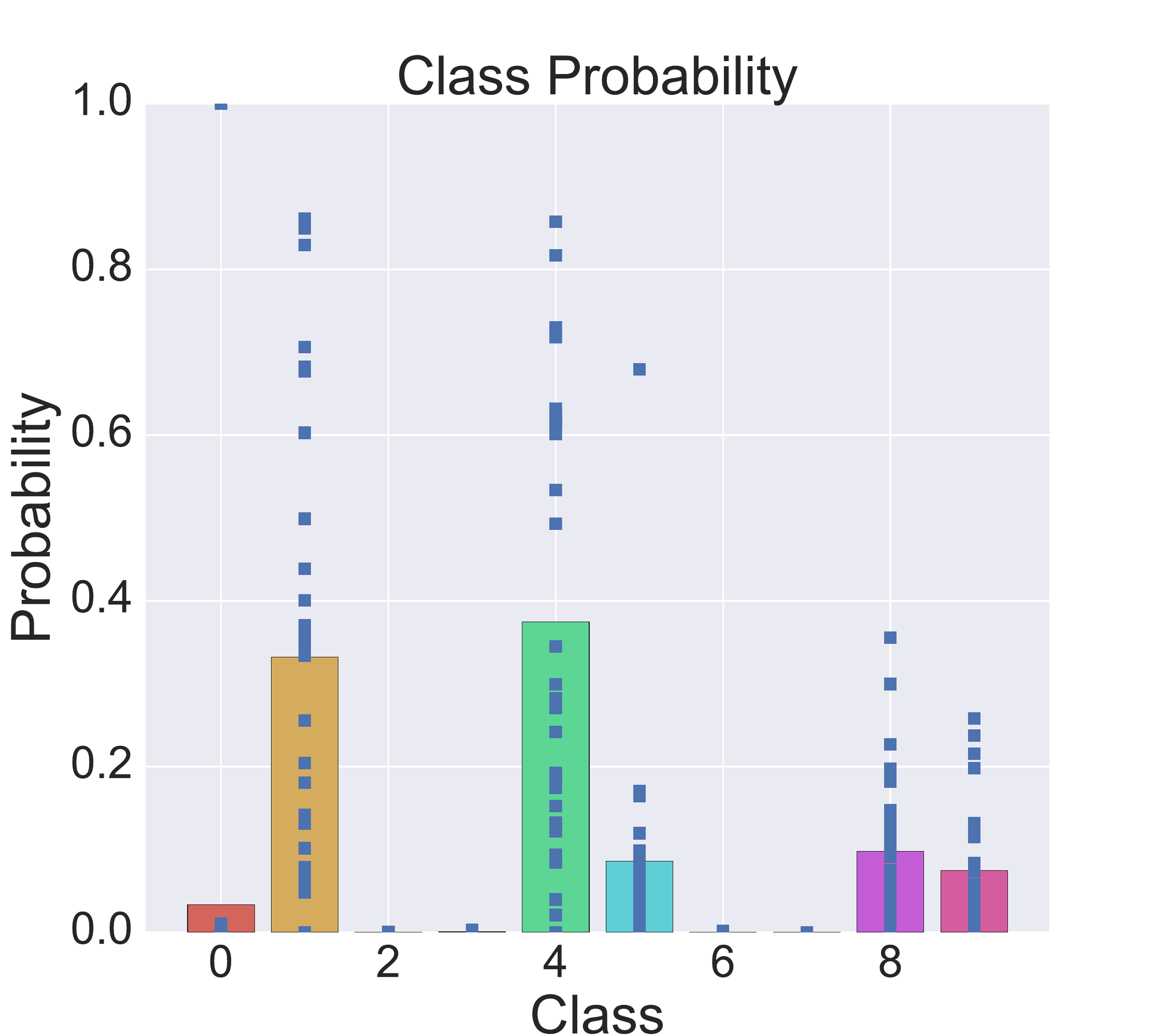}
		(b) SGHMC
	\end{minipage}
	\begin{minipage}[t]{0.46\linewidth}
		\centering
		\includegraphics[width=1\textwidth]{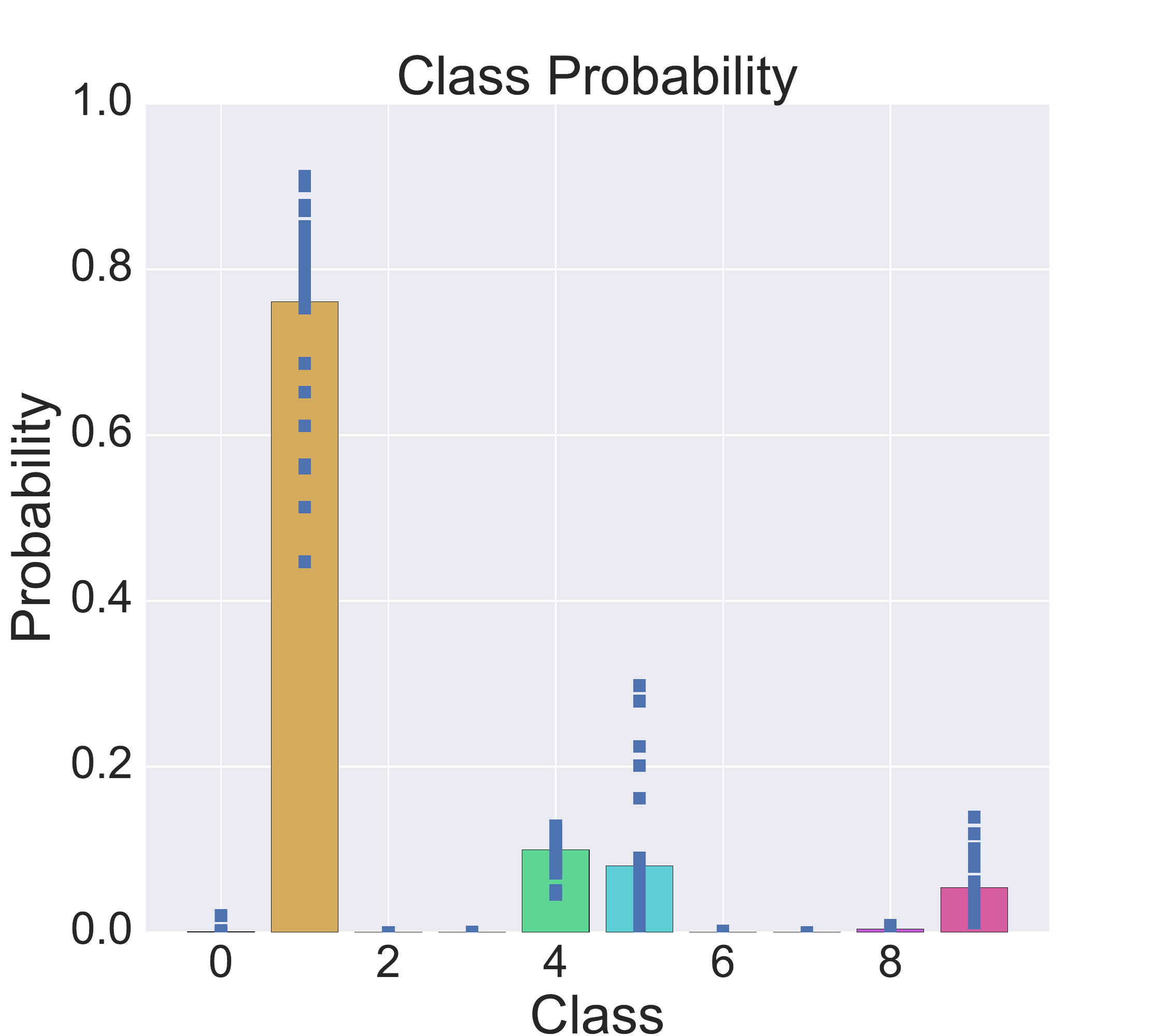}
		(c) SGLD
	\end{minipage}
	\begin{minipage}[t]{0.46\linewidth}
		\centering
		\includegraphics[width=1\textwidth]{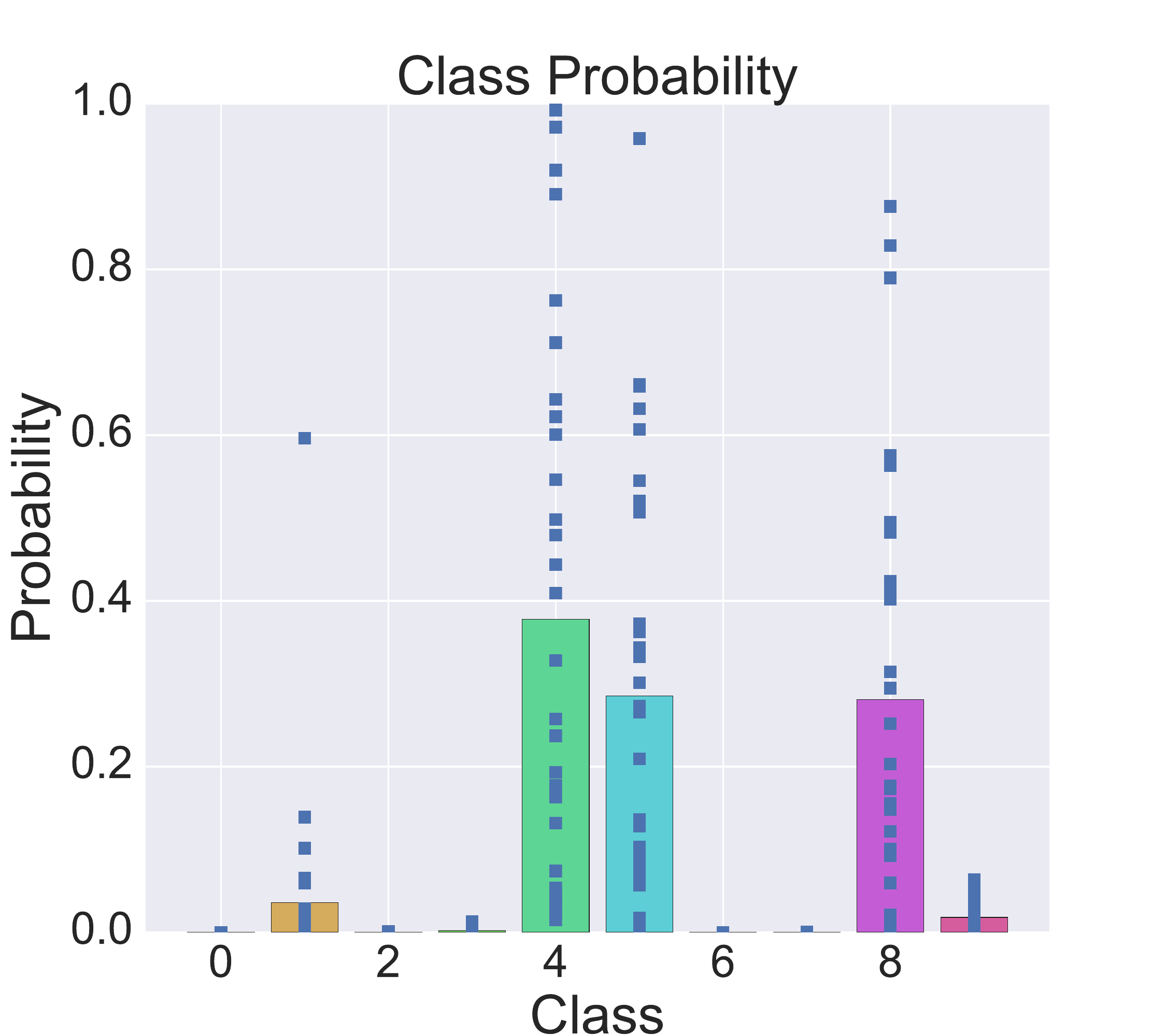}
		(d) D-SGHMC ($p = 0.9$)
	\end{minipage}
	\centering
	\begin{minipage}[t]{0.46\linewidth}
		\centering
		\includegraphics[width=1\textwidth]{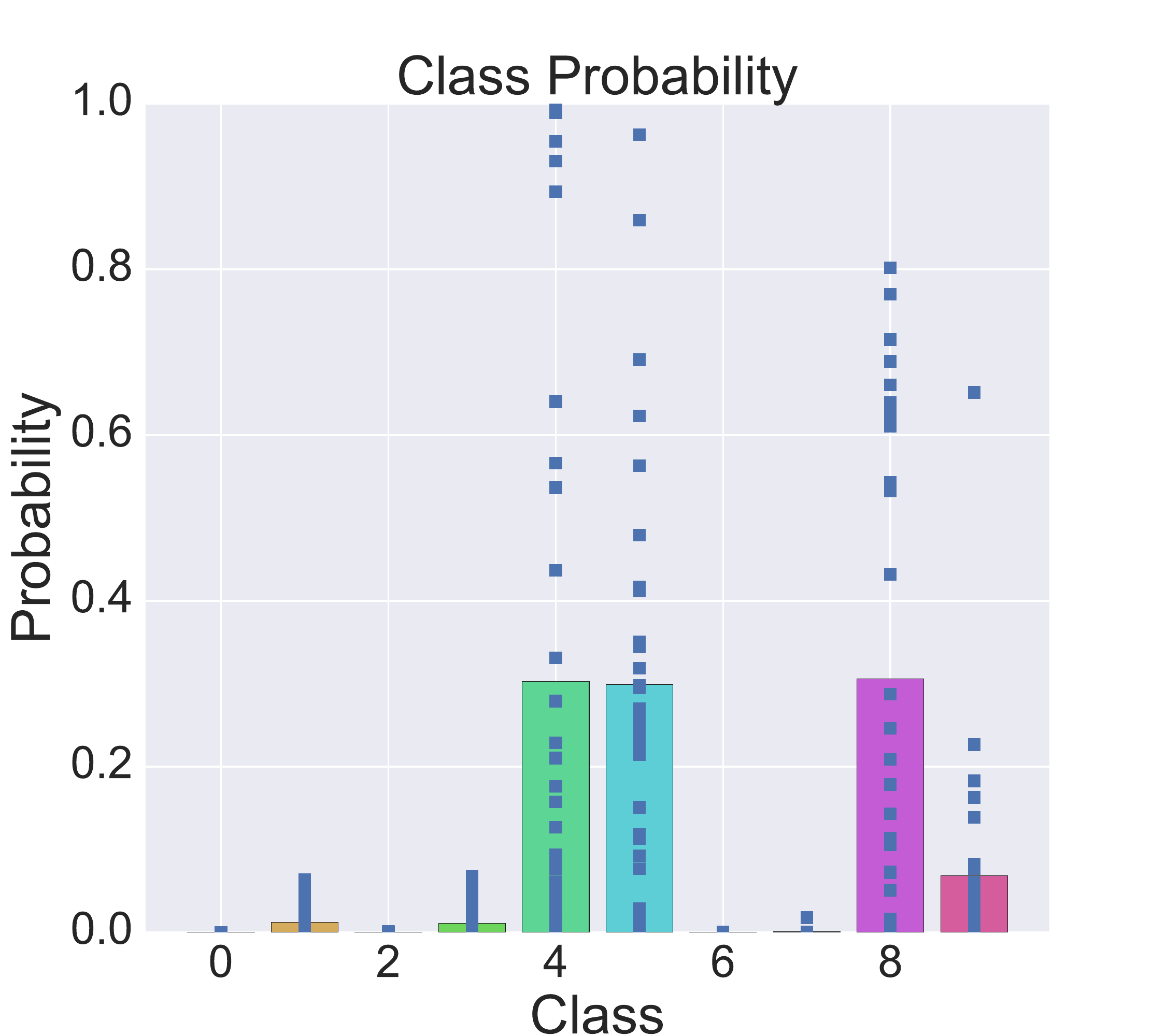}
		(e) D-SGHMC ($p = 0.7$)
	\end{minipage}
	\begin{minipage}[t]{0.46\linewidth}
		\centering
	\includegraphics[width=1\textwidth]{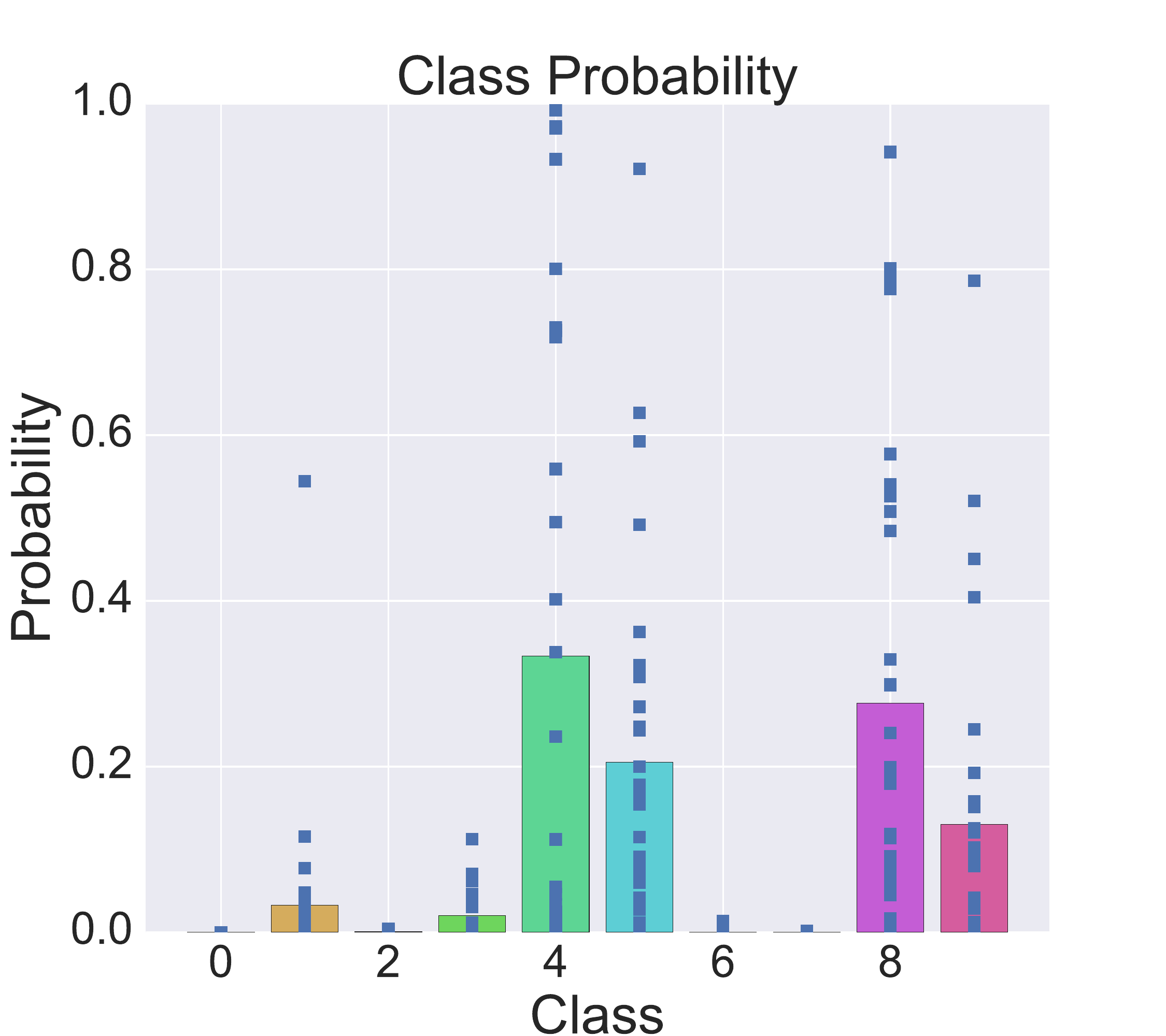}
	(f) D-SGHMC ($p = 0.5$)
	\end{minipage}%
	\caption{Bar plot of error rates for confusing example $8$. }
	\label{fig:ex8}
\end{figure}

In the second example (Fig. \ref{fig:ex8}.a), we observe that it is similar to the digits '$4$', '$5$' and even '$1$', for thus SGHMC (Fig. \ref{fig:ex8}.b) and SGLD (Fig. \ref{fig:ex8}.c) maintain a high confidence in this digits.
	
In this example, D-SGHMC does not significantly improve the classification results (Fig. \ref{fig:ex8}.d and \ref{fig:ex8}.f). However, it does improve the uncertainty estimates, generating higher confidence in the true label (Fig. \ref{fig:ex8}.e).

\clearpage
\subsection{Age Recognition on ADIENCE}
	
\begin{figure}
	\centering
	\begin{minipage}[t]{0.3\linewidth}
		\centering
		\includegraphics[width=\textwidth]{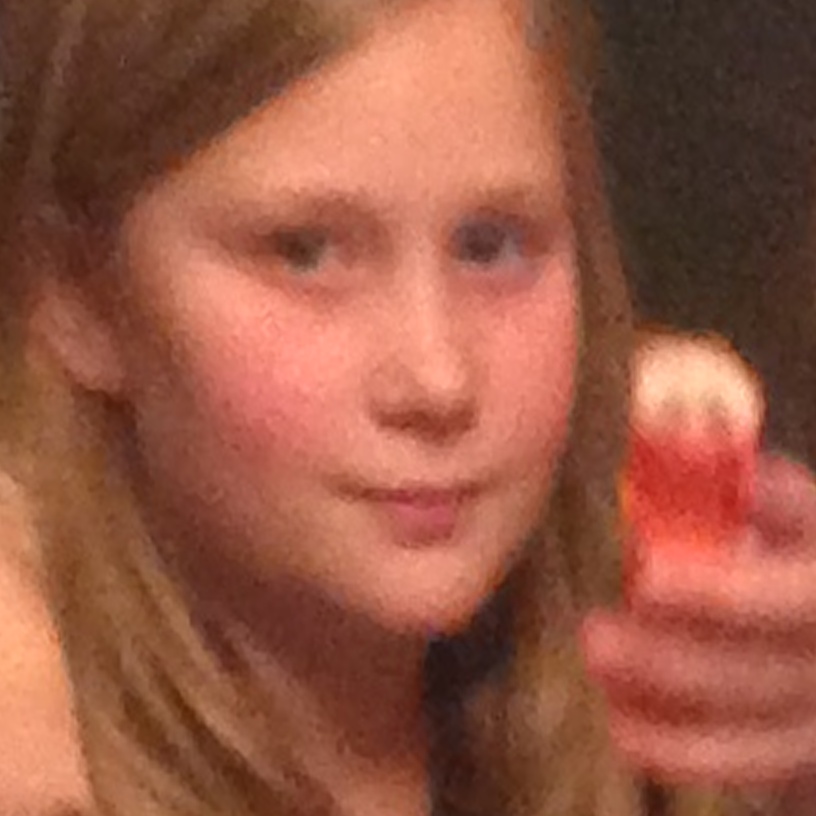}
	\end{minipage}
	\begin{minipage}[t]{0.3\linewidth}
		\centering
		\includegraphics[width=\textwidth]{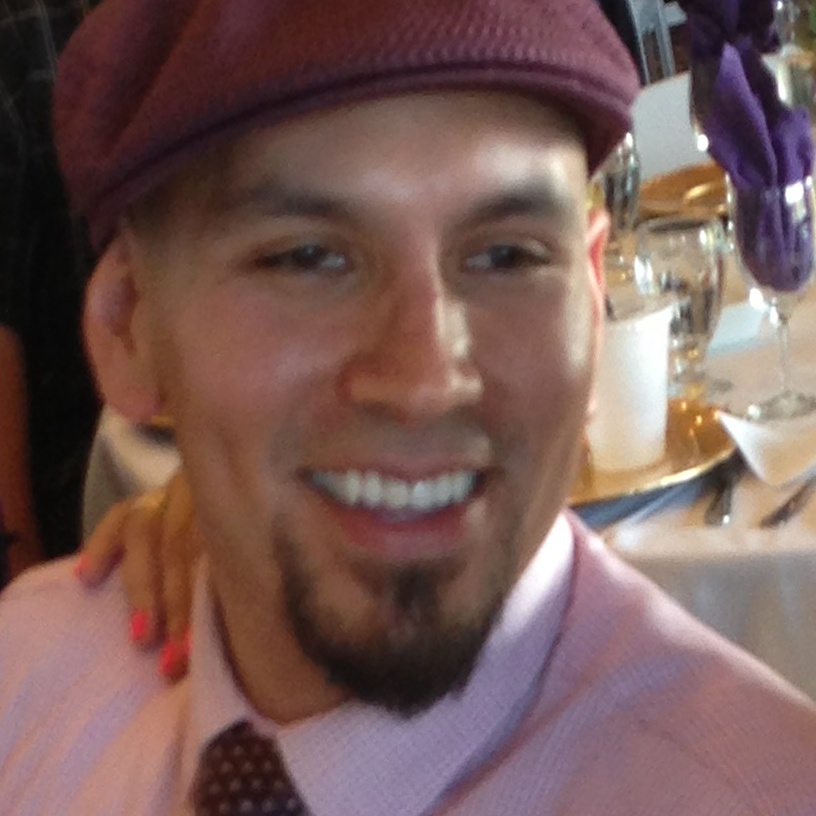}
	\end{minipage}
	\begin{minipage}[t]{0.3\linewidth}
		\centering
		\includegraphics[width=\textwidth]{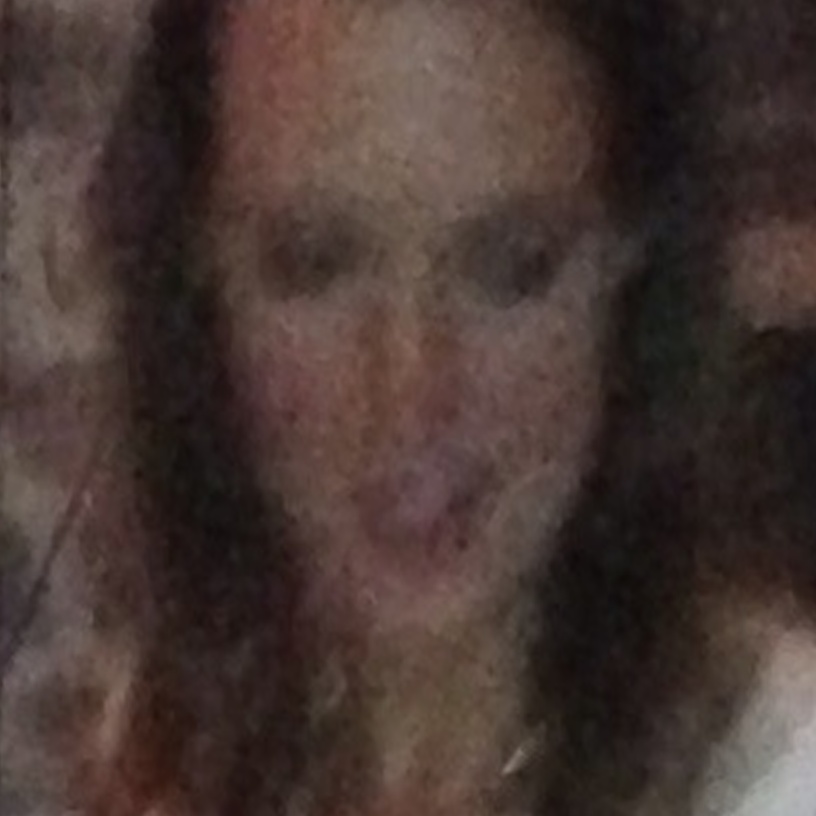}
	\end{minipage}

    \vspace{0.1cm}

	\begin{minipage}[t]{0.3\linewidth}
		\centering
		\includegraphics[width=\textwidth]{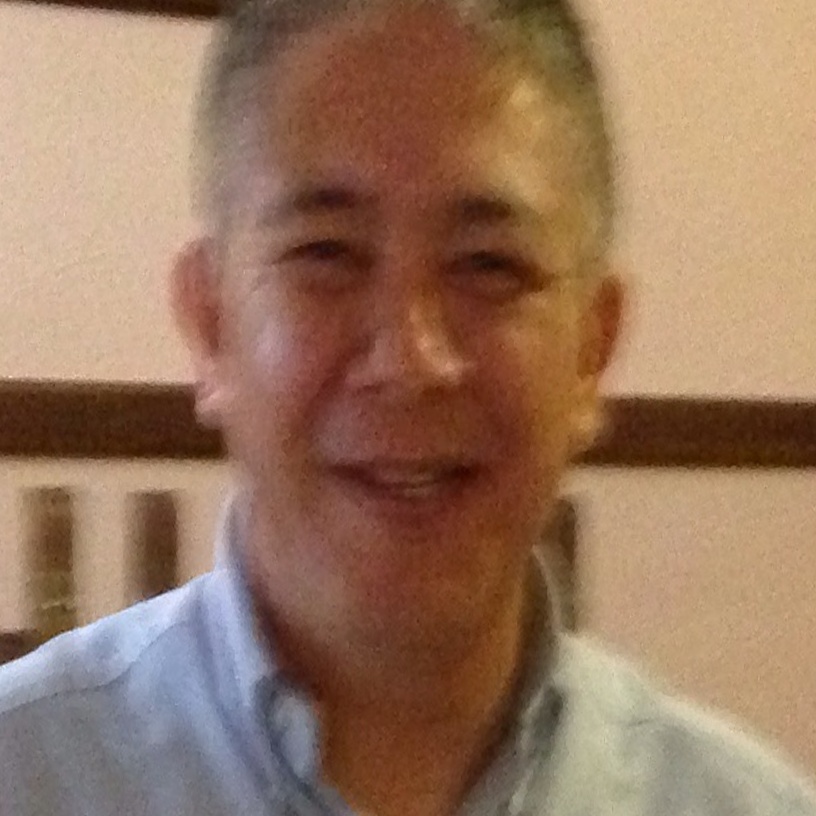}
	\end{minipage}
	\begin{minipage}[t]{0.3\linewidth}
		\centering
		\includegraphics[width=\textwidth]{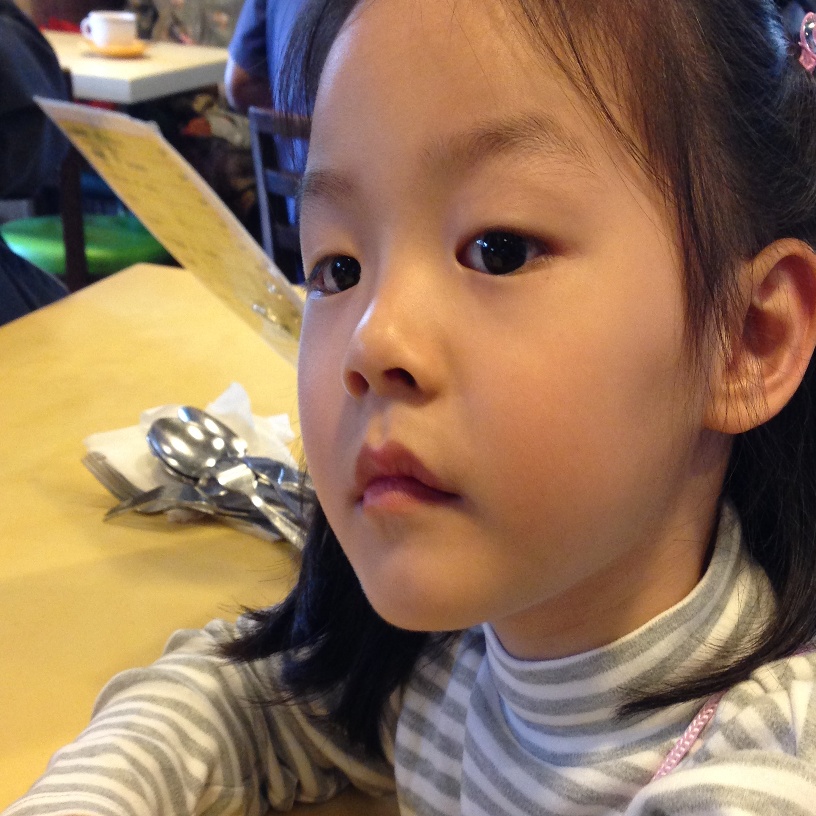}
	\end{minipage}
	\begin{minipage}[t]{0.3\linewidth}
		\centering
		\includegraphics[width=\textwidth]{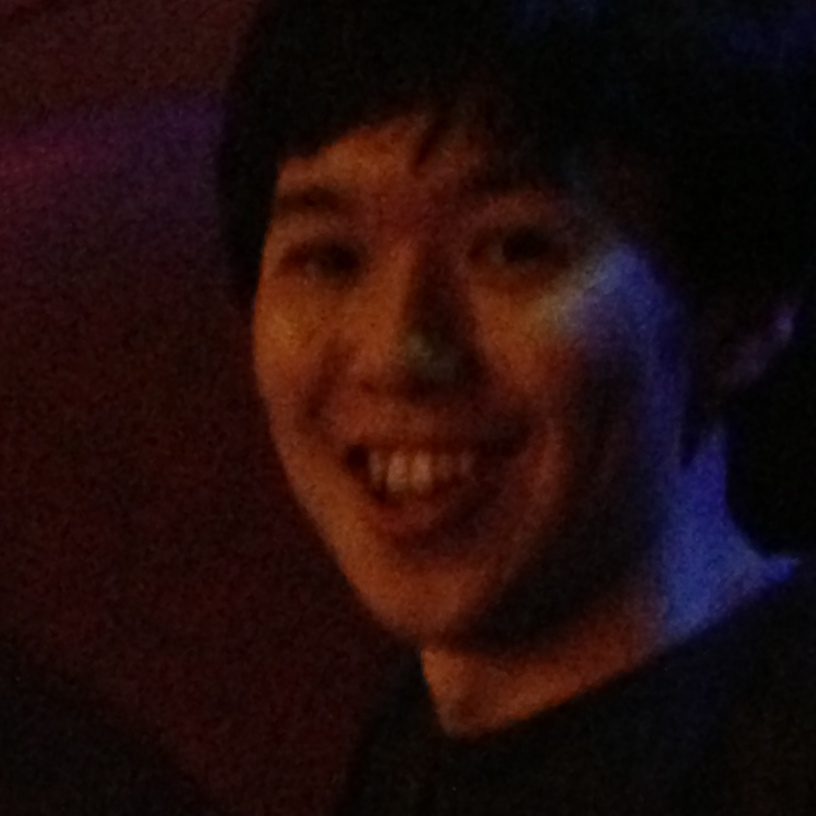}
	\end{minipage}
	\caption{Examples from ADIENCE database.}
\label{fig:adience_image_examples}
\end{figure}

	As part of the many challenges of facial recognition systems, age recognition has been recognized as difficult problem \cite{eidinger2014age}. The ADIENCE dataset (see Fig. \ref{fig:adience_image_examples}) contains $26.580$ images of $2.284$ different subjects and has been used to study the performance of age and gender recognition systems. After pre-processing and cleaning, the resulting data set is partitioned into $13.000$ training examples and $3.534$ test examples. The final number of samples can be seen in Table \ref{tab:preproc}. 	
	
		\begin{table}[h!]
		\centering
			\begin{tabular}{|l|c|l|l|l|} \hline
				\textbf{Age}   & \textbf{ID Class} & \textbf{Train}  & \textbf{Test}  & \textbf{Total}  \\ \hline
				{[}$0-2${]}   & $0$  & $1.201$  & $199$   & $1.400$  \\
				{[}$4-6${]}   & $1$  & $1.566$  & $573$   & $2.139$  \\
				{[}$8-13${]}  & $2$  & $1.942$  & $343$   & $2.285$  \\
				{[}$15-20${]} & $3$ & $1.385$  & $255$   & $1.640$  \\
				{[}$25-32${]} & $4$  & $3.940$  & $1.099$ & $5.039$  \\
				{[}$38-43${]} & $5$  & $1.794$  & $546$   & $2.340$  \\
				{[}$48-53${]} & $6$  & $579$    & $246$   & $825$    \\
				{[}$60- ${]}  & $7$  & $593$    & $273$   & $866$    \\ \hline
				\textbf{Total}       & $8$  & $13.000$ & $3.534$ & $16.534$\\ \hline
		\end{tabular}
		\caption{Final examples distribution of the ADIENCE data set.}
		\label{tab:preproc}
	\end{table}

	Transfer learning from VGG-Face CNN (Convolutional Neural Networks)\cite{Parkhi15} with AVG pooling is used as  a convolutional descriptor. For each example of ADIENCE database, VGG-Face computes a descriptor of $512$ features.
	
	A total number of $5$ independent runs is used for each one of the methods. Using the hyper-parameter setting set in table \ref{tab:hyper-setting}, where state-of-art results are achieved \cite{levi2015age}. Comparison with baseline methods can be seen in Table \ref{tab:adience}. 
	
		\begin{table}[h!]
		\centering
		\begin{tabular}{|l|c|}\hline
			Method& Total Accuracy ($\%$) \\\hline
			SGHMC	&	$45.6 \pm 1.23$\\
			SGLD&	$44.0 \pm 1.10$ \\
			D-SGHMC($p = 0.9)$	& $48.2 \pm 0.56$\\
			D-SGHMC($p = 0.5$)&	$51.6 \pm 0.51$\\
			D-SGHMC($p = 0.1$)&	$48.3 \pm 0.88$\\\hline
		\end{tabular}	
		\caption{Test accuracy for ADIENCE data Set.}
		\label{tab:adience}
	\end{table}

\begin{figure}
	\centering
	\includegraphics[ width=1\linewidth]{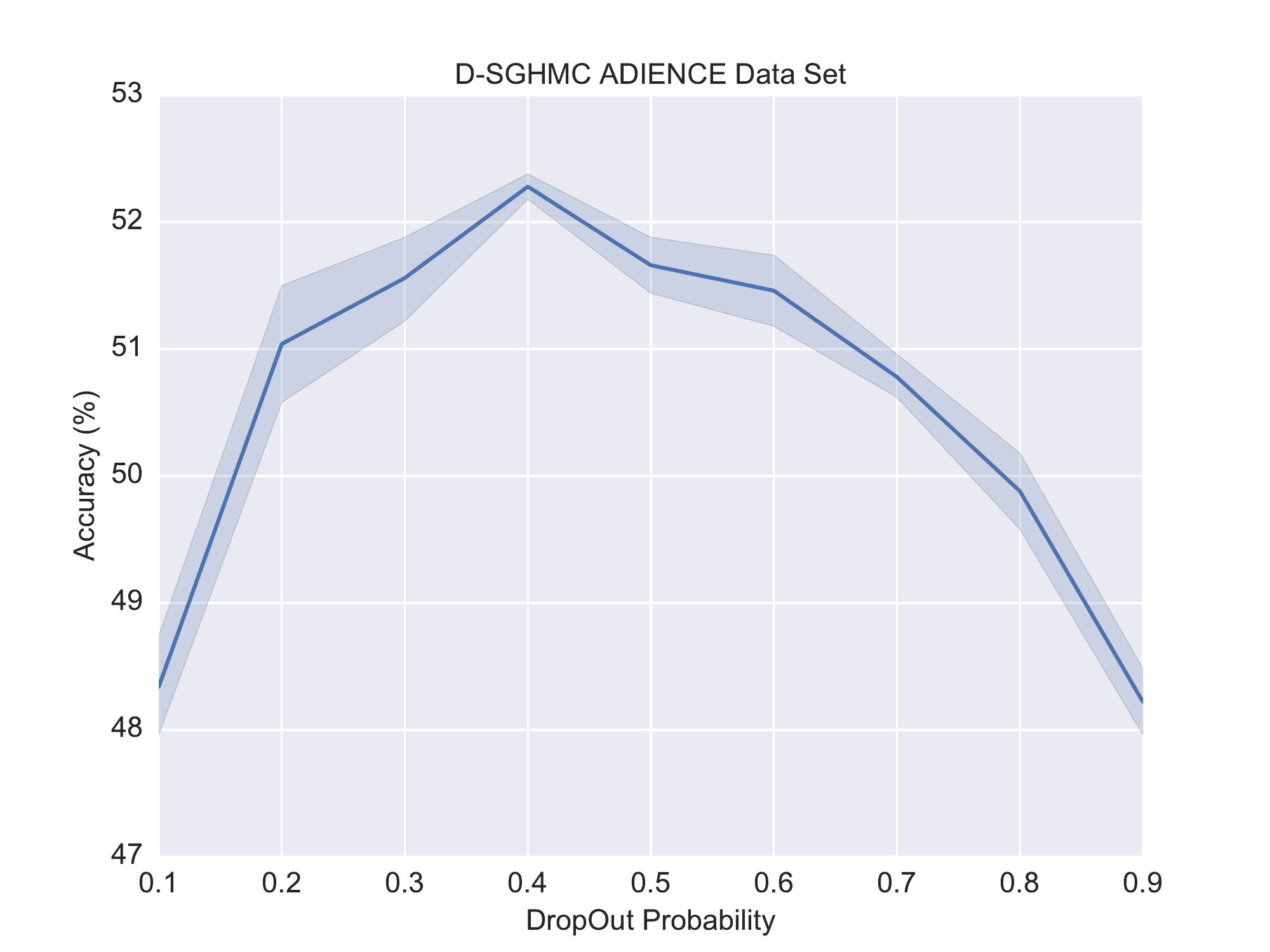}
	\caption{Sensitivity analysis for D-SGHMC with Dropout rate $p$.}
	\label{fig:sensitivity2}
\end{figure}

 The role of the Dropout rate $p$ on the performance is analyzed. As shown in Figure \ref{fig:sensitivity2}, best results are obtained with values between $0.3$ and $0.5$.

\begin{figure}
	\begin{minipage}[t]{0.46\linewidth}
		\centering
		\includegraphics[width=\linewidth]{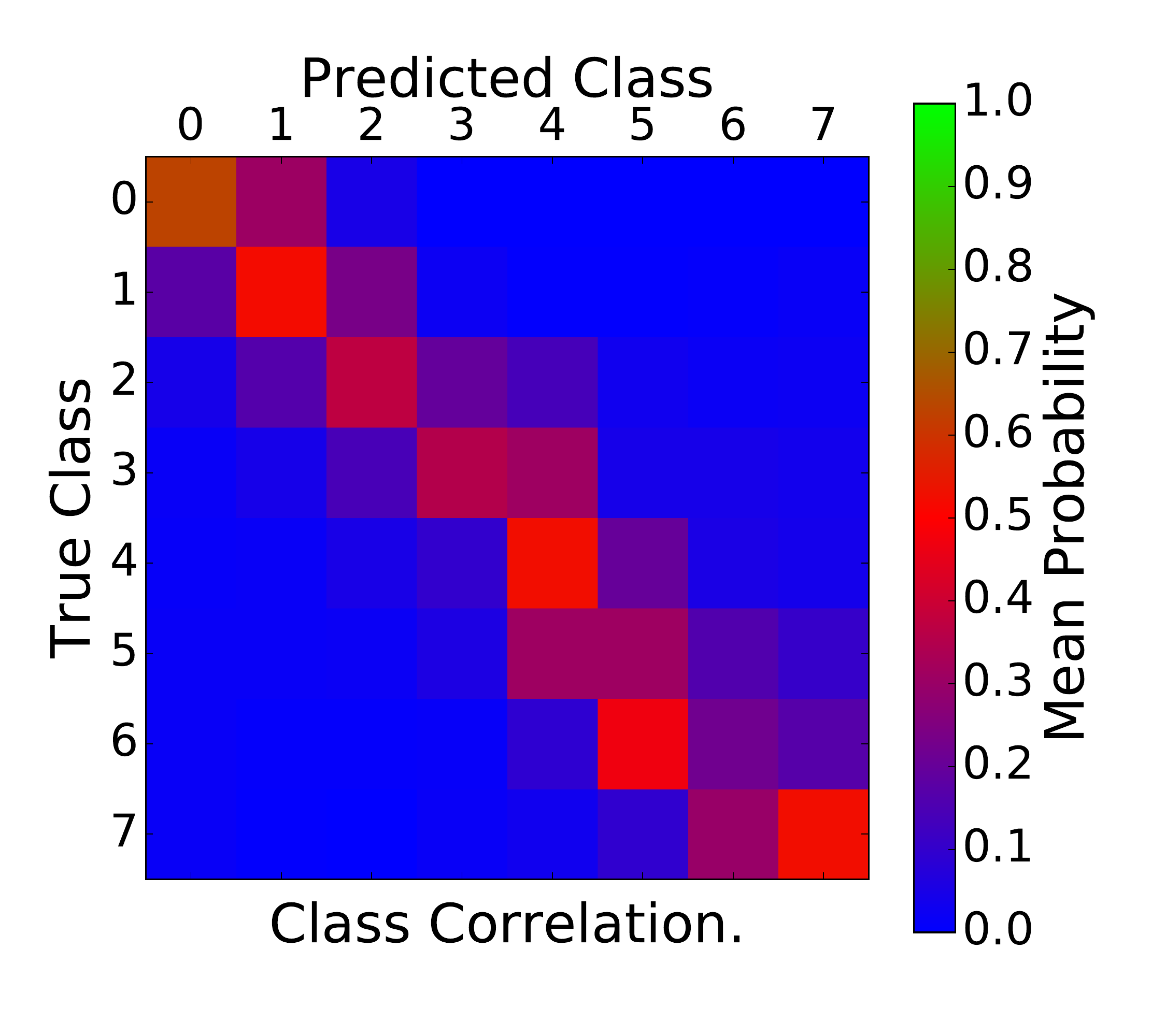}
		(a) SGHMC
	\end{minipage}
	\begin{minipage}[t]{0.46\linewidth}
		\centering
		\includegraphics[width=\linewidth]{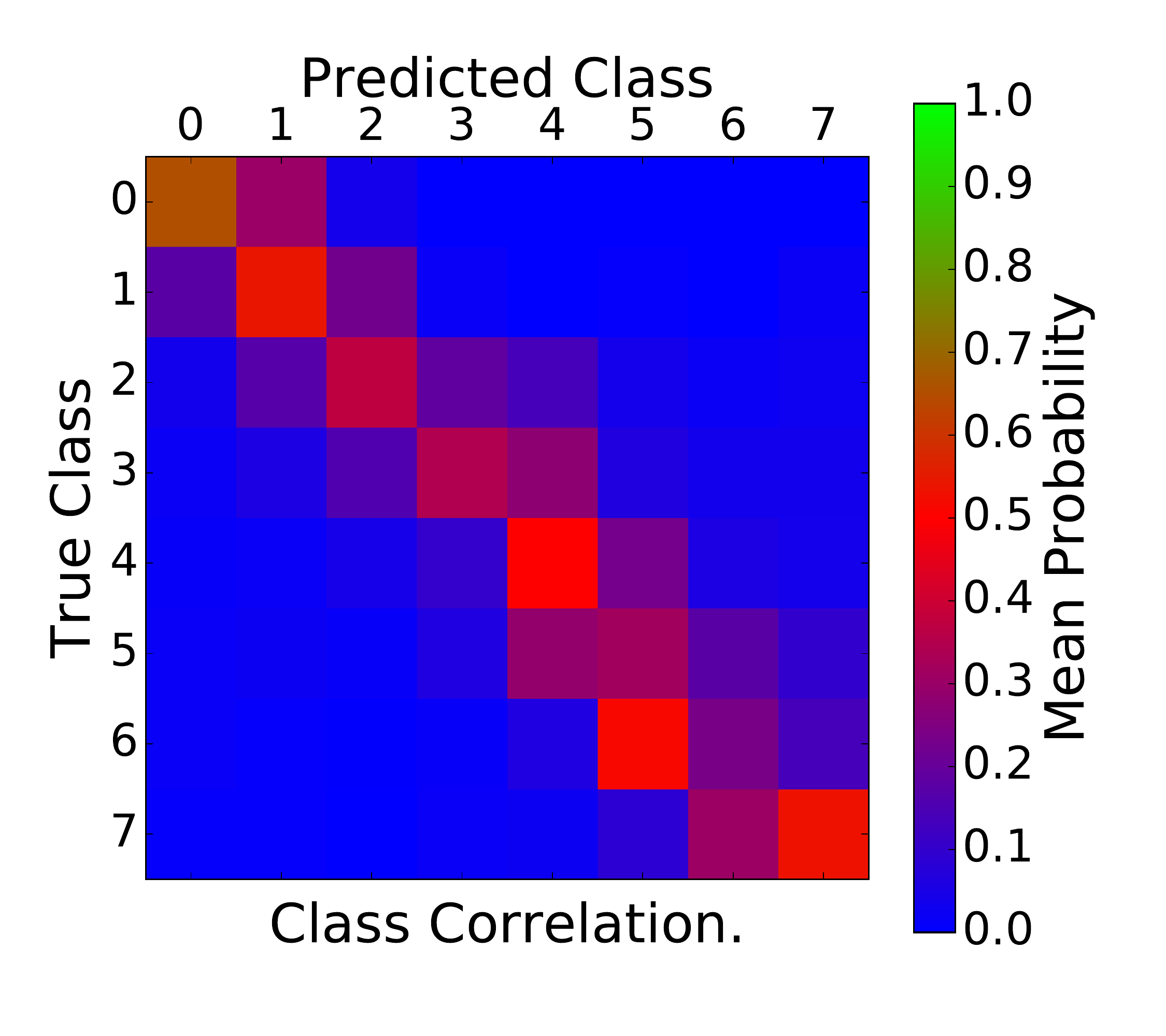}
		(b) SGLD
	\end{minipage}
	\begin{minipage}[t]{0.46\linewidth}
		\centering
		\includegraphics[width=\linewidth]{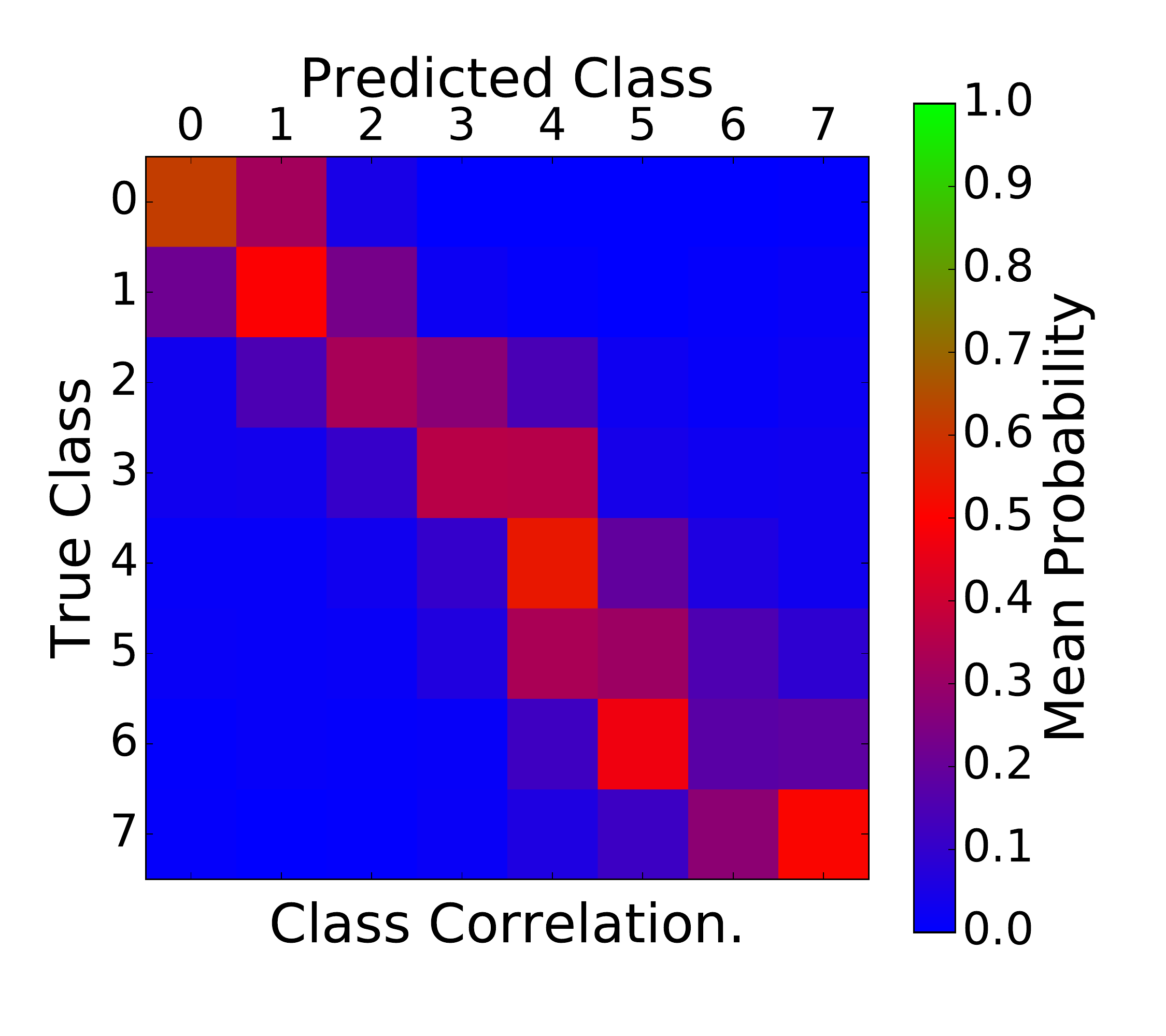}
		(c) D-SGHMC ($p = 0.9$)
	\end{minipage}
	\begin{minipage}[t]{0.46\linewidth}
		\centering
		\includegraphics[width=\linewidth]{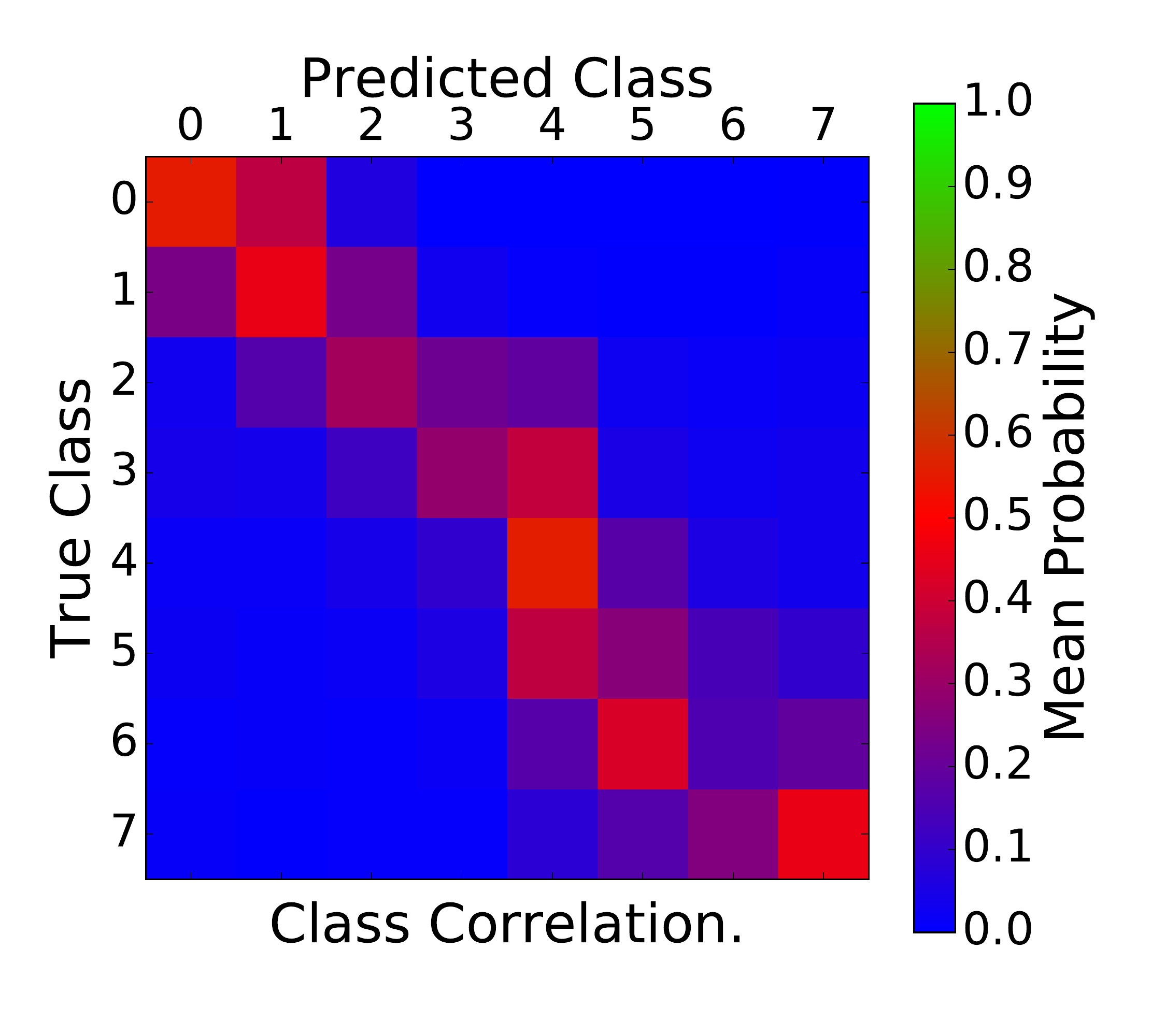}
		(d) D-SGHMC ($p = 0.5$)
	\end{minipage}
	\centering
	\begin{minipage}[t]{0.46\linewidth}
		\centering
		\includegraphics[width=\linewidth]{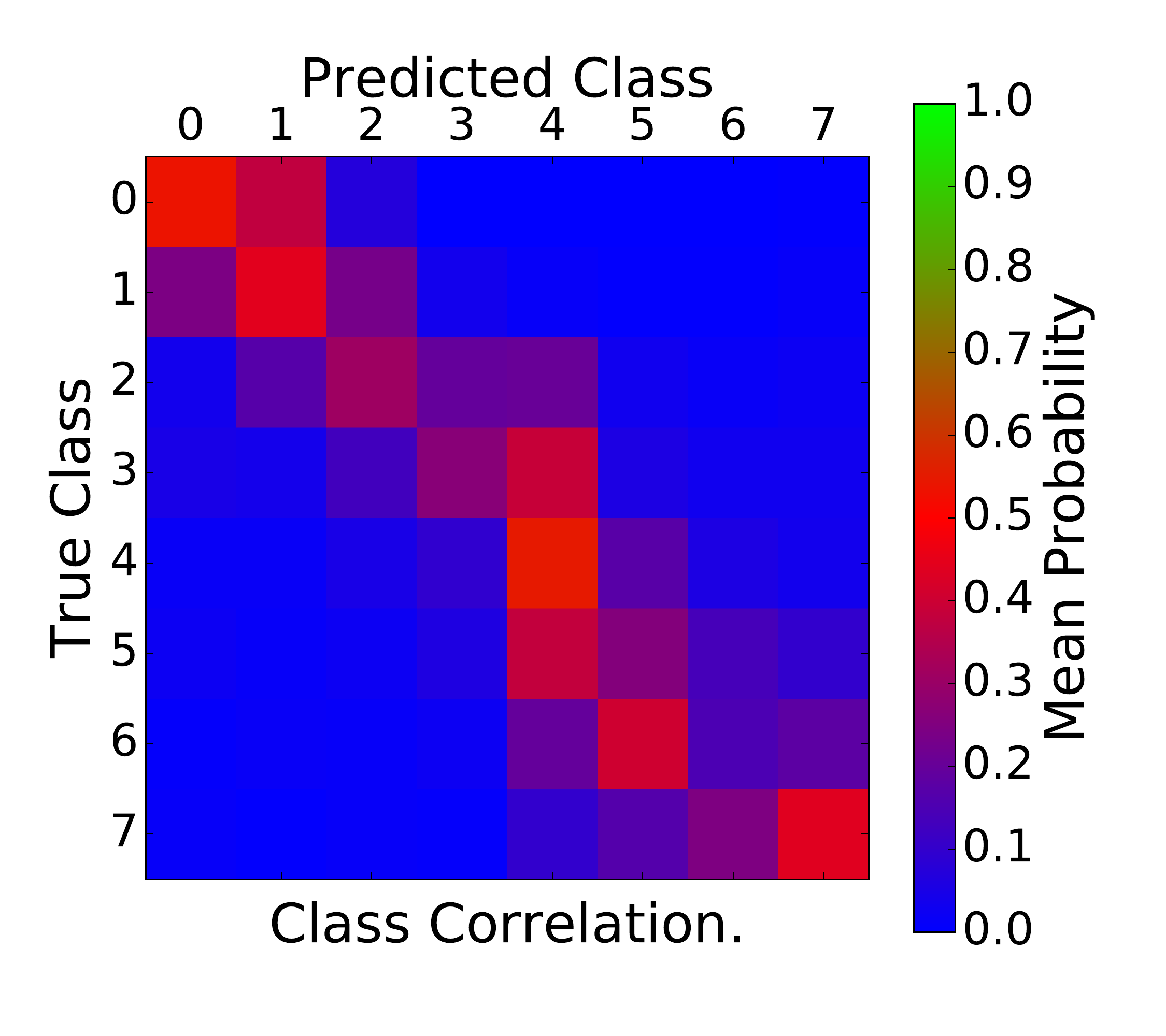}
		(e) D-SGHMC ($p = 0.4$)
	\end{minipage}%
	\begin{minipage}[t]{0.46\linewidth}
		\centering
		\includegraphics[width=\linewidth]{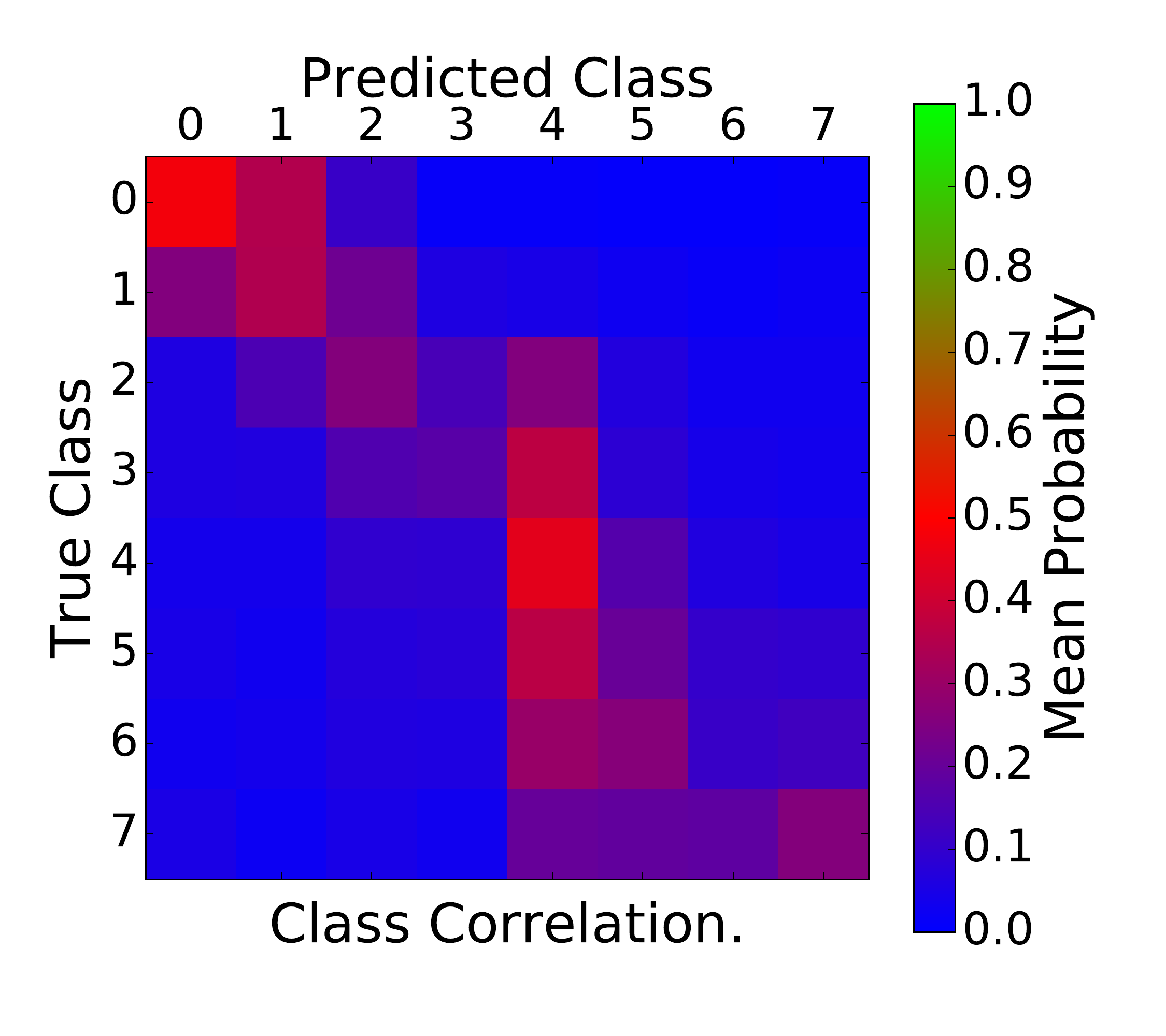}
		(f) D-SGHMC ($p = 0.1$)
	\end{minipage}%
	\caption{Matrix Correlation plot of probabilities for all classes. $5$ independent chains on ADIENCE.}
	\label{fig:adience_prob_class}
\end{figure}

In ADIENCE, uncertainty is increased for neighboring classes. This can be seen in Figure \ref{fig:adience_prob_class}, which shows the true and the predicted labels.

\clearpage
\subsubsection{Accuracy}

\begin{figure}
	\begin{minipage}[t]{0.46\linewidth}
		\centering
		\includegraphics[width=\linewidth]{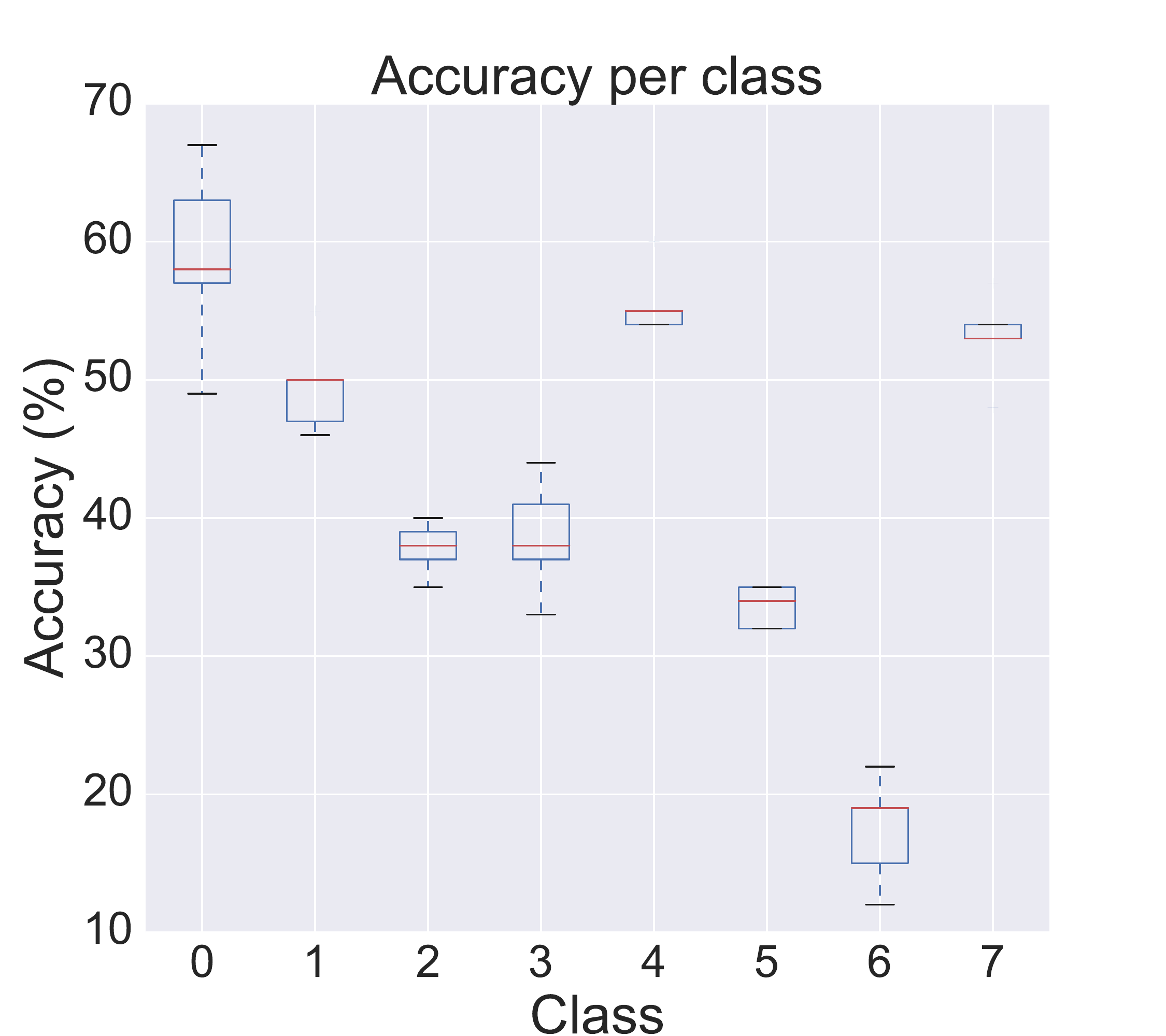}
		(a) SGHMC
	\end{minipage}
	\begin{minipage}[t]{0.46\linewidth}
		\centering
		\includegraphics[width=\linewidth]{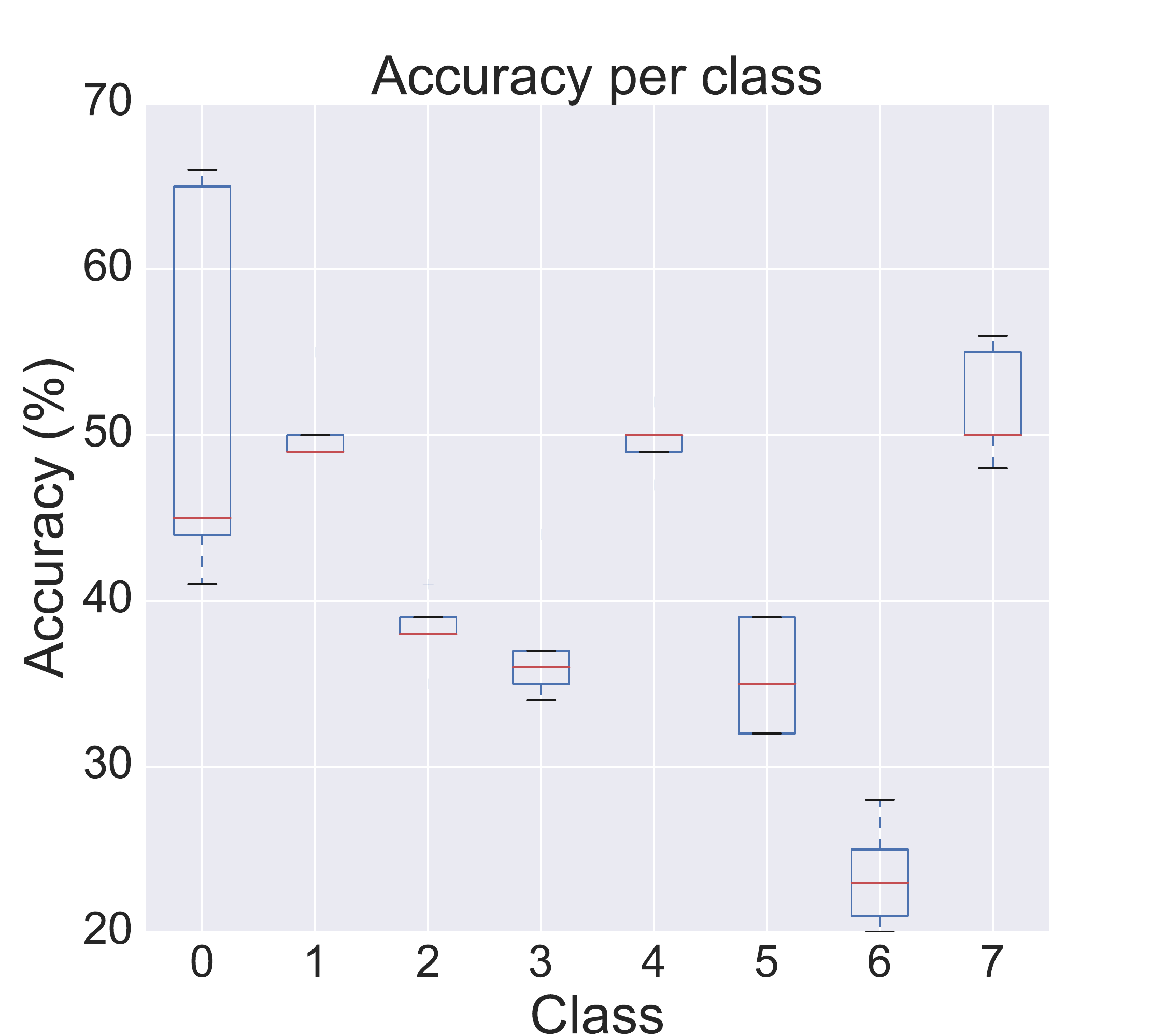}
		(b) SGLD
	\end{minipage}
	\begin{minipage}[t]{0.46\linewidth}
		\centering
		\includegraphics[width=\linewidth]{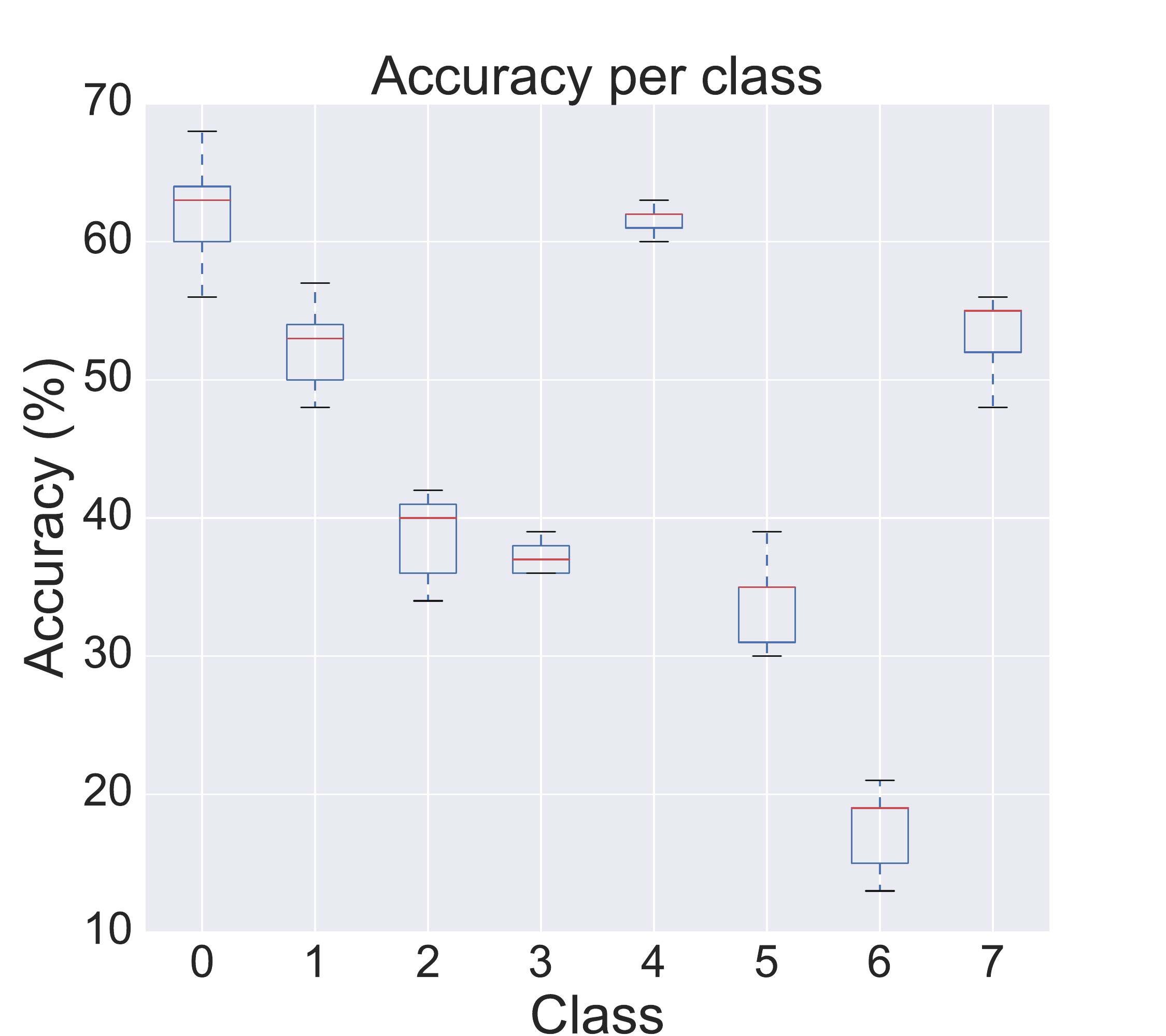}
		(c) D-SGHMC ($p = 0.9$)
	\end{minipage}
	\begin{minipage}[t]{0.46\linewidth}
		\centering
		\includegraphics[width=\linewidth]{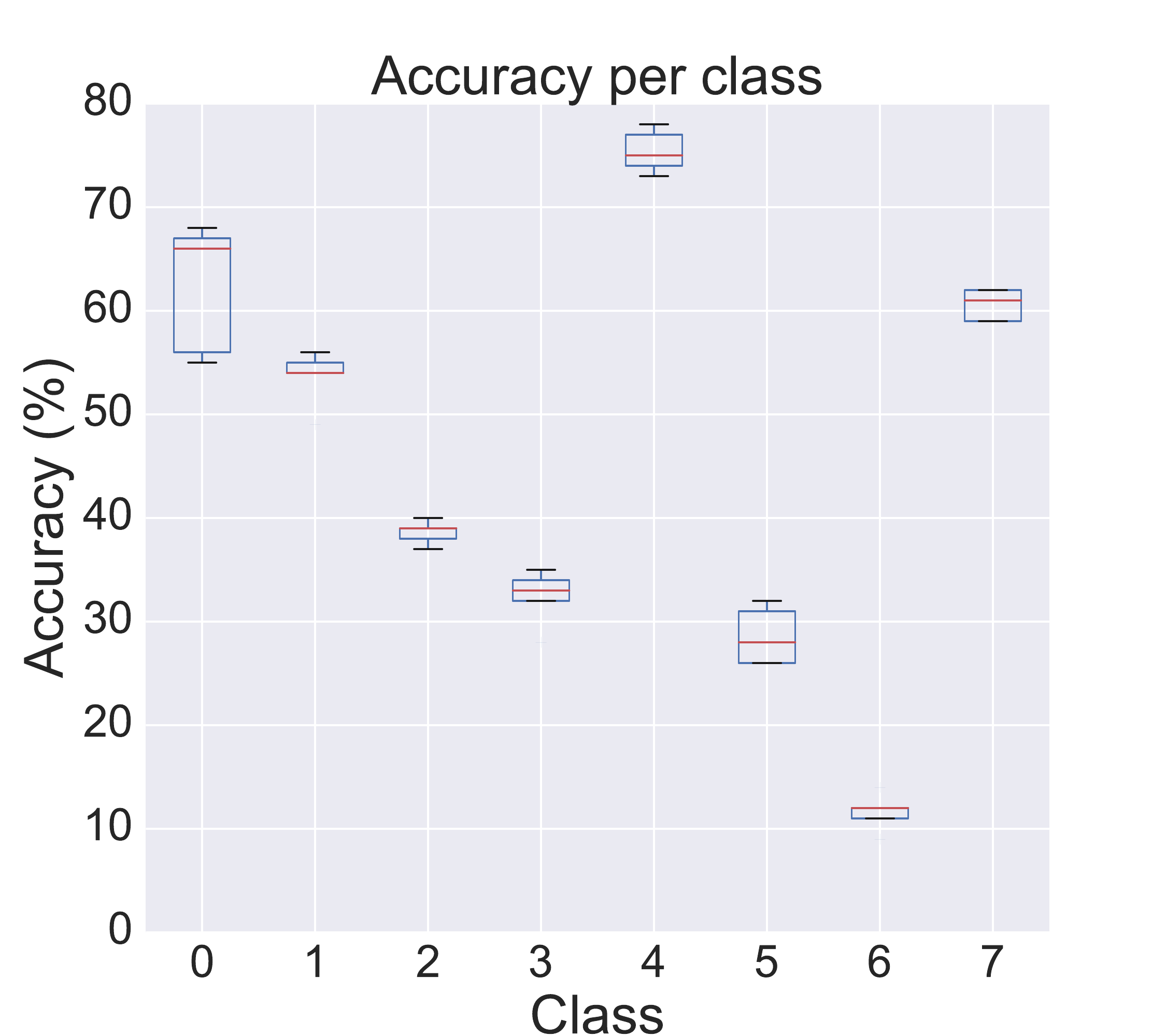}
		(d) D-SGHMC ($p = 0.5$)
	\end{minipage}
	\centering
	\begin{minipage}[t]{0.46\linewidth}
		\centering
		\includegraphics[width=\linewidth]{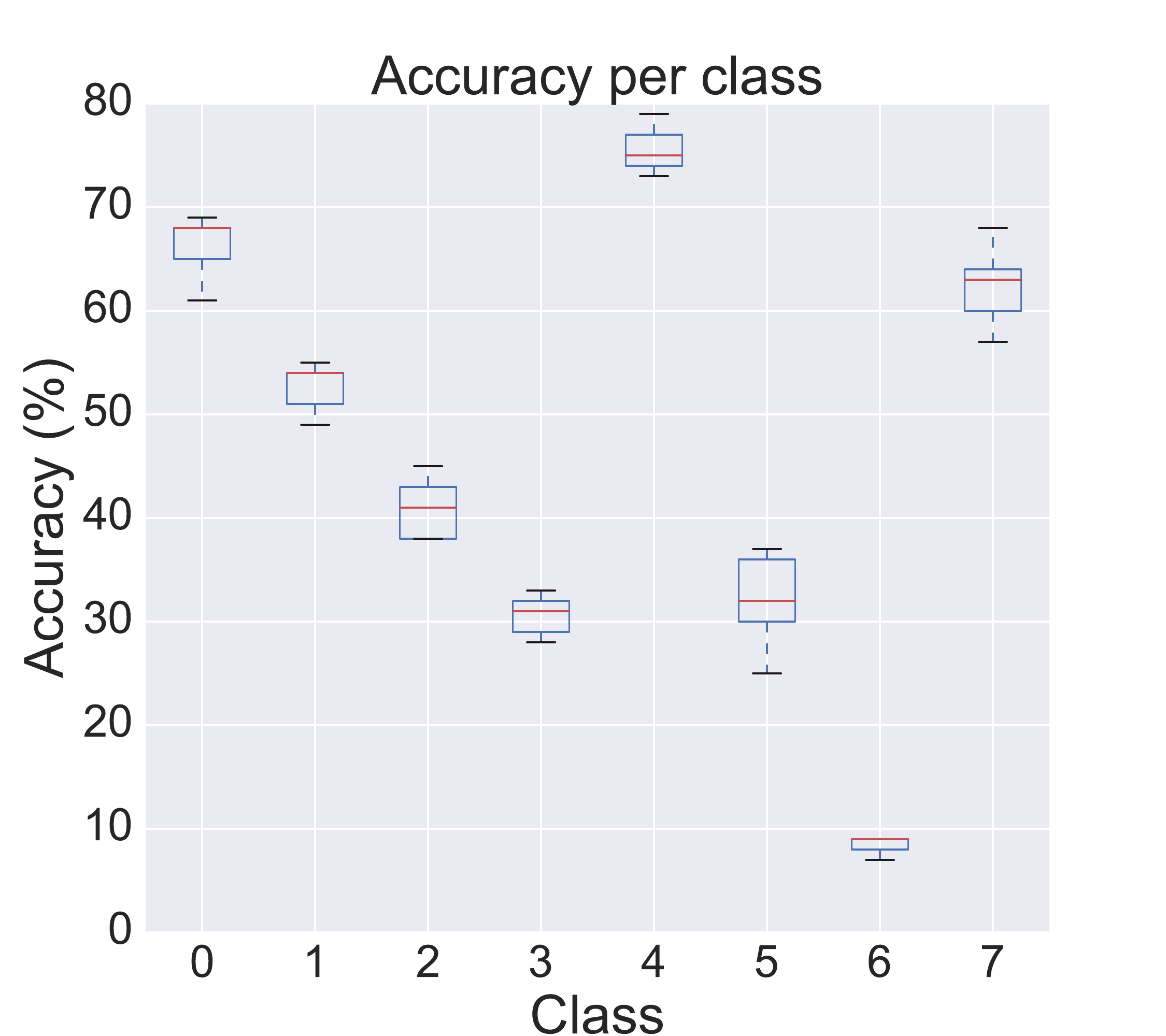}
		(e) D-SGHMC ($p = 0.4$)
	\end{minipage}%
	\begin{minipage}[t]{0.46\linewidth}
		\centering
		\includegraphics[width=\linewidth]{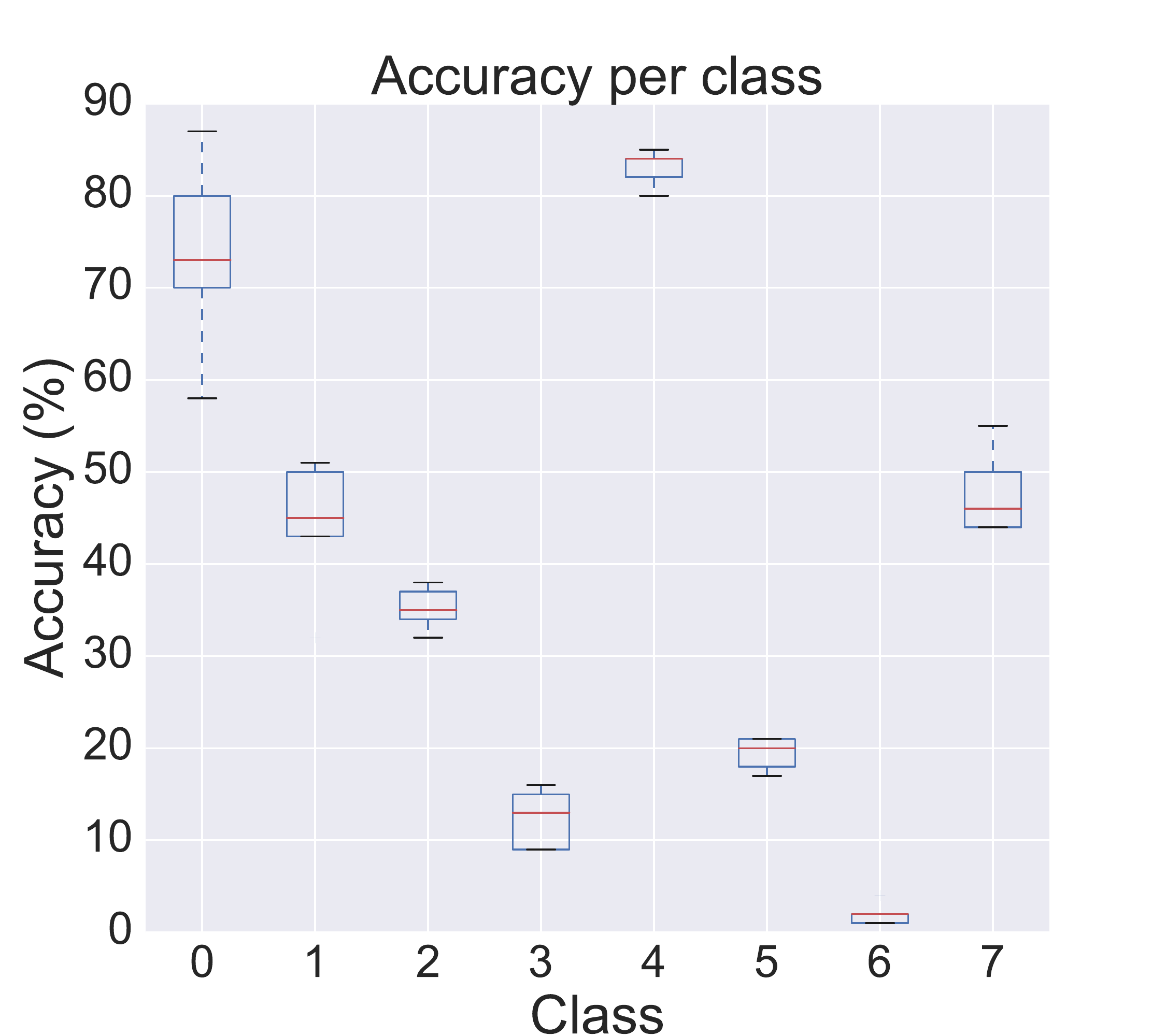}
		(f) D-SGHMC ($p = 0.1$)
	\end{minipage}%
	\caption{Box plot of accuracies for all classes on ADIENCE.}
	\label{fig:prob_class}
\end{figure}

Class imbalance plays an important role in the final classification results for the age recognition problem in ADIENCE. D-SGHMC (see Figure \ref{fig:prob_class}.d) improves results in most of classes, sacrificing to a lesser extent the accuracy for classes with less number of examples.

As the Dropout rate decreases, the imbalance problem becomes more evident and the accuracy results are also dropped (see Figure \ref{fig:sensitivity2} ).

\subsubsection{Predictive Uncertainty for the Adience Dataset}	

\begin{figure*}
    \centering
    \begin{minipage}[t]{\textwidth}
        \begin{minipage}[b]{0.14\textwidth} \centering{Class} \newline\end{minipage}
        \begin{minipage}[b]{0.15\textwidth} \centering{SGHMC} \newline \end{minipage}
        \begin{minipage}[b]{0.15\textwidth} \centering{SGLD} \newline \end{minipage}
        \begin{minipage}[b]{0.15\textwidth} \centering{D-SGHMC ($p = 0.9$)} \end{minipage}
        \begin{minipage}[b]{0.15\textwidth} \centering{D-SGHMC ($p = 0.5$)} \end{minipage}
        \begin{minipage}[b]{0.15\textwidth} \centering{D-SGHMC ($p = 0.1$)} \end{minipage}
    \end{minipage}

    \begin{minipage}[t]{\textwidth}
        \begin{minipage}[b]{0.14\textwidth}
            \centering{Class 0}\\ \includegraphics[width=\textwidth]{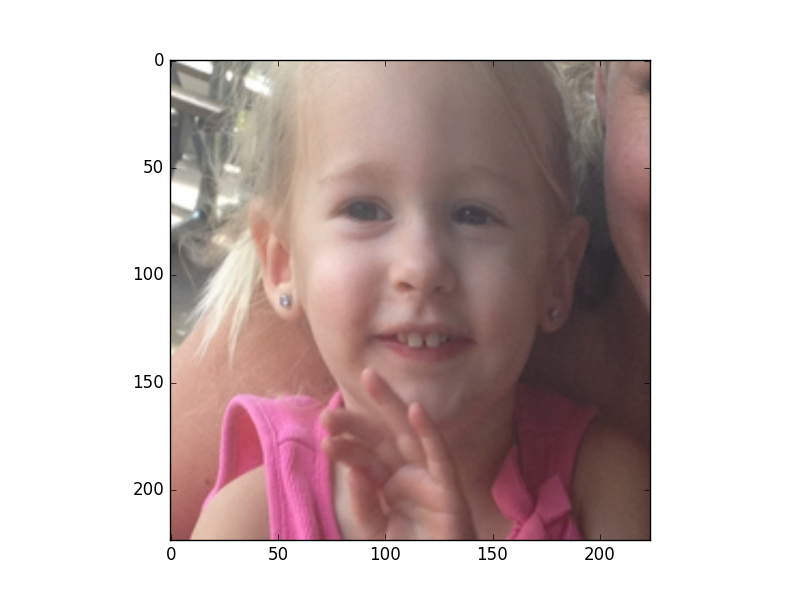}
        \end{minipage}        
        \includegraphics[width=0.15\textwidth]{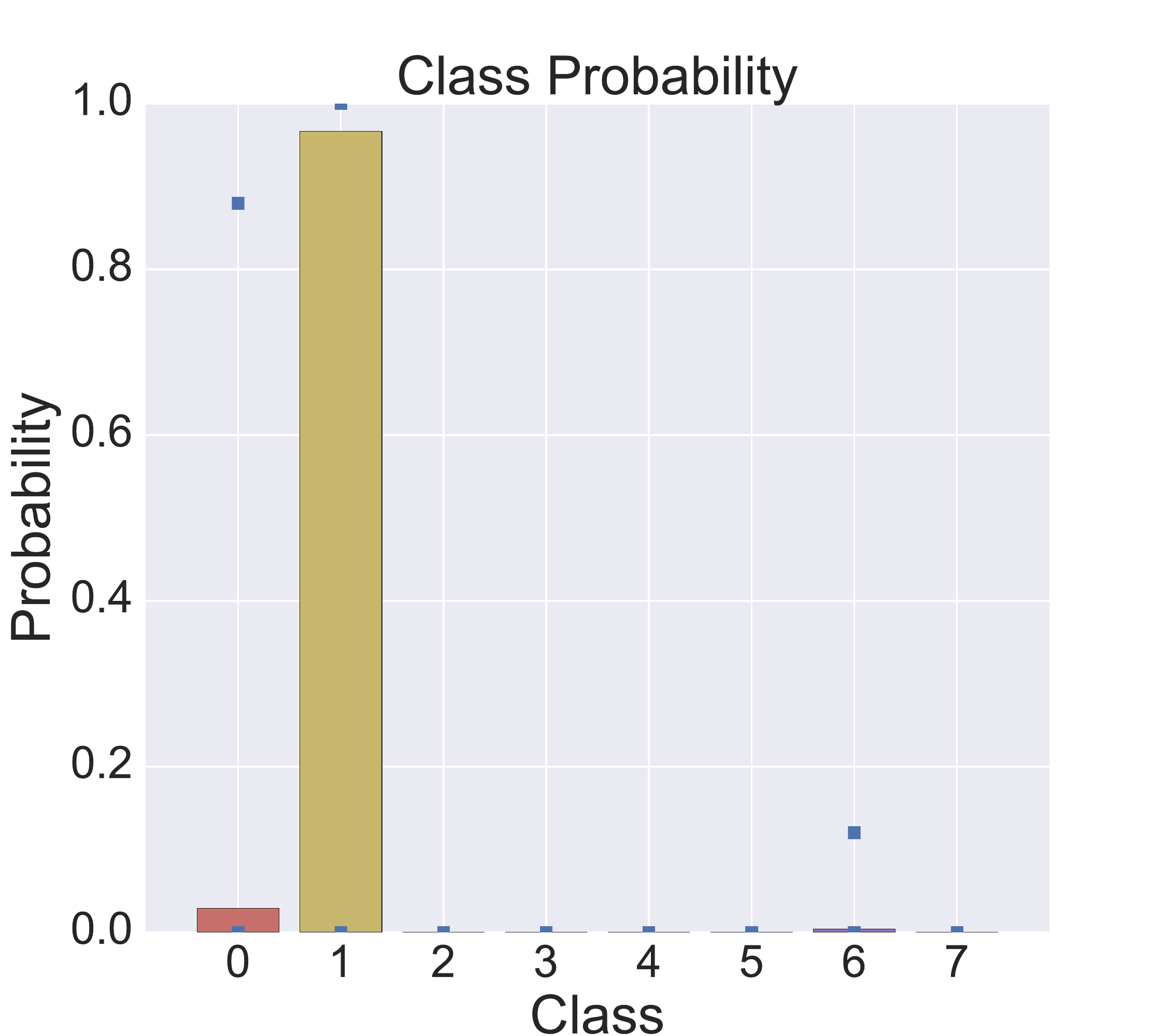}
        \includegraphics[width=0.15\textwidth]{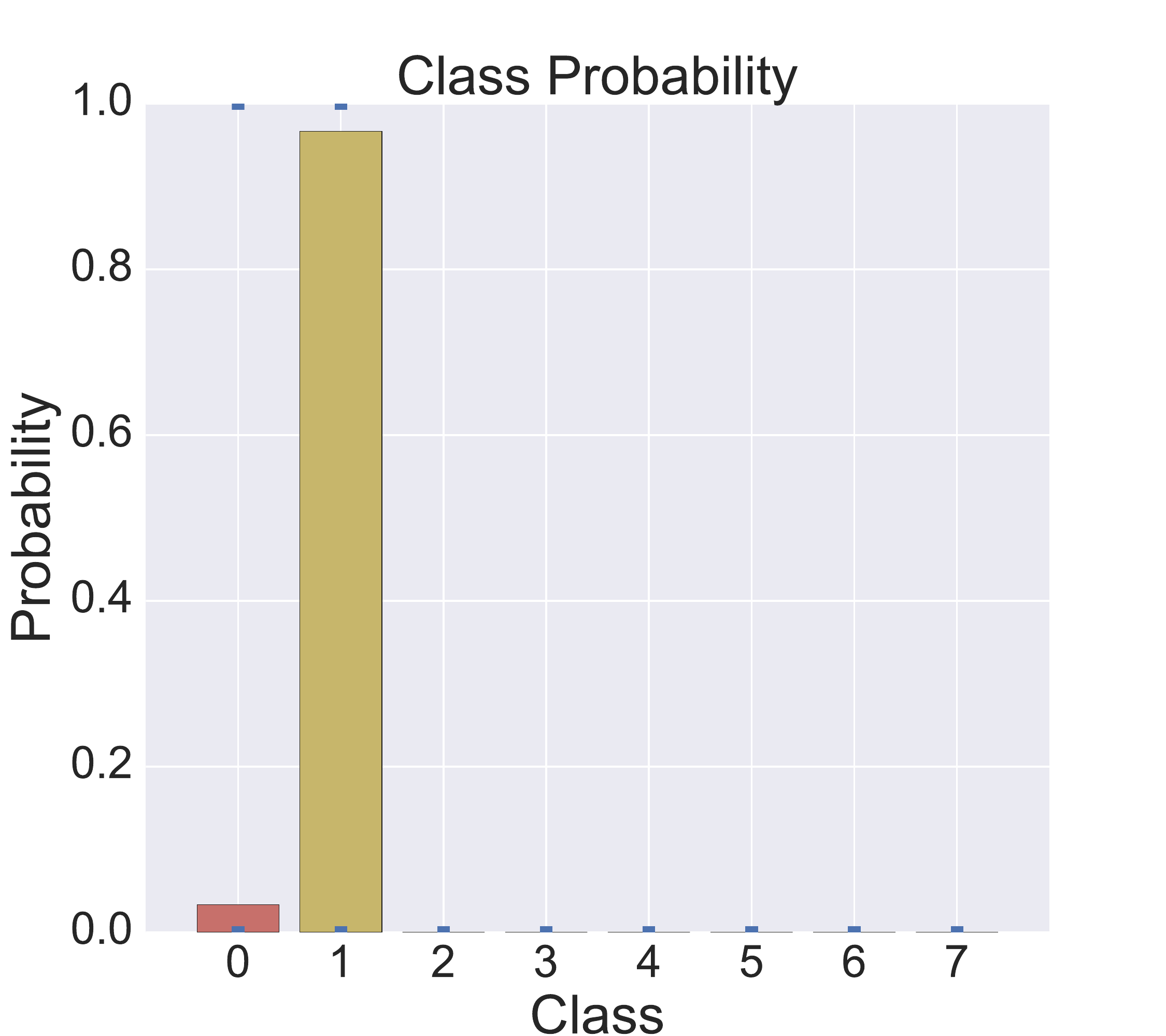}
        \includegraphics[width=0.15\textwidth]{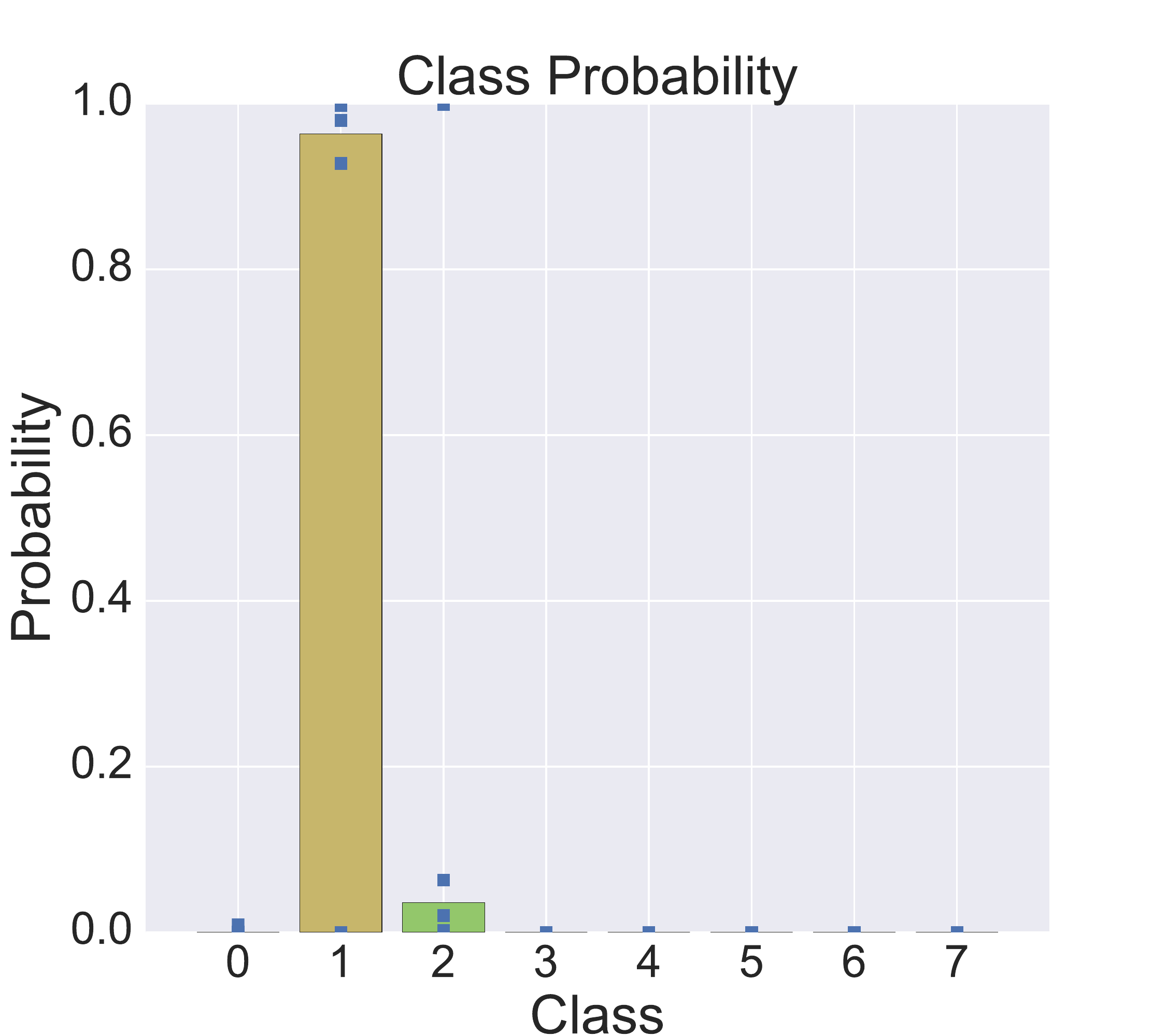}
        \includegraphics[width=0.15\textwidth]{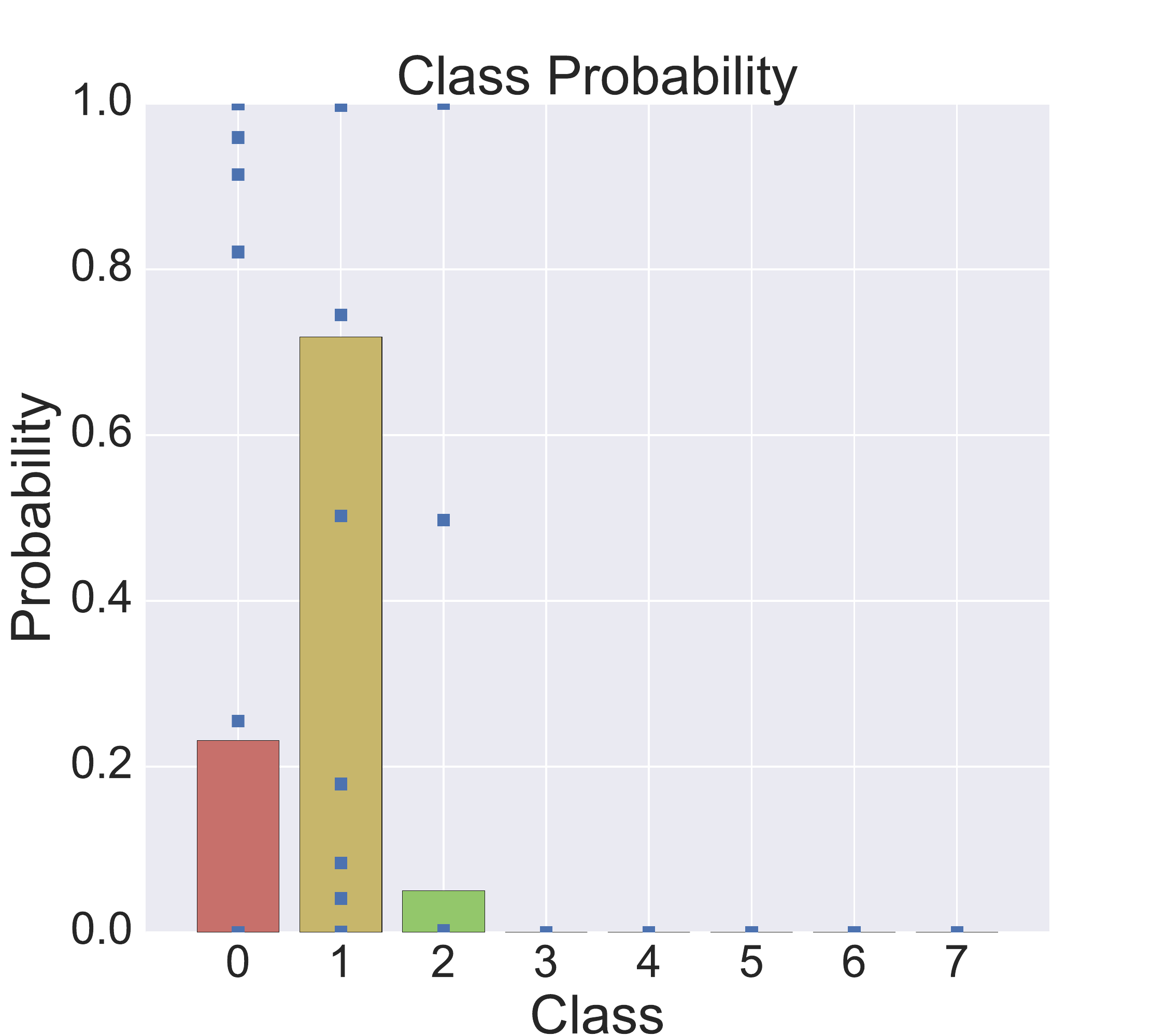}
        \includegraphics[width=0.15\textwidth]{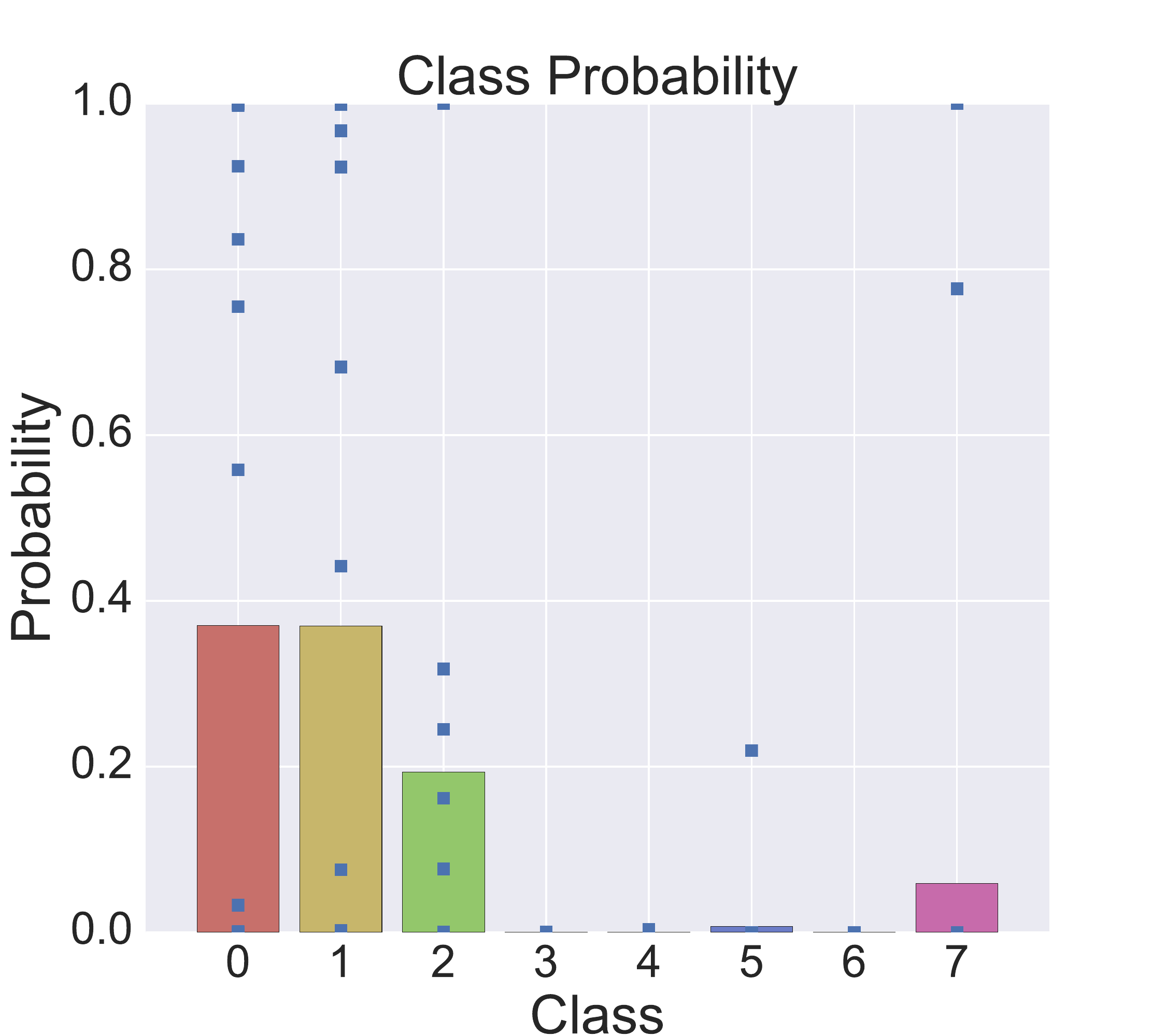}
    \end{minipage}

    \begin{minipage}[t]{\textwidth}
        \begin{minipage}[b]{0.14\textwidth}
            \centering{Class 1}\\ \includegraphics[width=\textwidth]{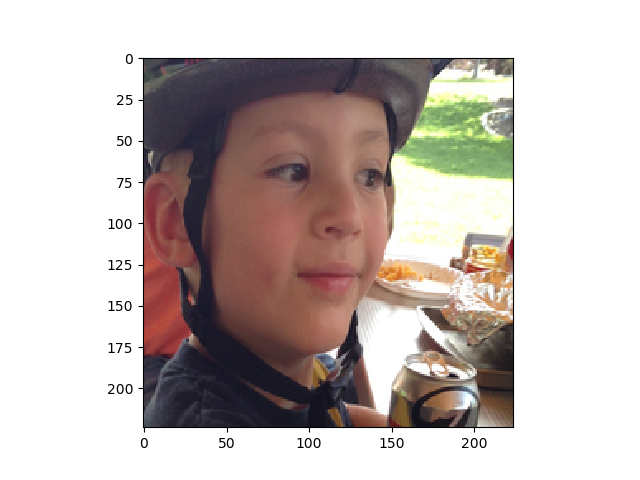}
        \end{minipage}        
        \includegraphics[width=0.15\textwidth]{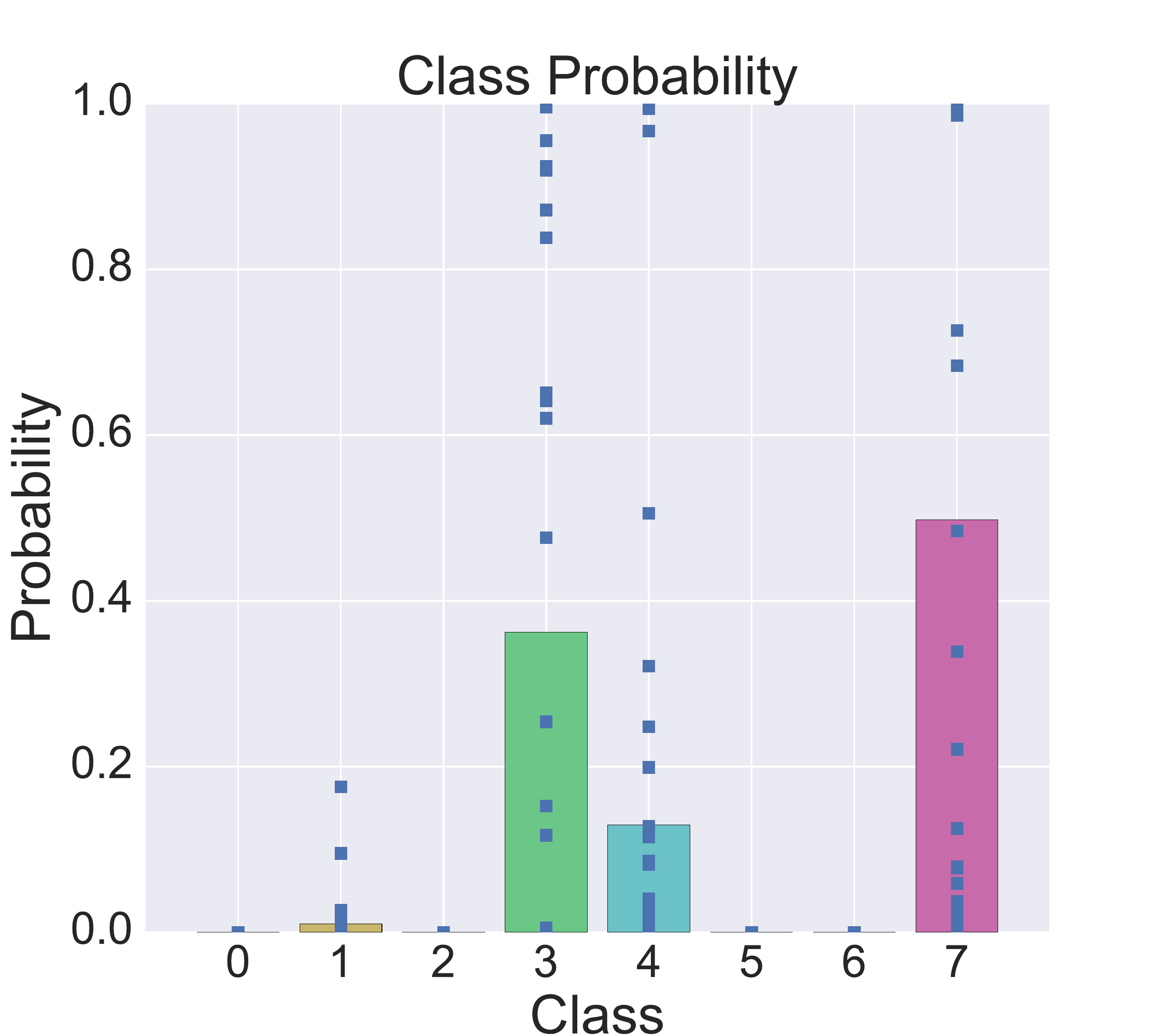}
        \includegraphics[width=0.15\textwidth]{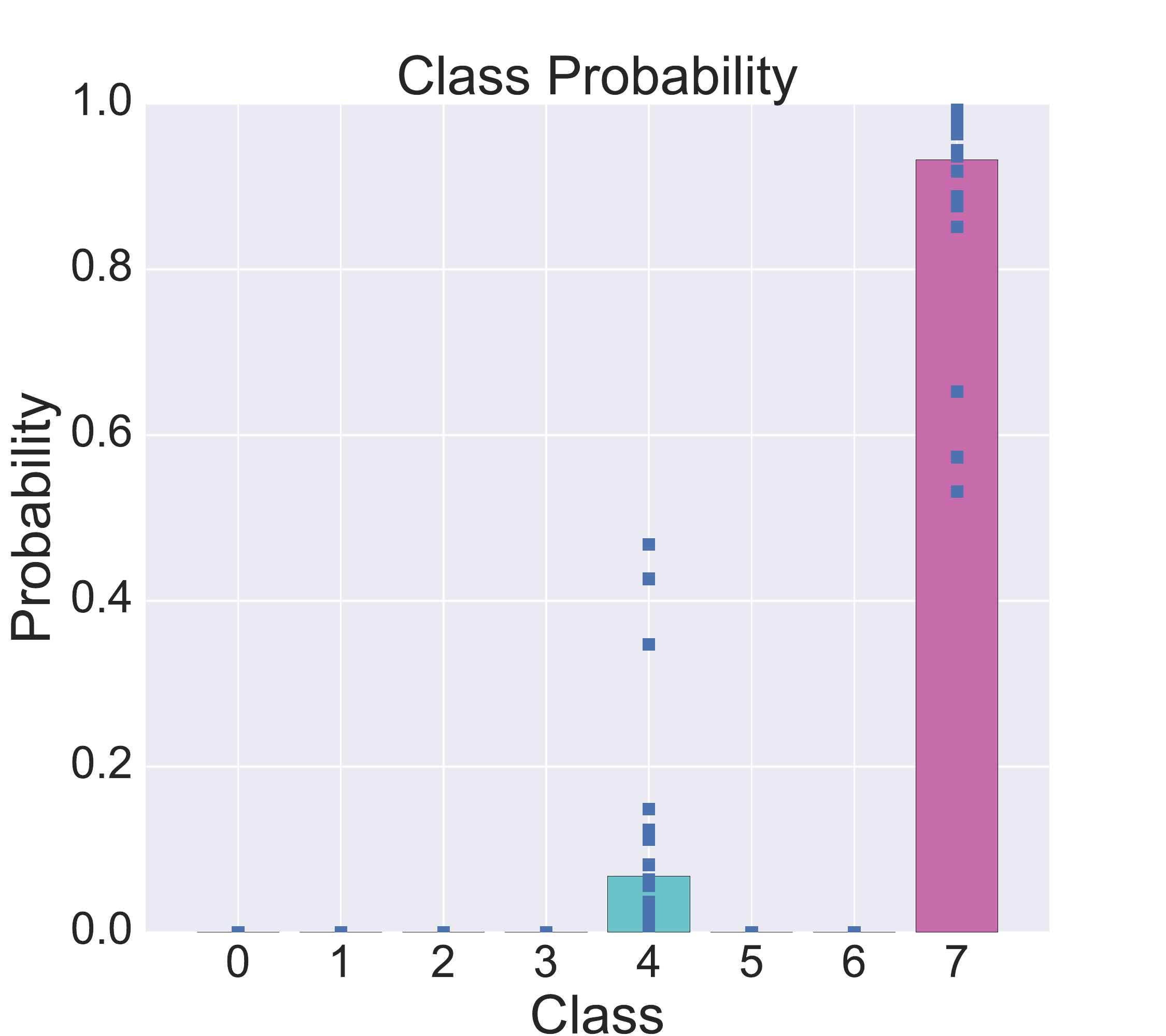}
        \includegraphics[width=0.15\textwidth]{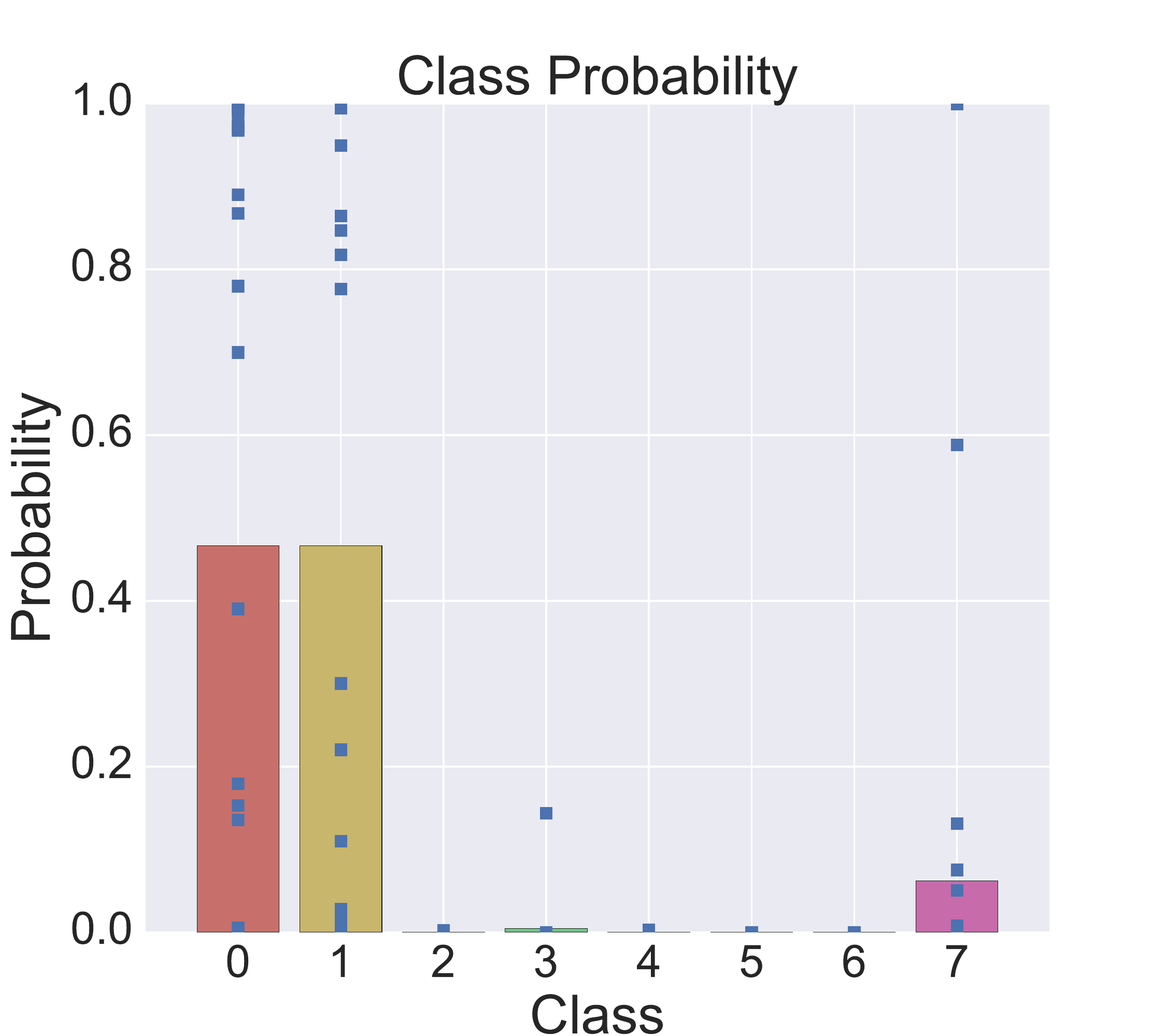}
        \includegraphics[width=0.15\textwidth]{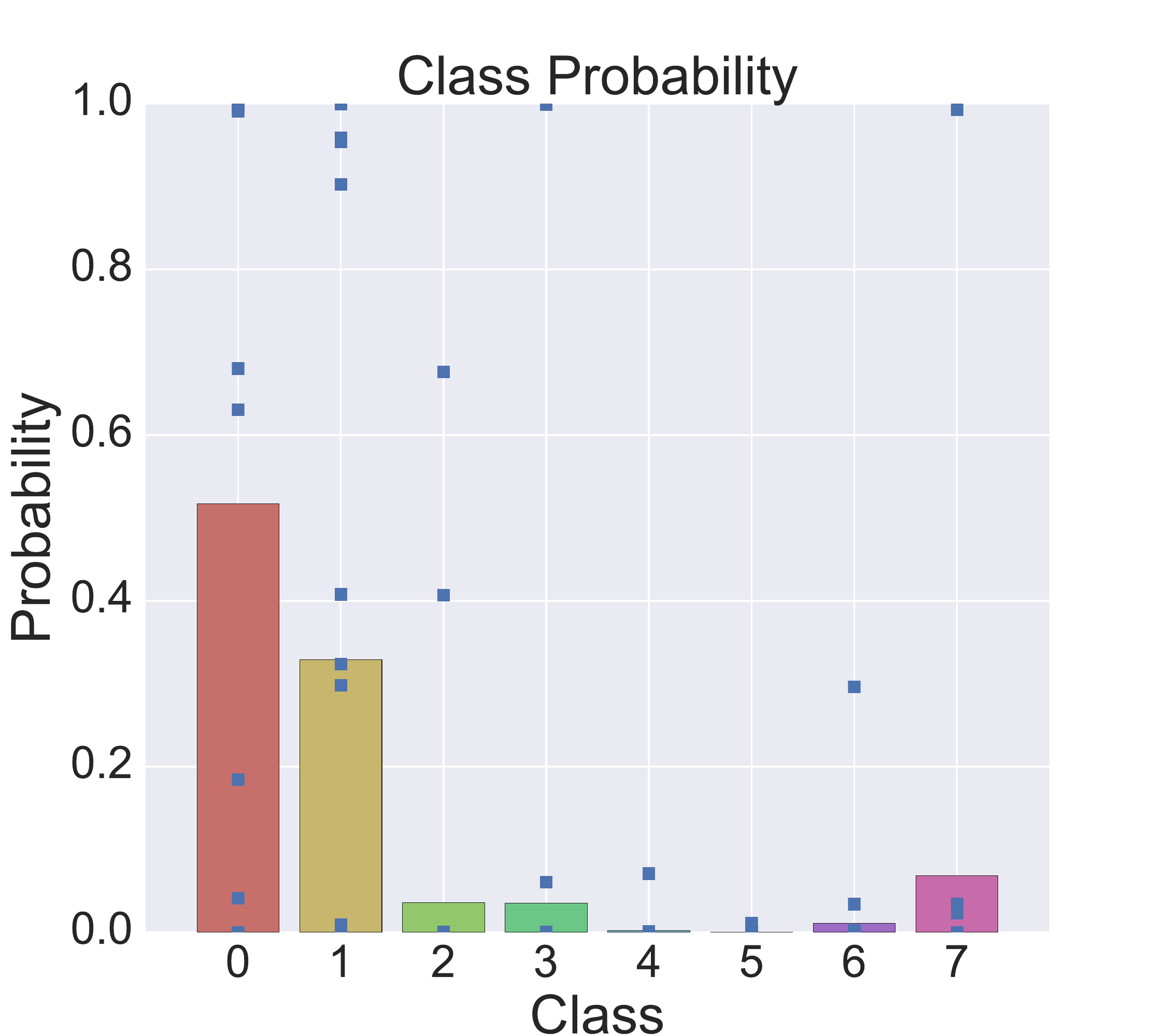}
        \includegraphics[width=0.15\textwidth]{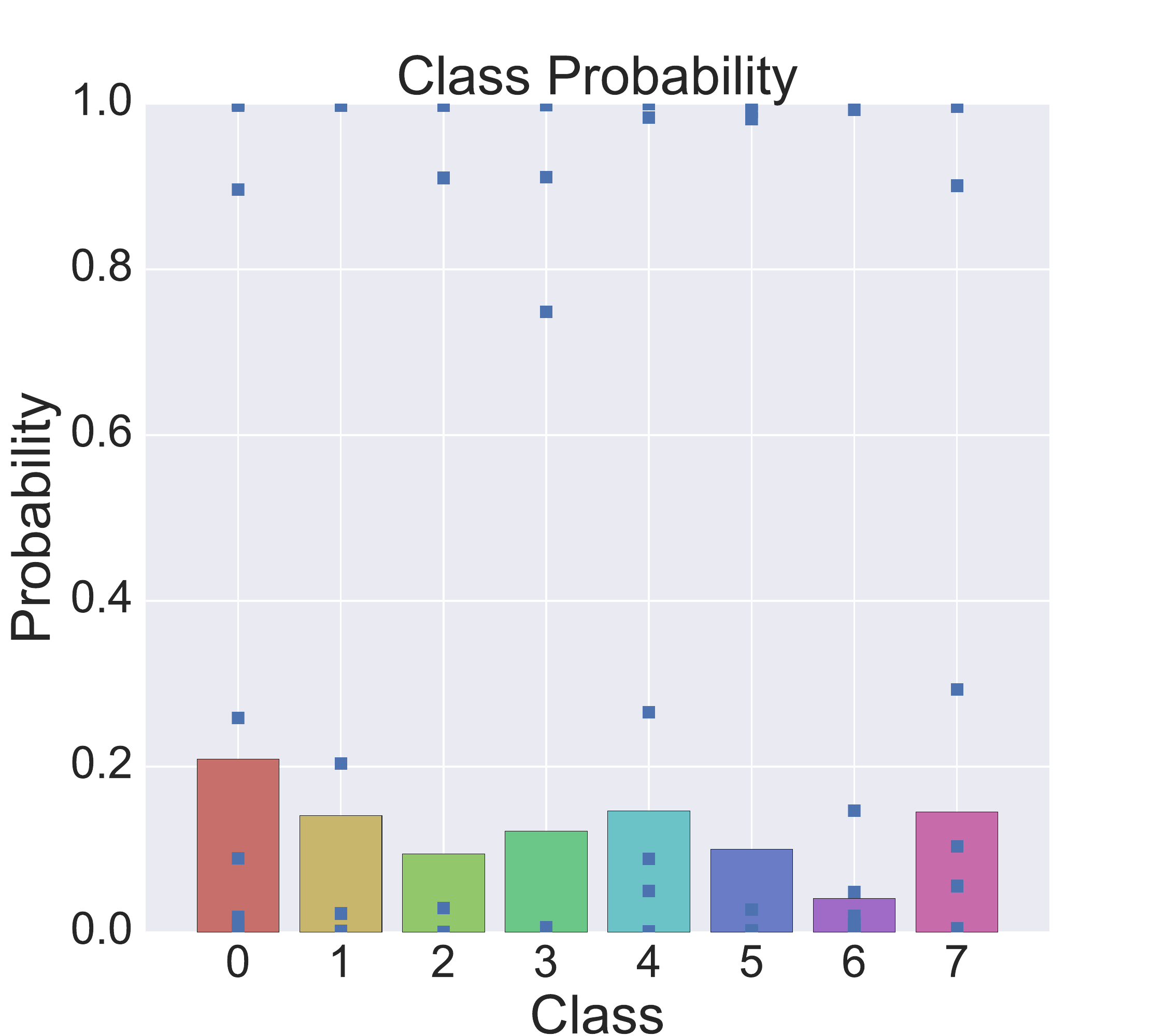}
    \end{minipage}
    
    \begin{minipage}[t]{\textwidth}
        \begin{minipage}[b]{0.14\textwidth}
            \centering{Class 2]}\\ \includegraphics[width=\textwidth]{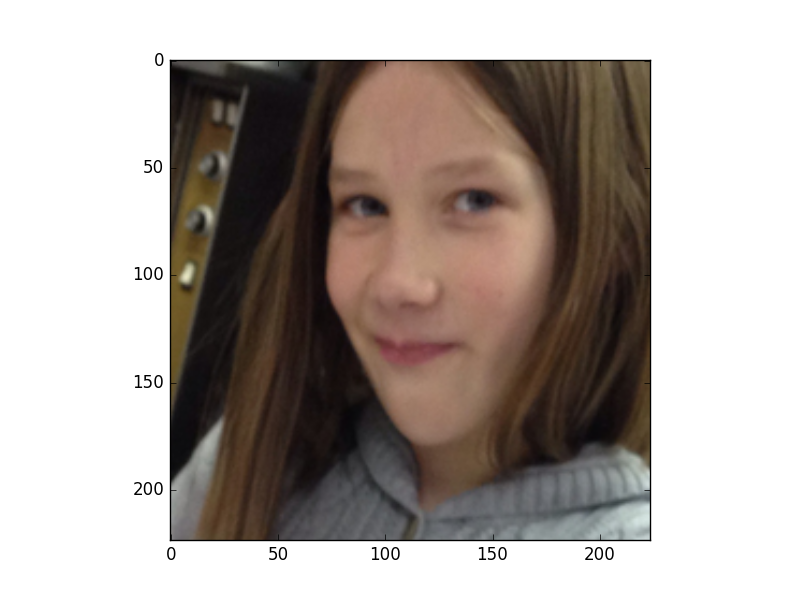}
        \end{minipage}        
        \includegraphics[width=0.15\textwidth]{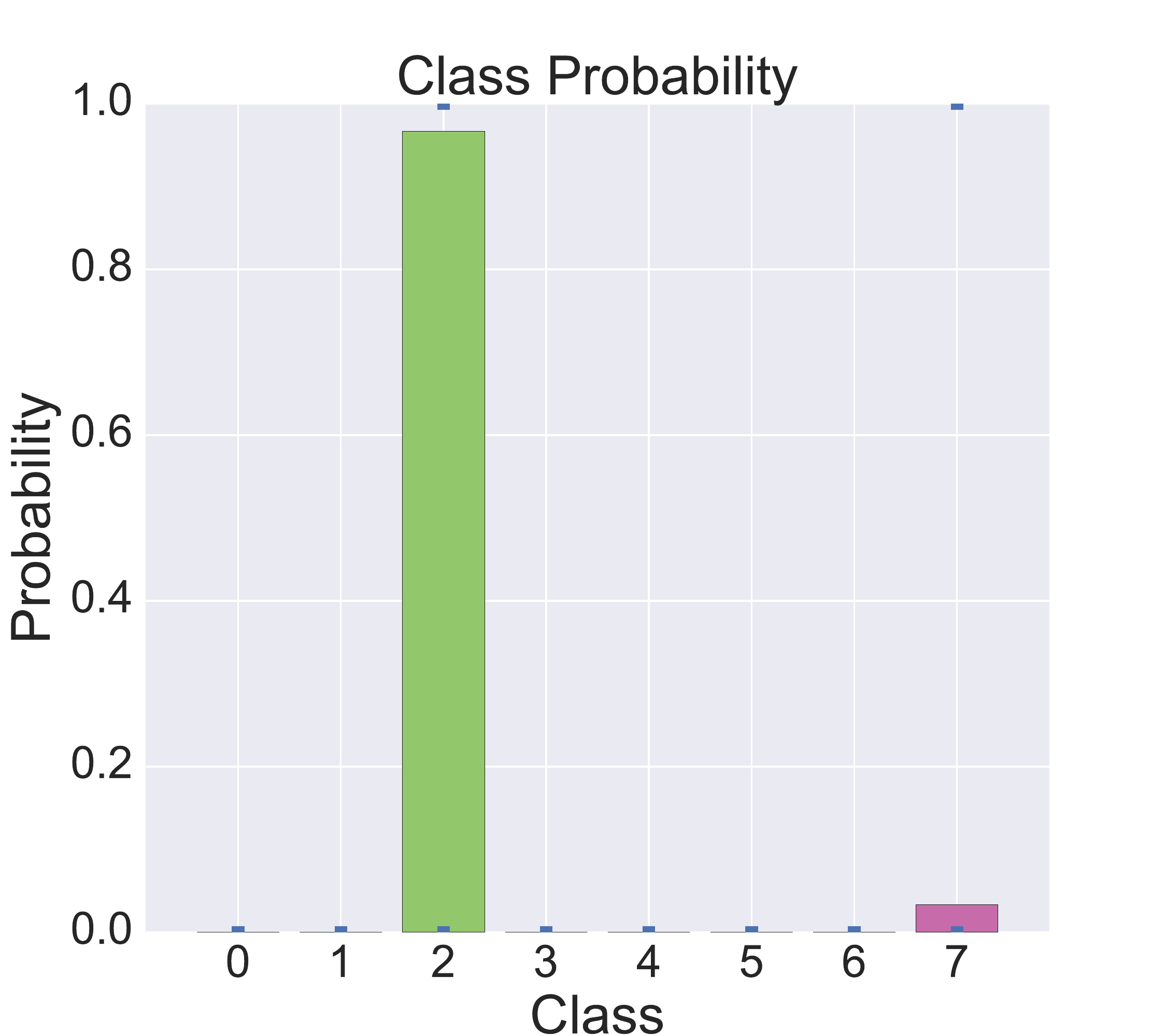}
        \includegraphics[width=0.15\textwidth]{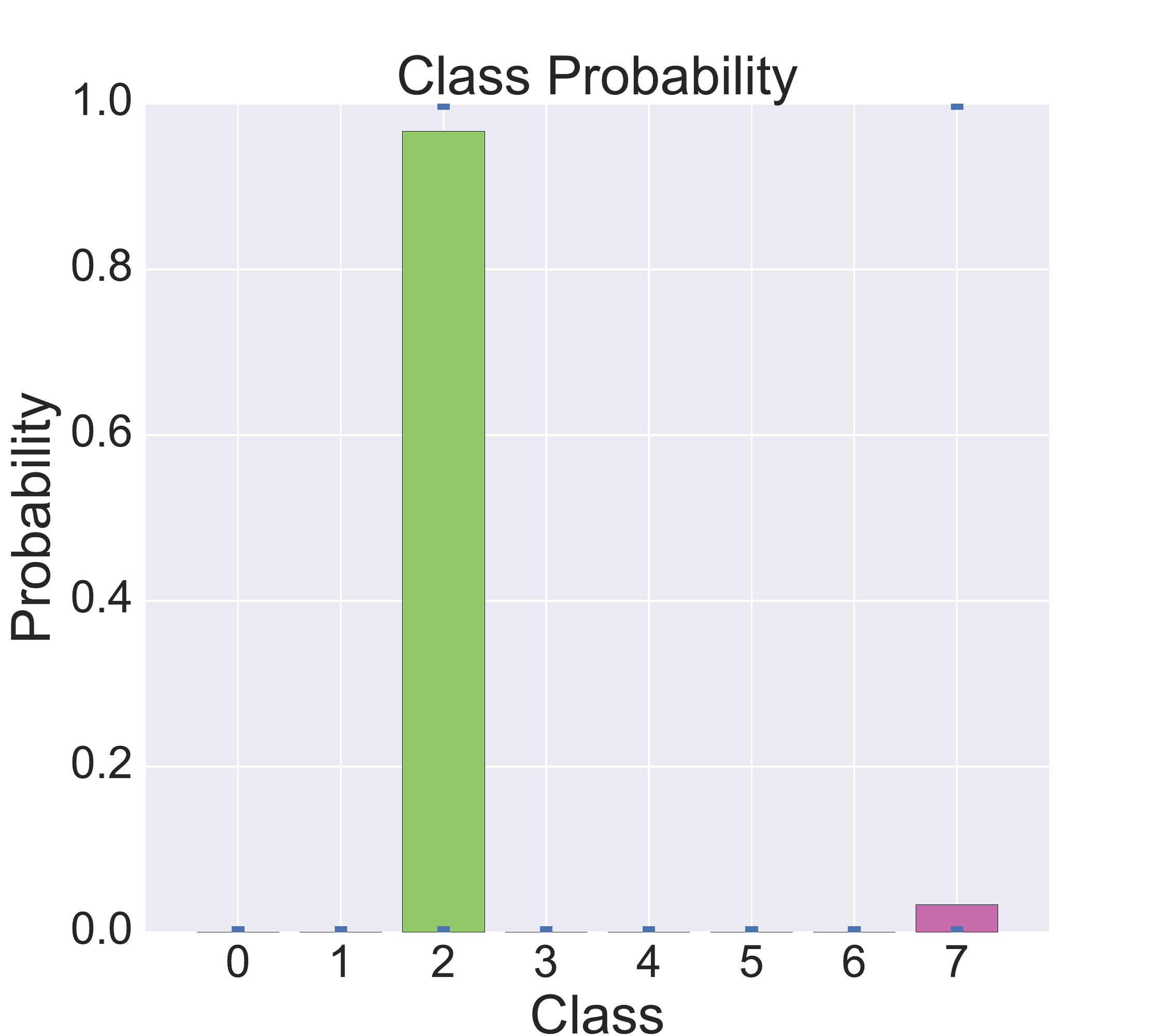}
        \includegraphics[width=0.15\textwidth]{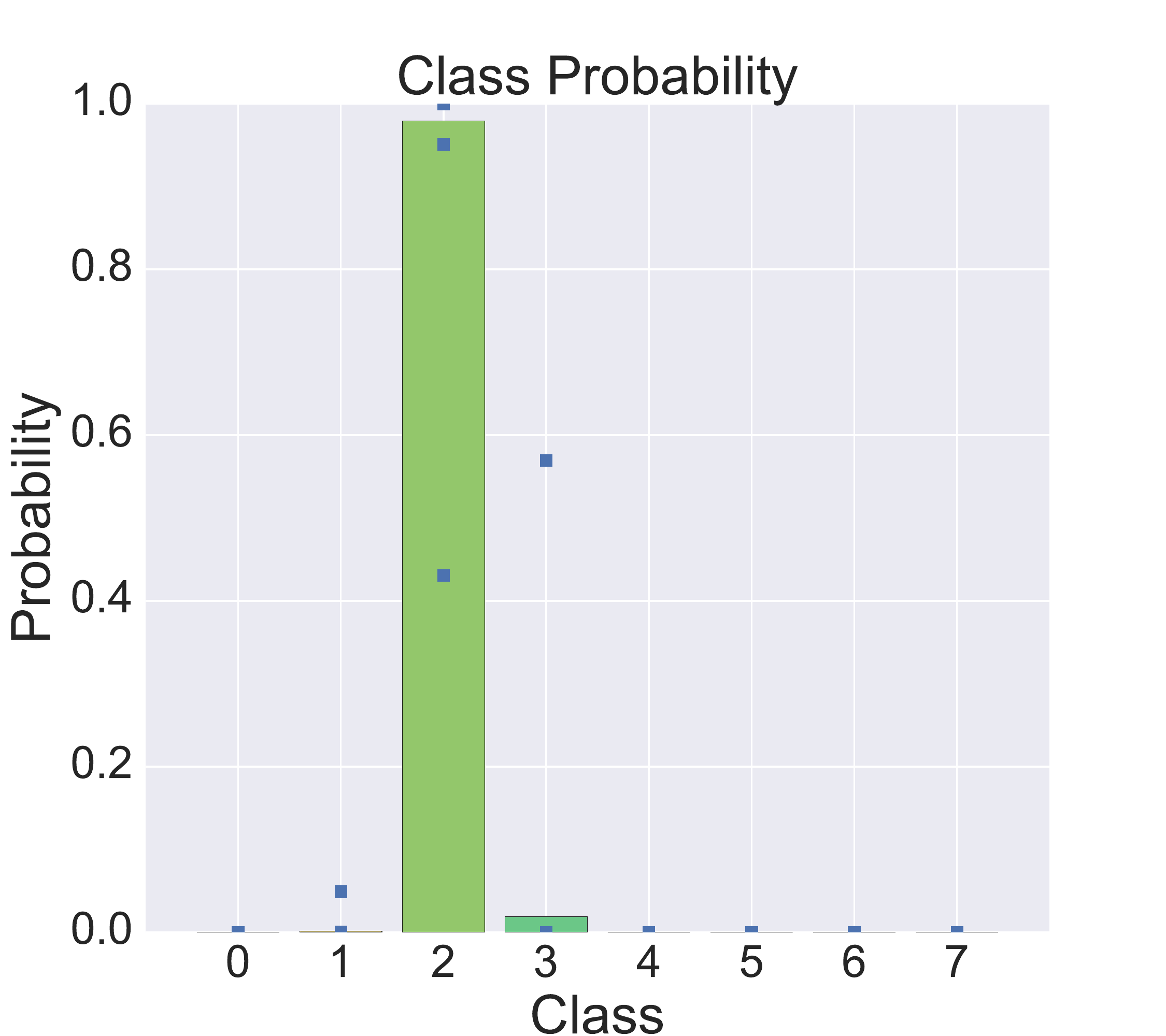}
        \includegraphics[width=0.15\textwidth]{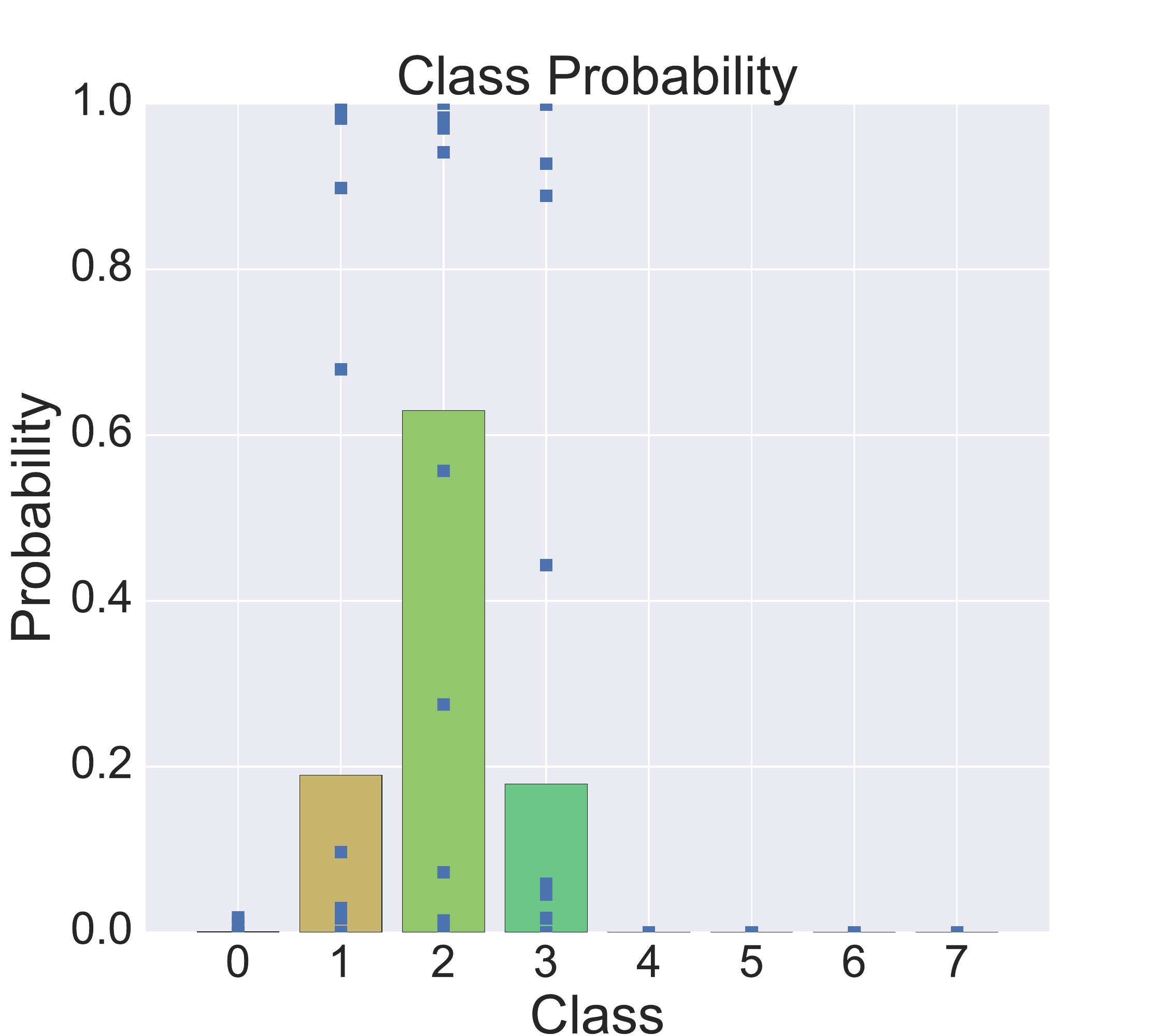}
        \includegraphics[width=0.15\textwidth]{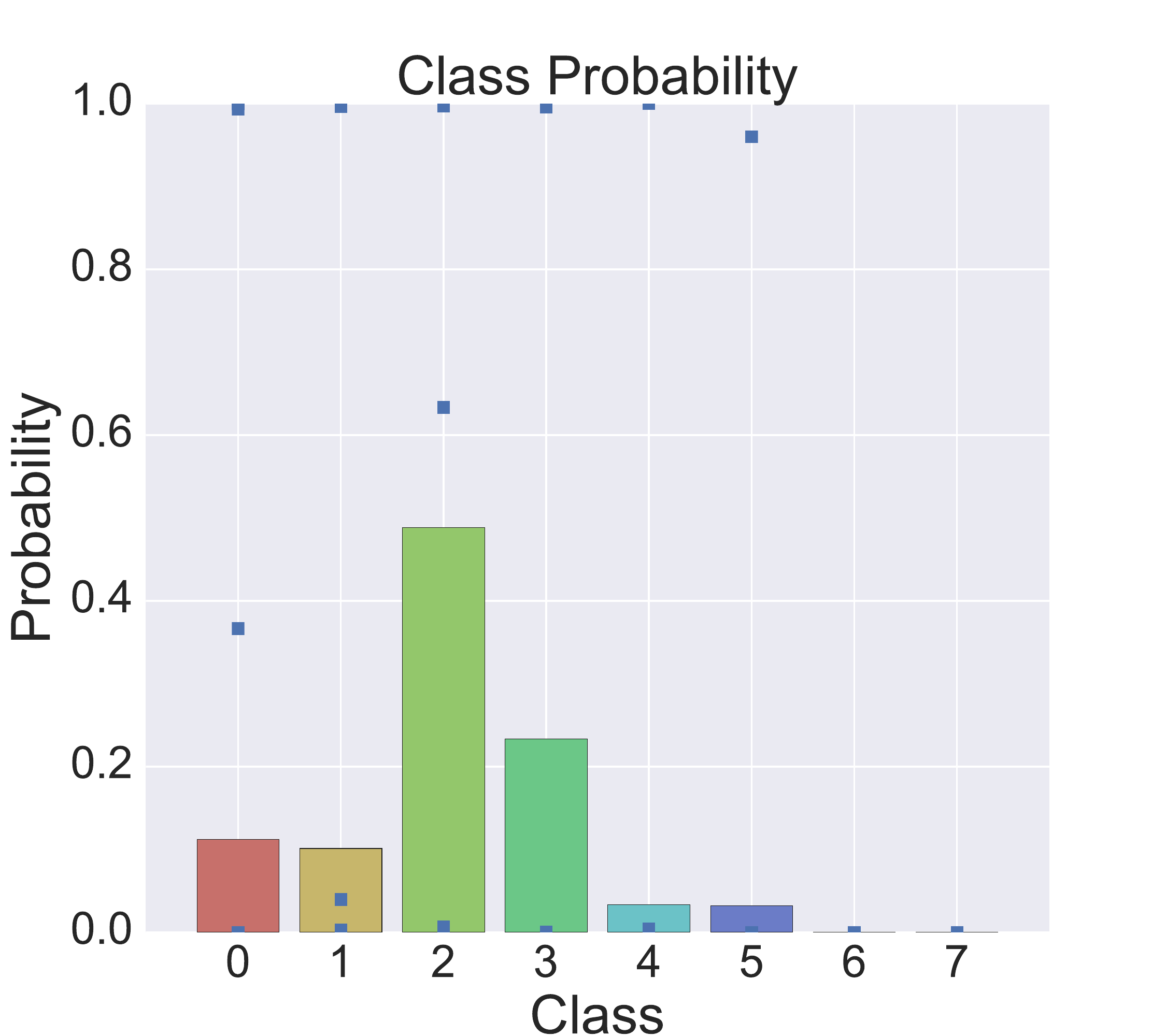}
    \end{minipage}
    
    \begin{minipage}[t]{\textwidth}
        \begin{minipage}[b]{0.14\textwidth}
            \centering{Class 3}\\ \includegraphics[width=\textwidth]{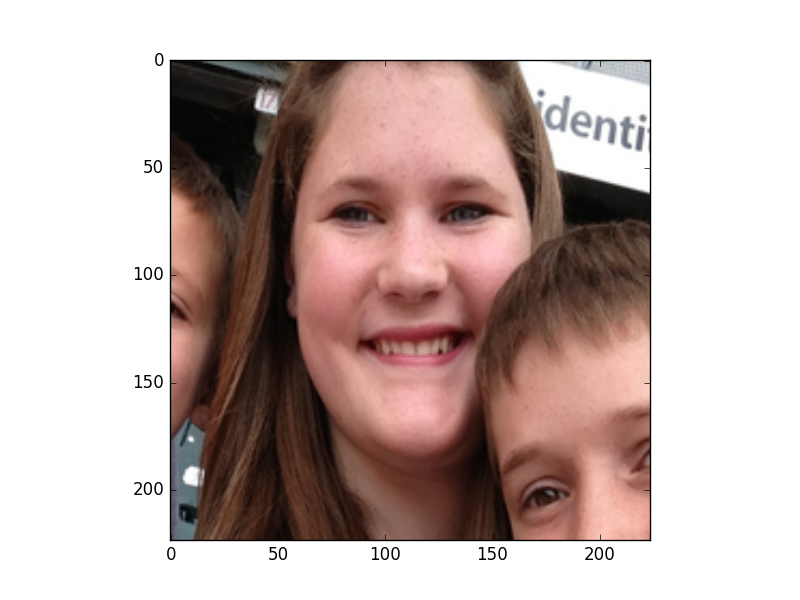}
        \end{minipage}        
        \includegraphics[width=0.15\textwidth]{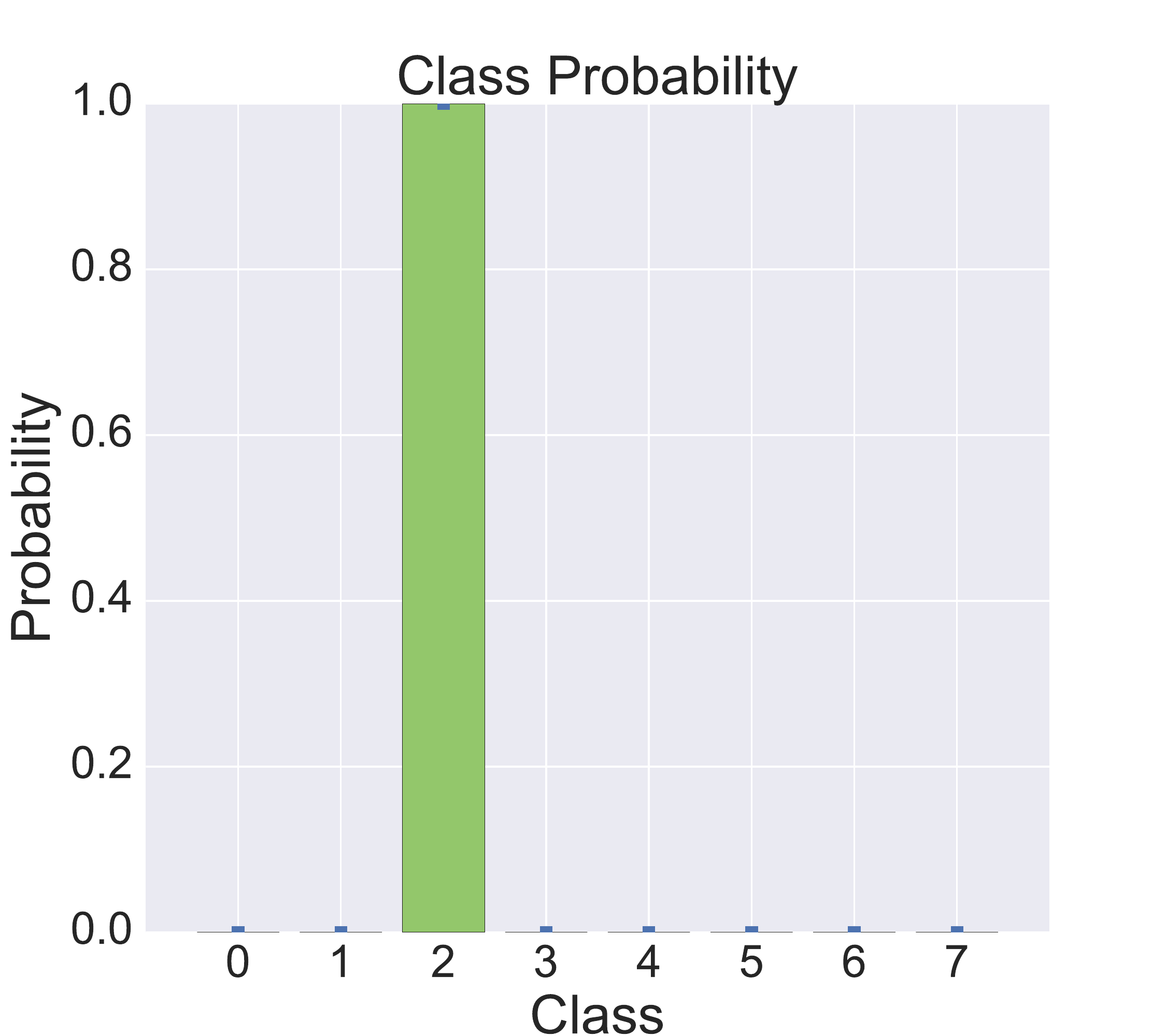}
        \includegraphics[width=0.15\textwidth]{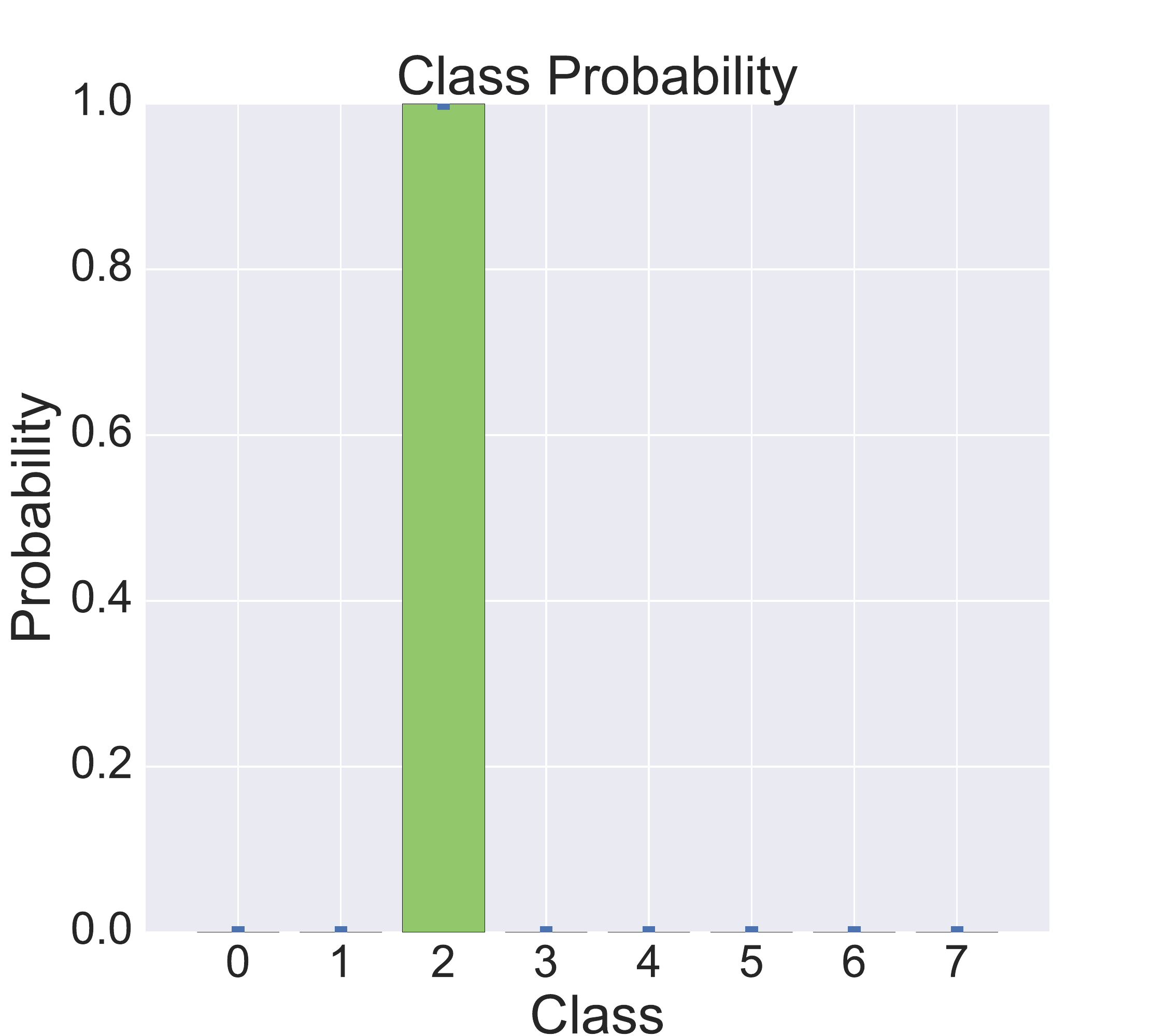}
        \includegraphics[width=0.15\textwidth]{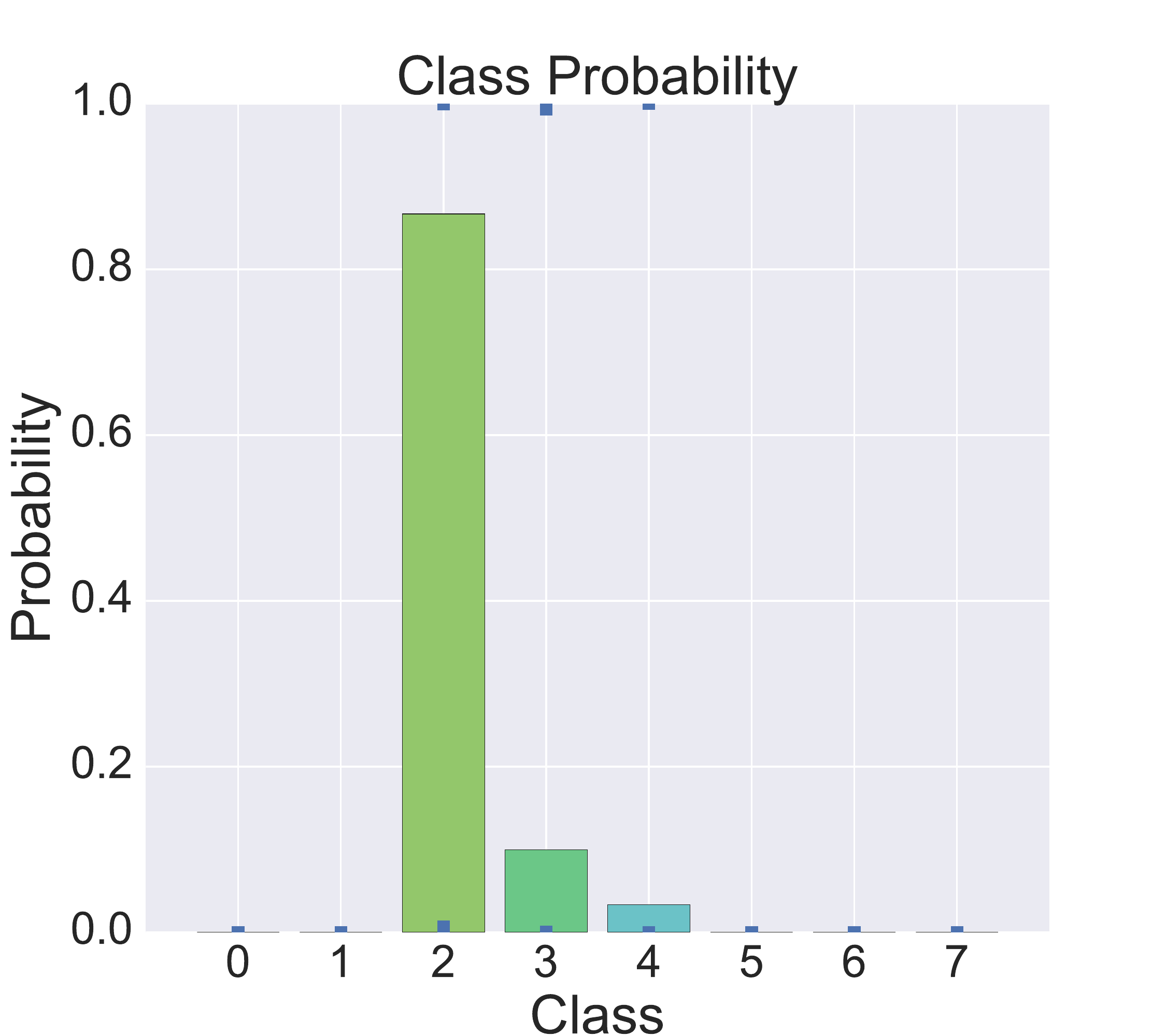}
        \includegraphics[width=0.15\textwidth]{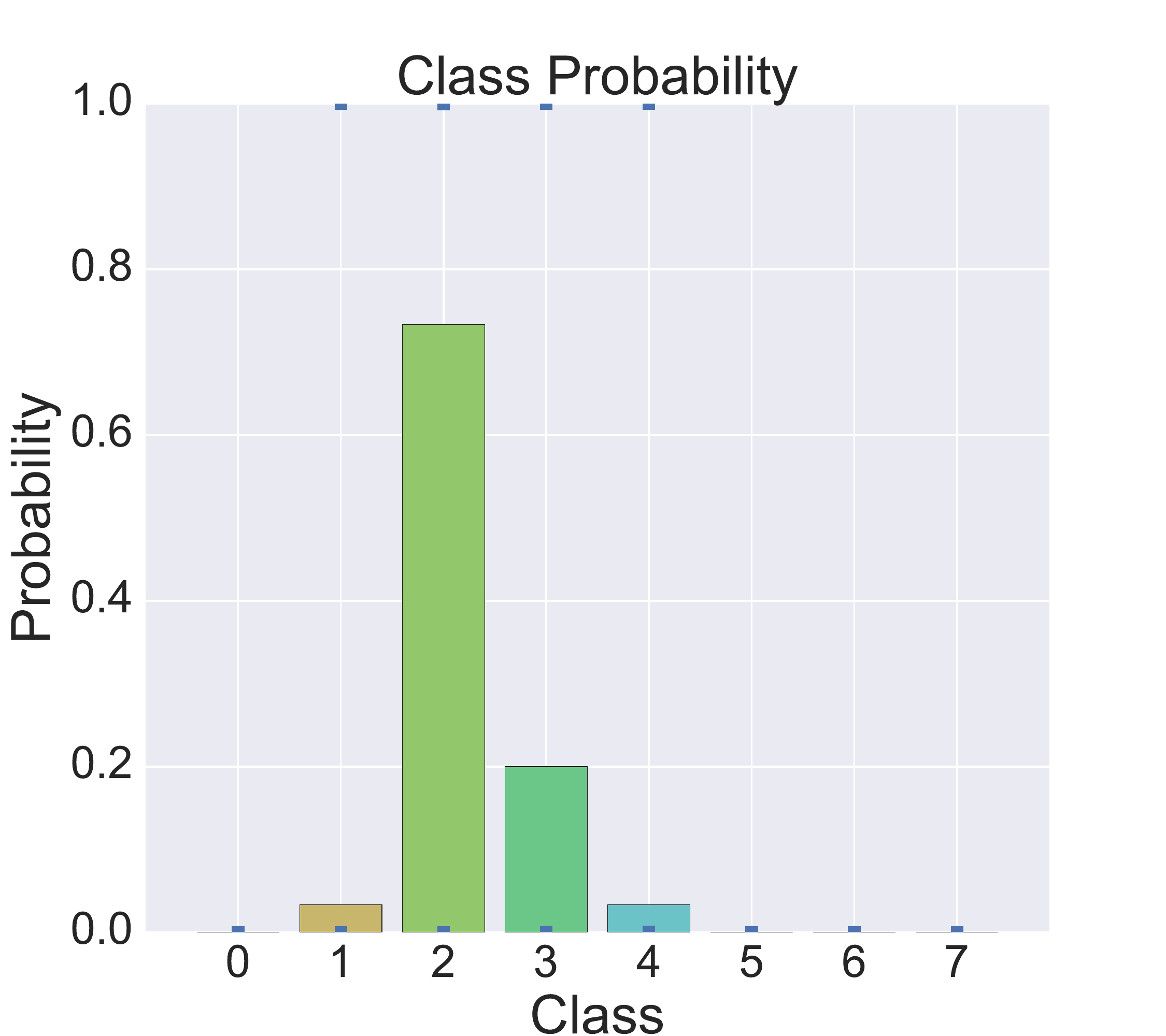}
        \includegraphics[width=0.15\textwidth]{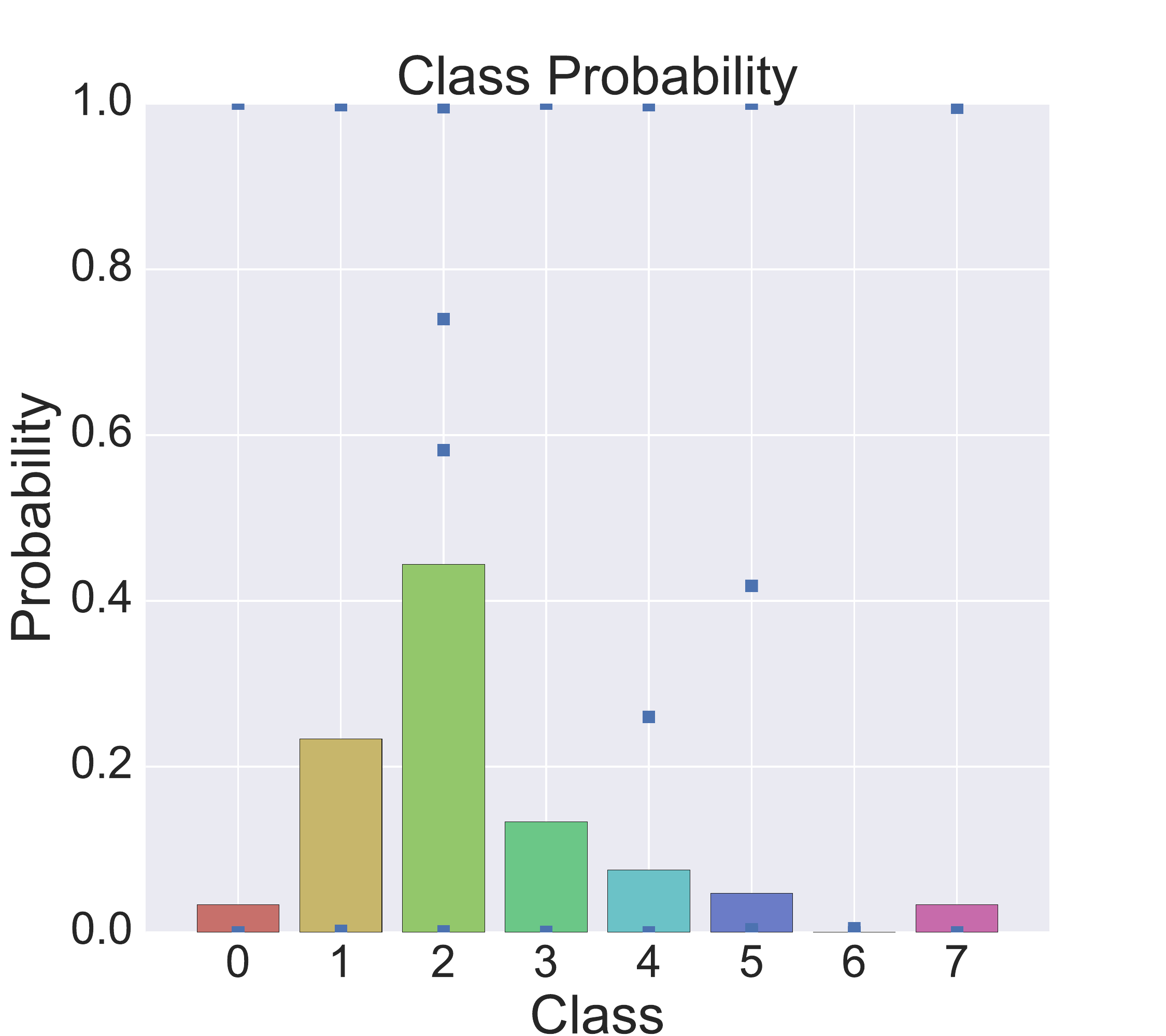}
    \end{minipage}
    
    \begin{minipage}[t]{\textwidth}
        \begin{minipage}[b]{0.14\textwidth}
            \centering{Class 4}\\ \includegraphics[width=\textwidth]{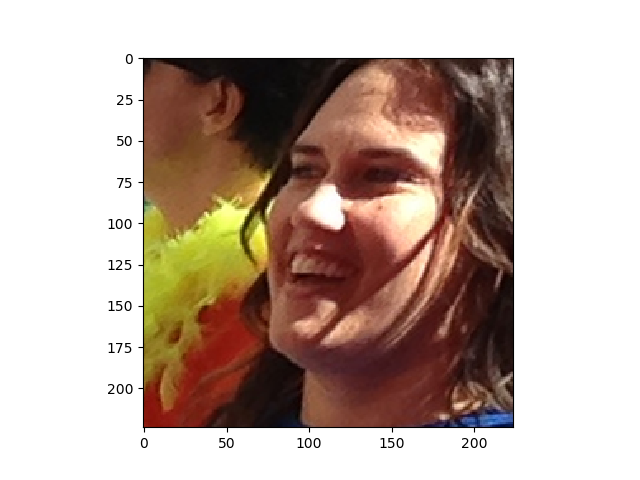}
        \end{minipage}        
        \includegraphics[width=0.15\textwidth]{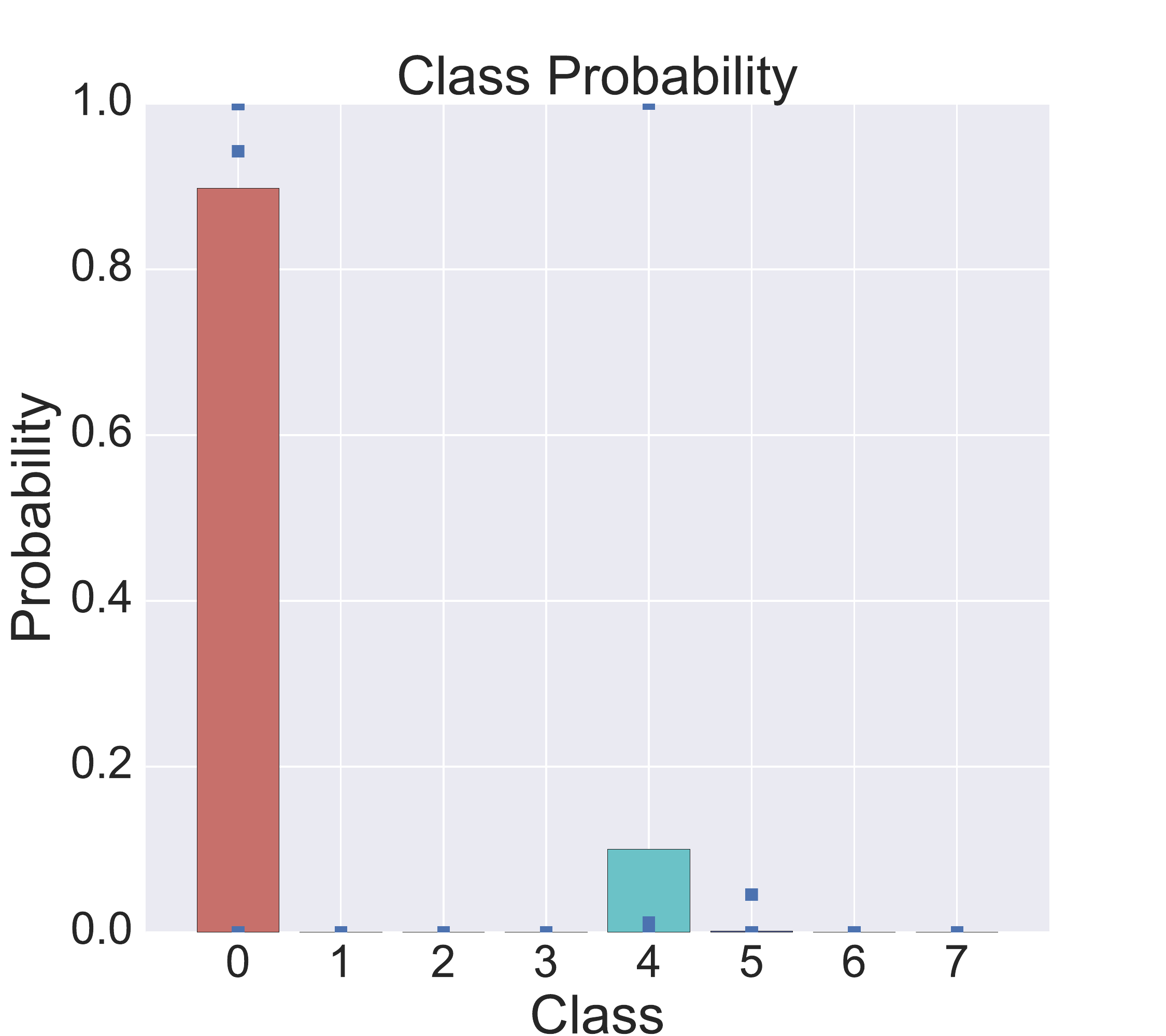}
        \includegraphics[width=0.15\textwidth]{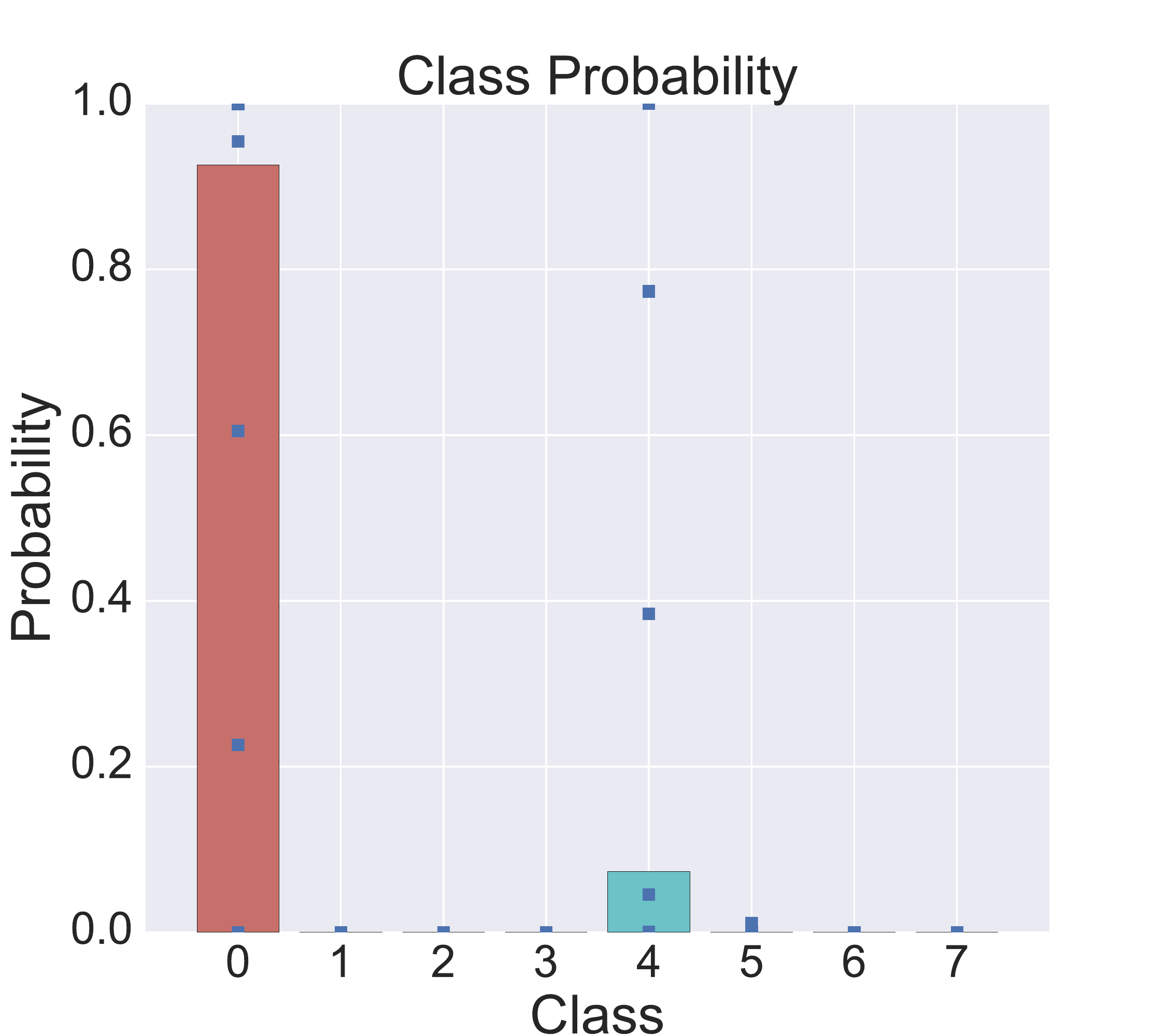}
        \includegraphics[width=0.15\textwidth]{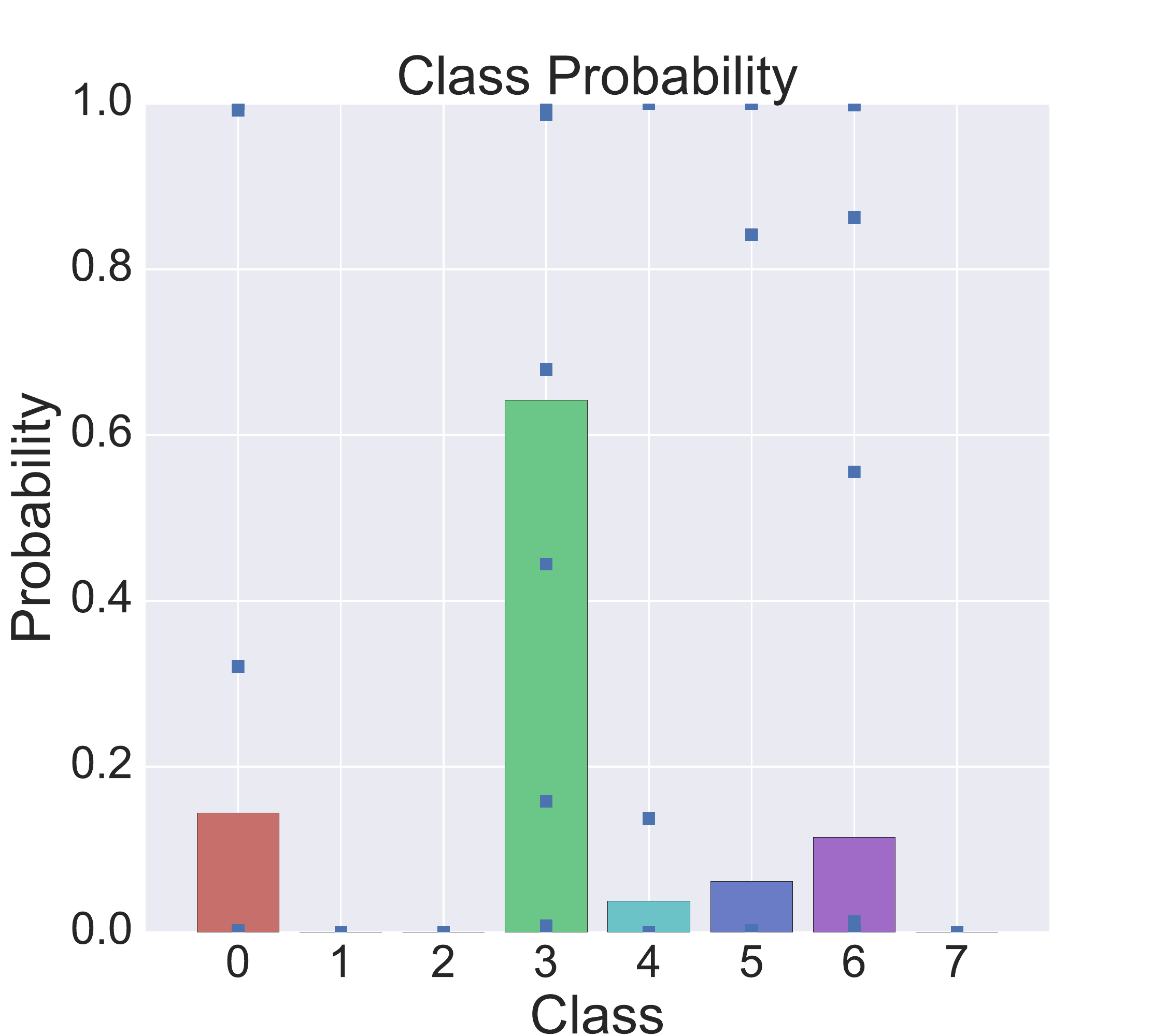}
        \includegraphics[width=0.15\textwidth]{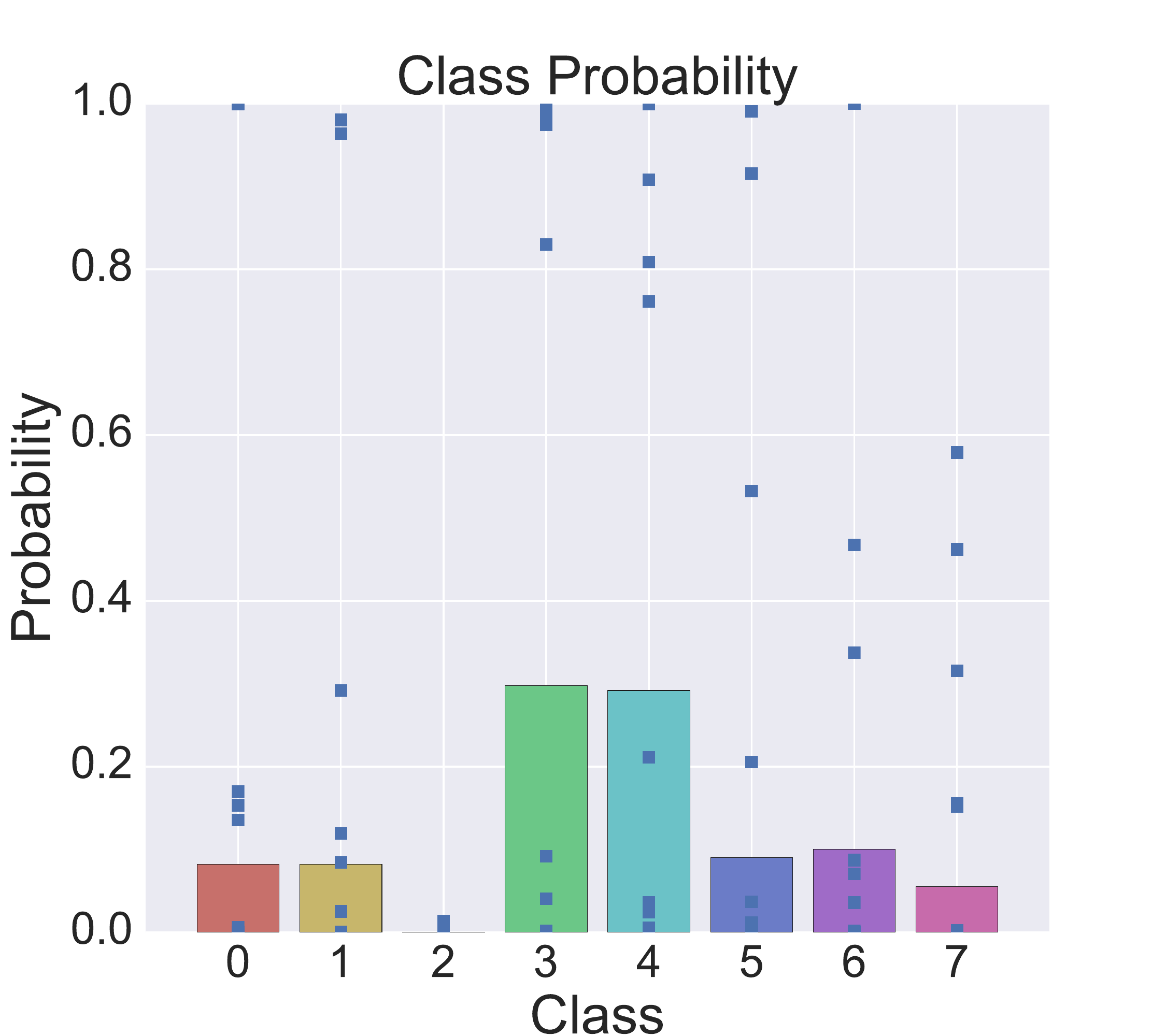}
        \includegraphics[width=0.15\textwidth]{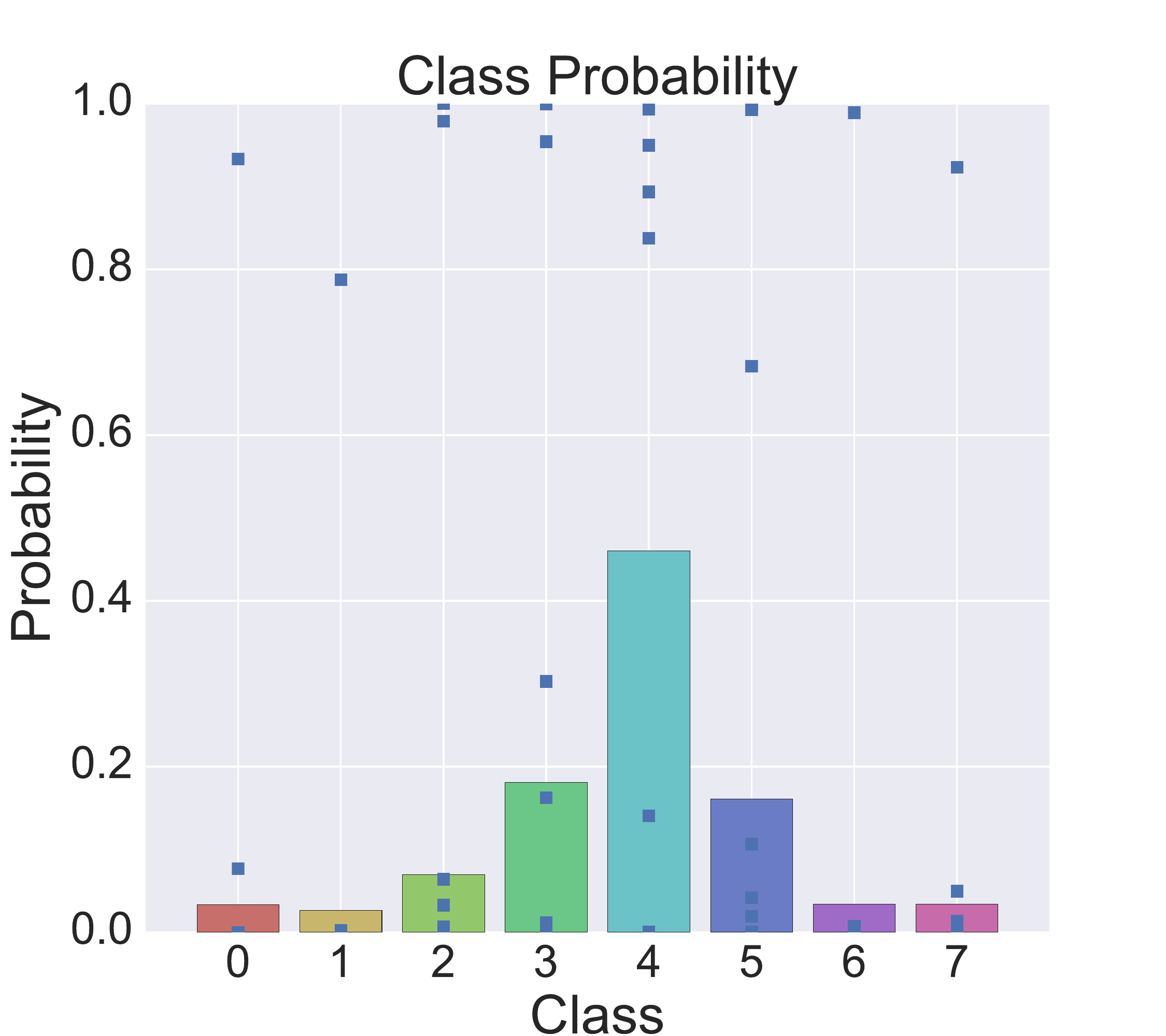}
    \end{minipage}

    \begin{minipage}[t]{\textwidth}
        \begin{minipage}[b]{0.14\textwidth}
            \centering{Class 5}\\ \includegraphics[width=\textwidth]{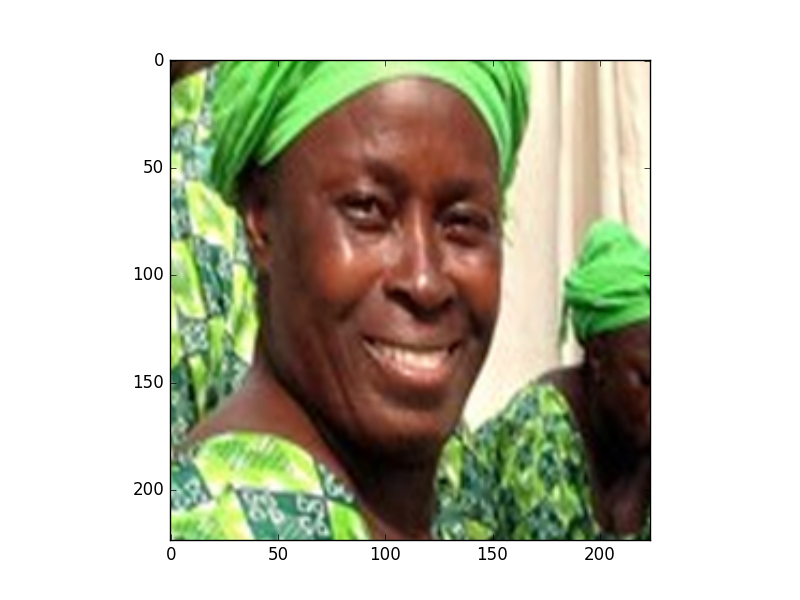}
        \end{minipage}        
        \includegraphics[width=0.15\textwidth]{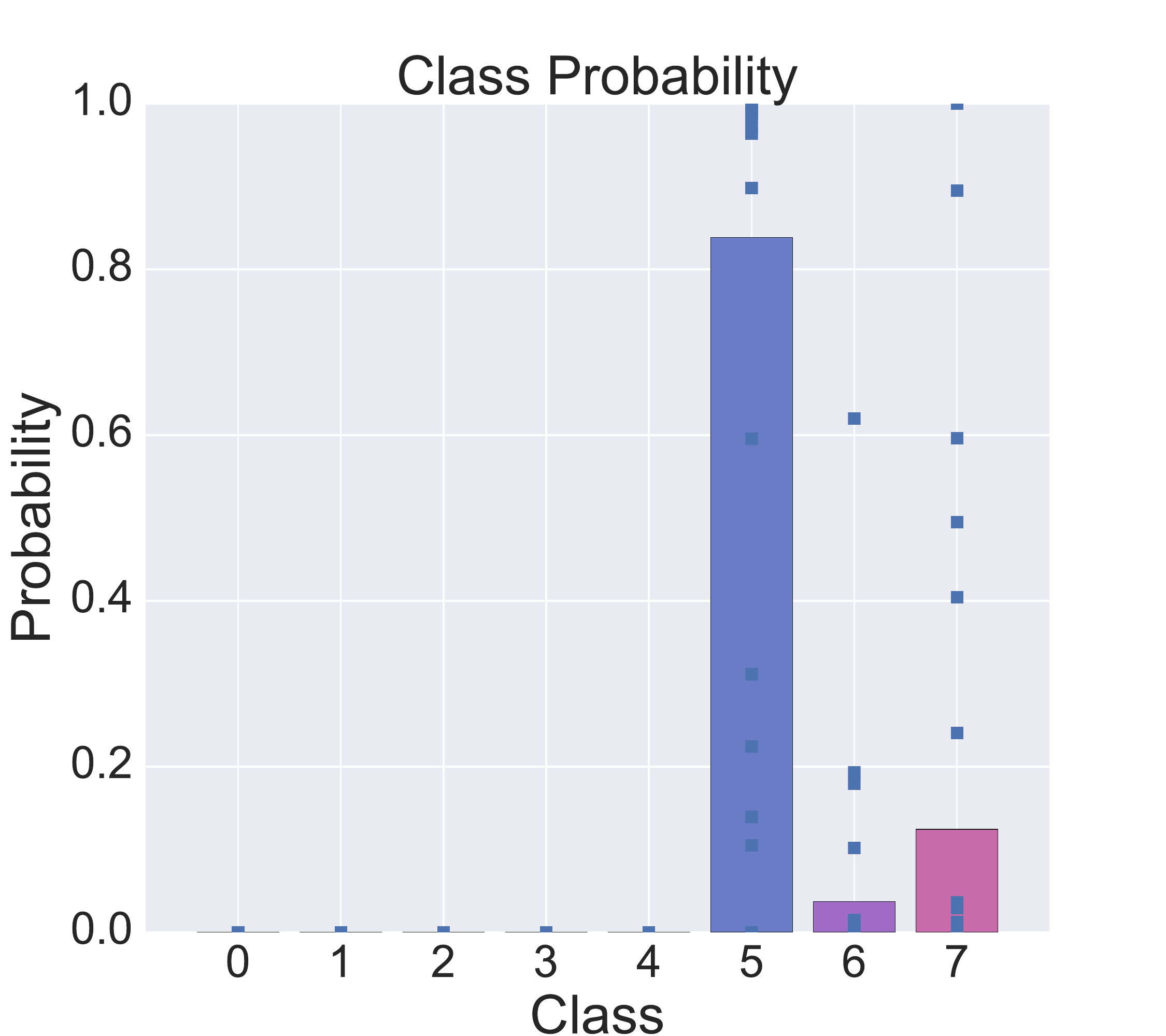}
        \includegraphics[width=0.15\textwidth]{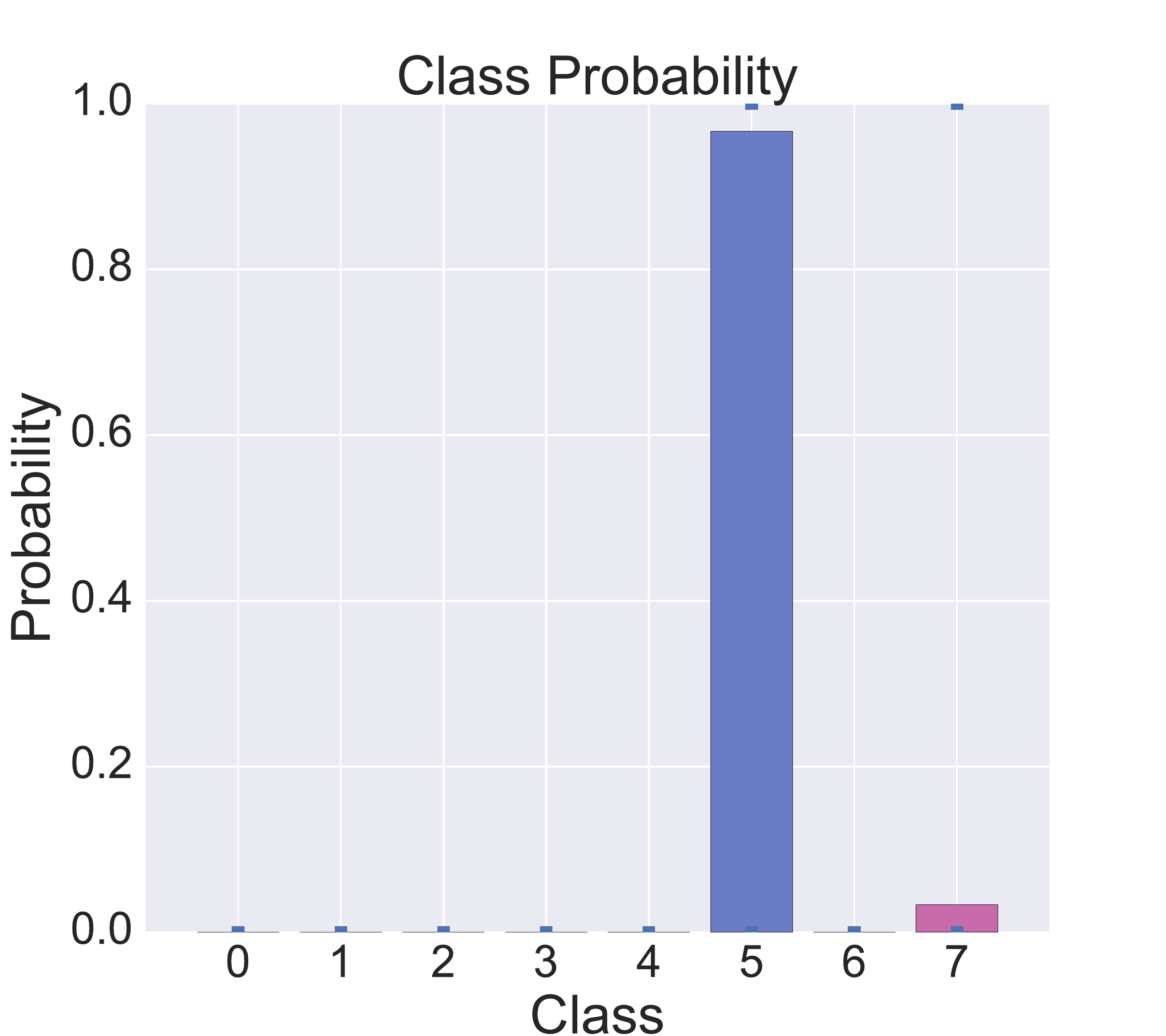}
        \includegraphics[width=0.15\textwidth]{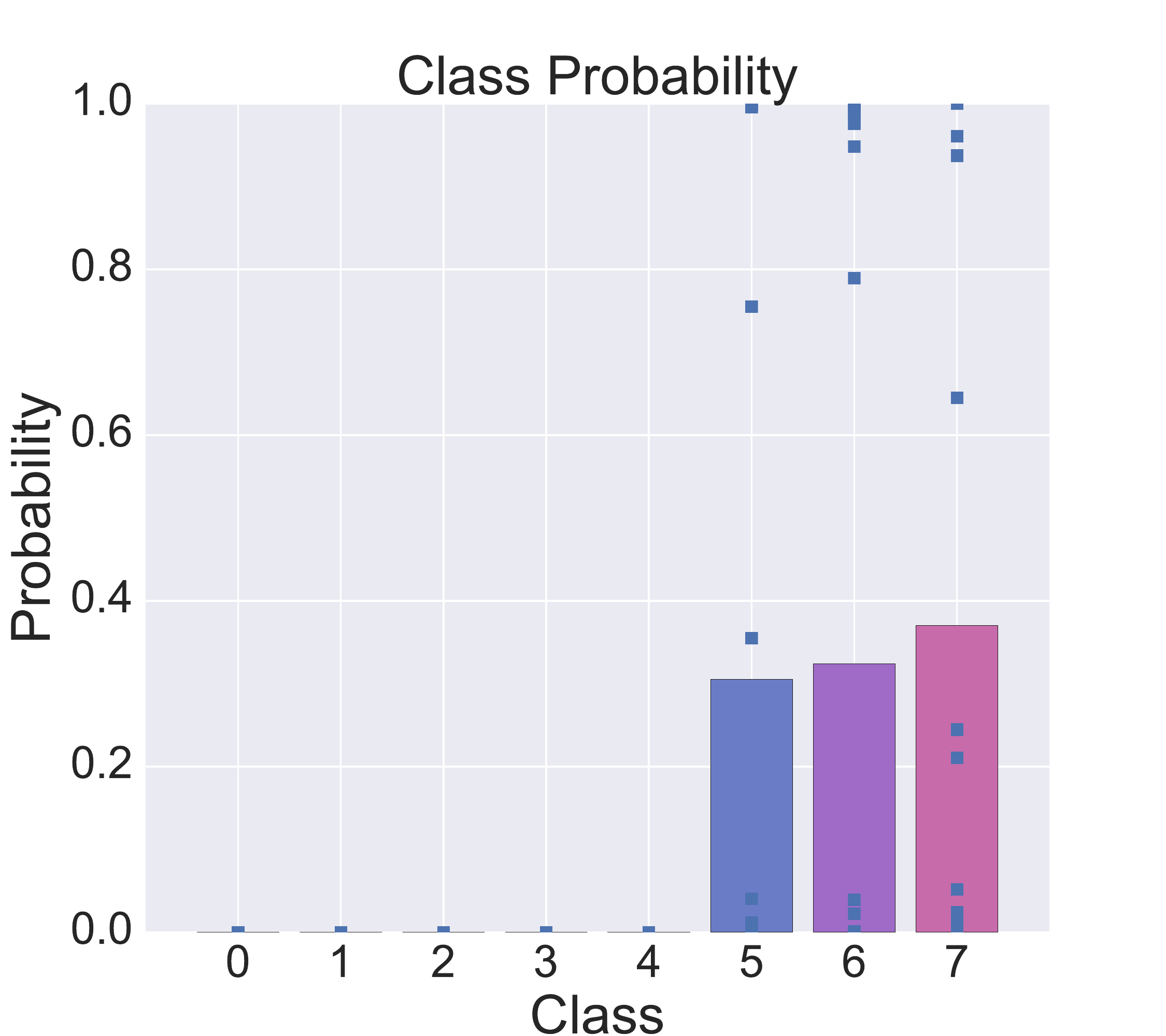}
        \includegraphics[width=0.15\textwidth]{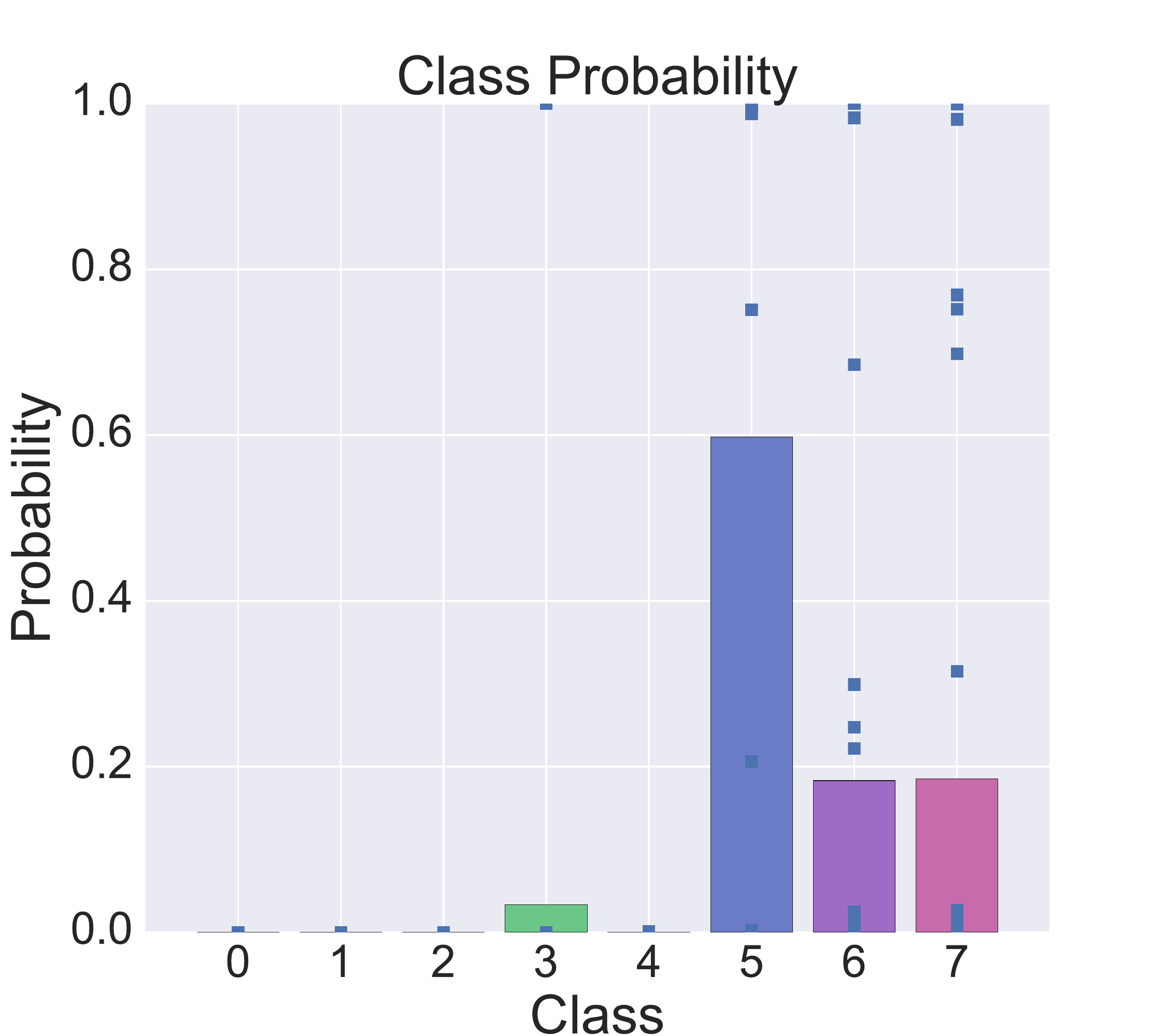}
        \includegraphics[width=0.15\textwidth]{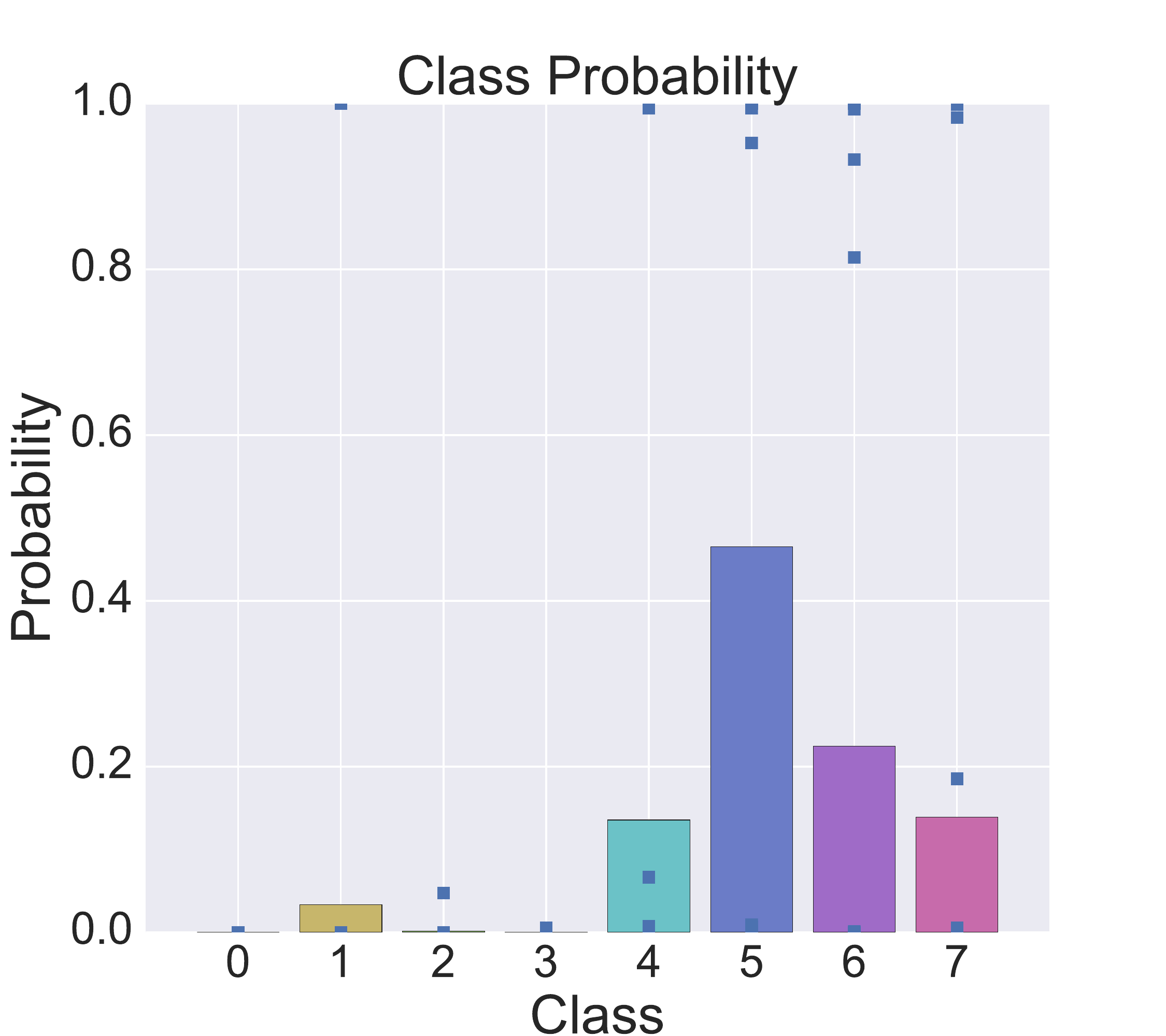}
    \end{minipage}

    \begin{minipage}[t]{\textwidth}
        \begin{minipage}[b]{0.14\textwidth}
            \centering{Class 6}\\ \includegraphics[width=\textwidth]{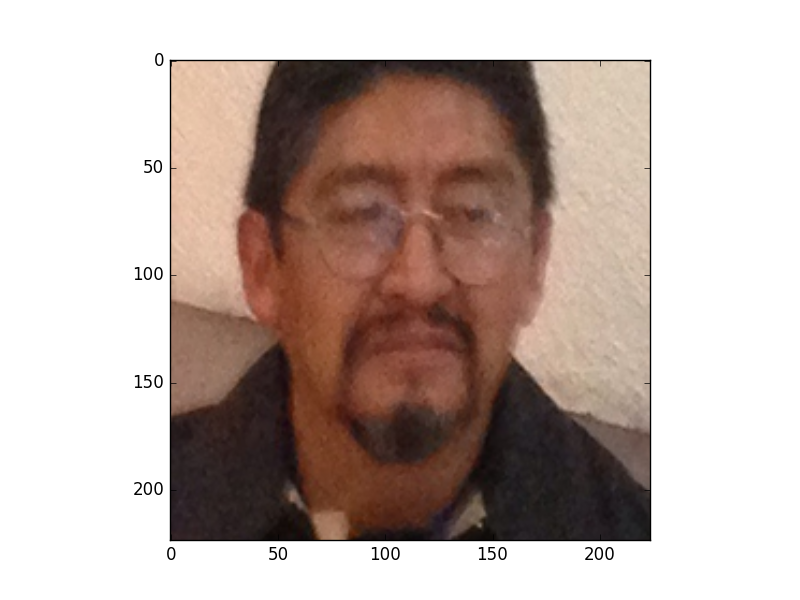}
        \end{minipage}        
        \includegraphics[width=0.15\textwidth]{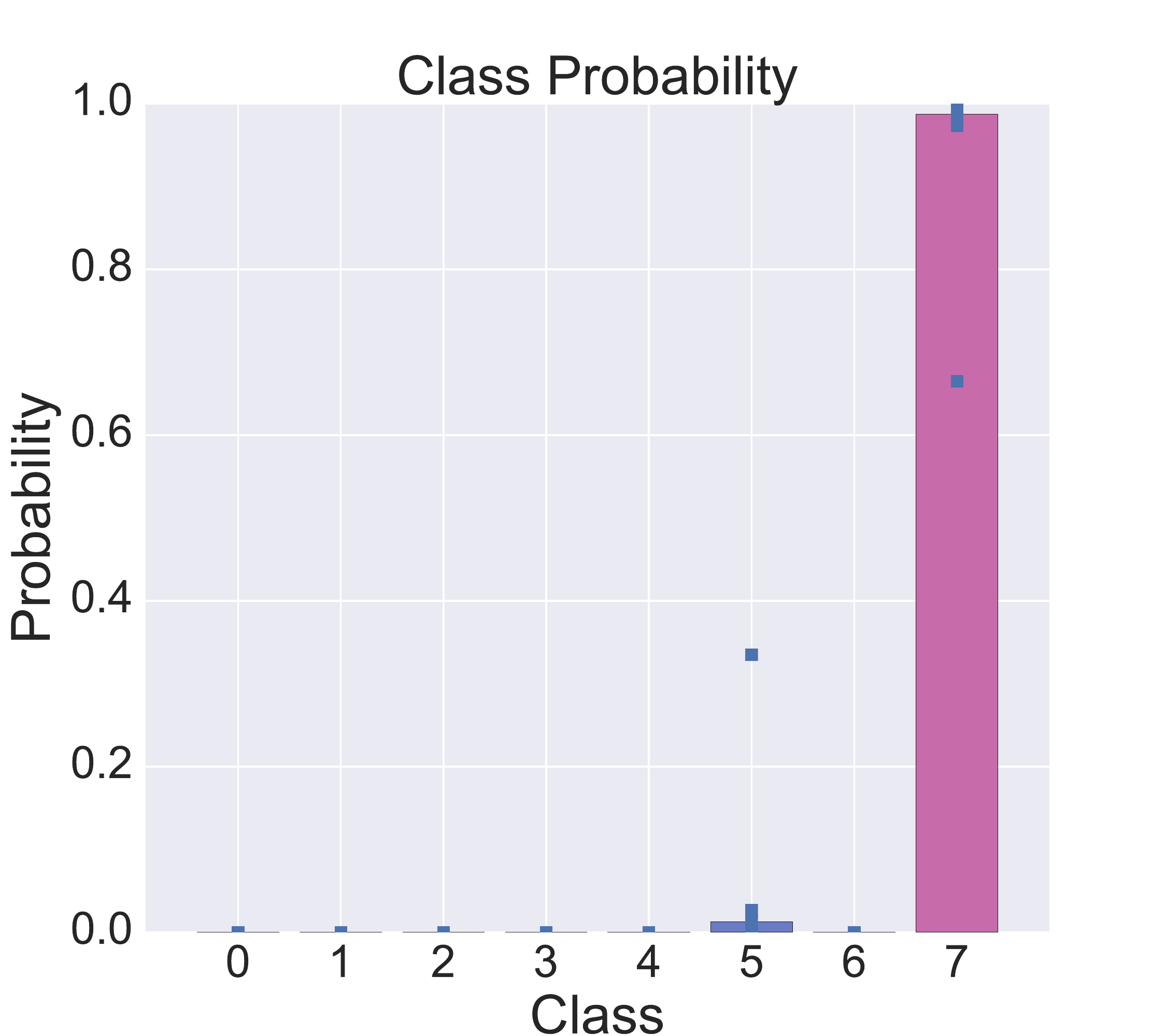}
        \includegraphics[width=0.15\textwidth]{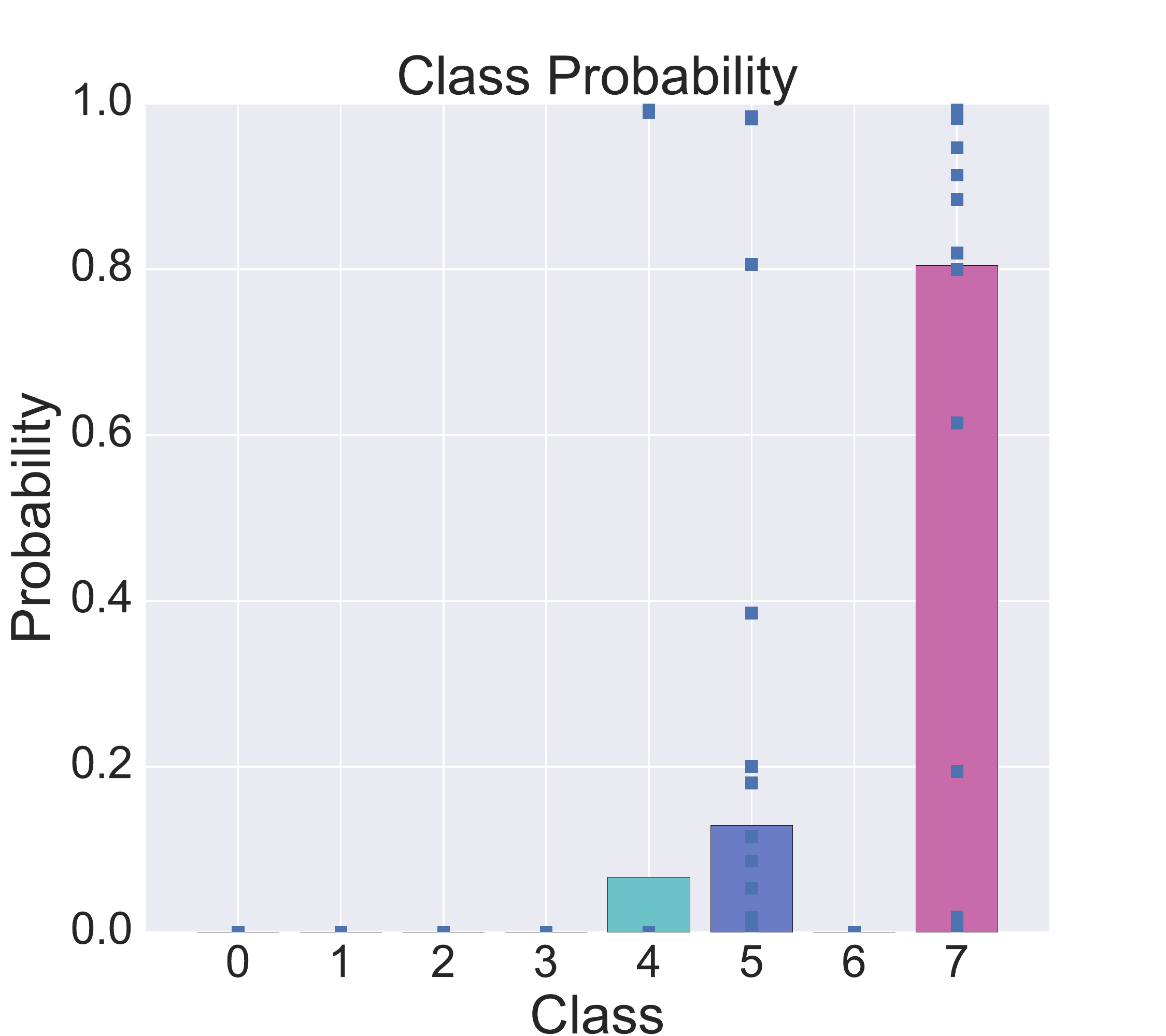}
        \includegraphics[width=0.15\textwidth]{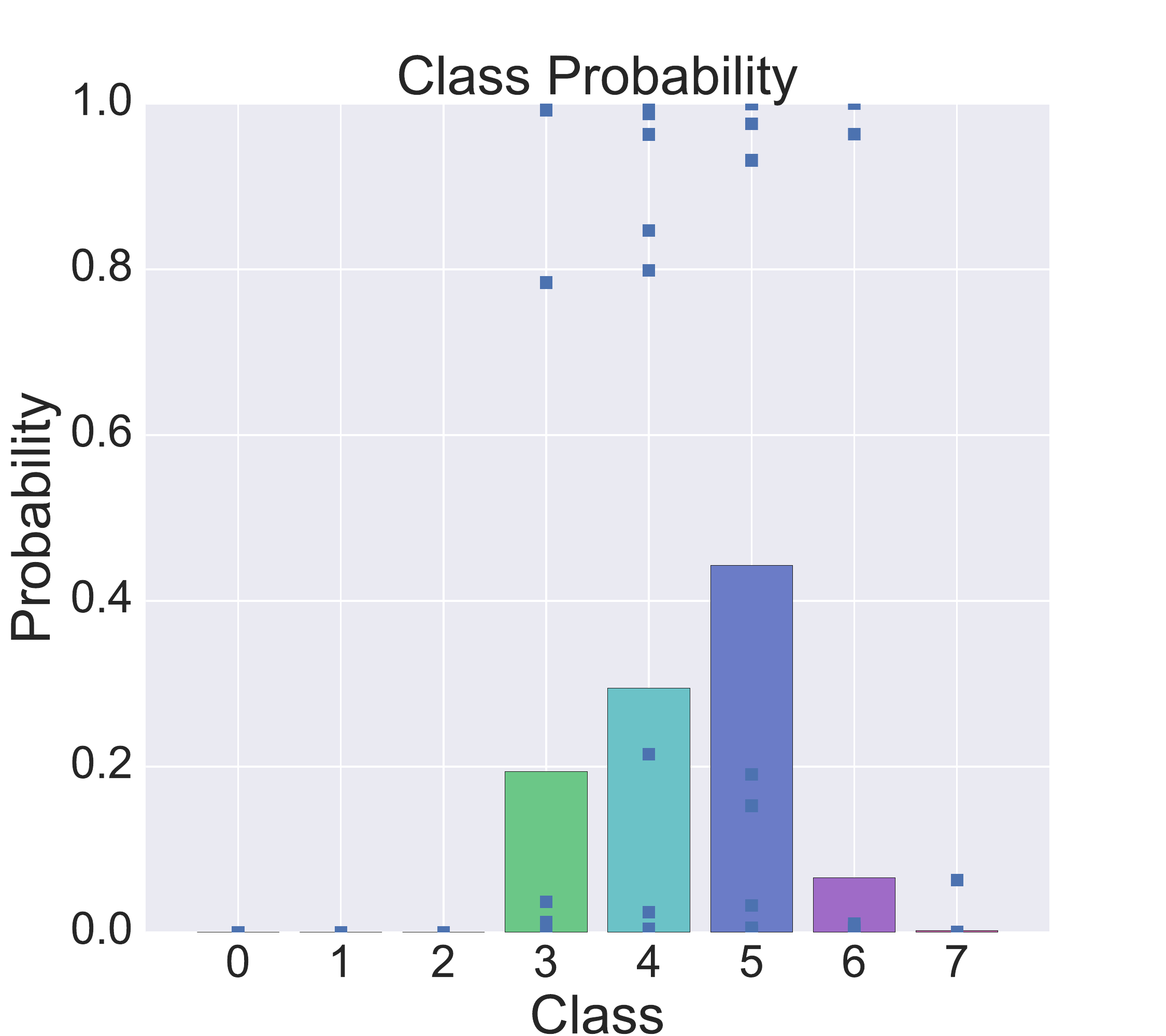}
        \includegraphics[width=0.15\textwidth]{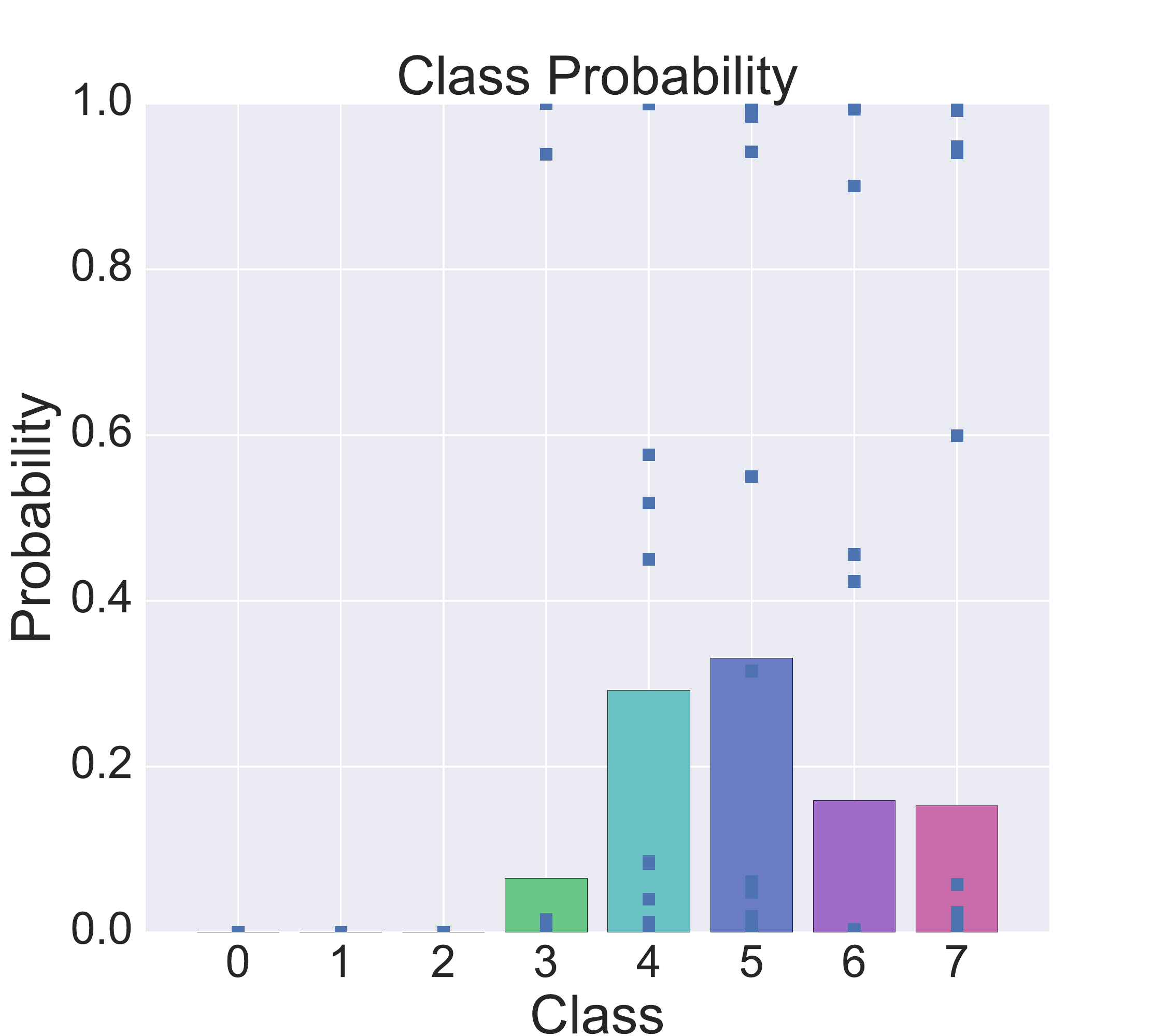}
        \includegraphics[width=0.15\textwidth]{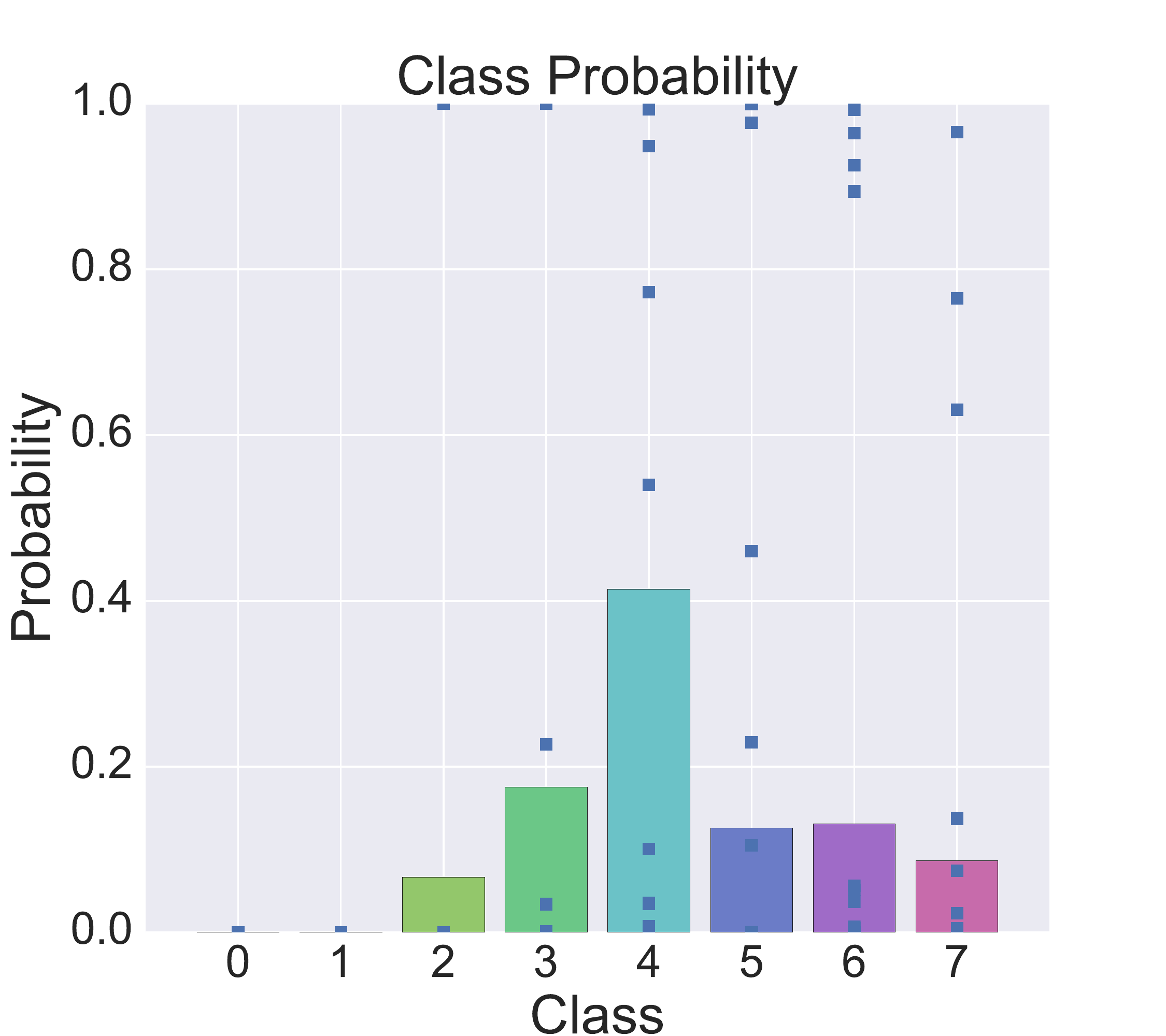}
    \end{minipage}

    \begin{minipage}[t]{\textwidth}
        \begin{minipage}[b]{0.14\textwidth}
            \centering{Class 7}\\ \includegraphics[width=\textwidth]{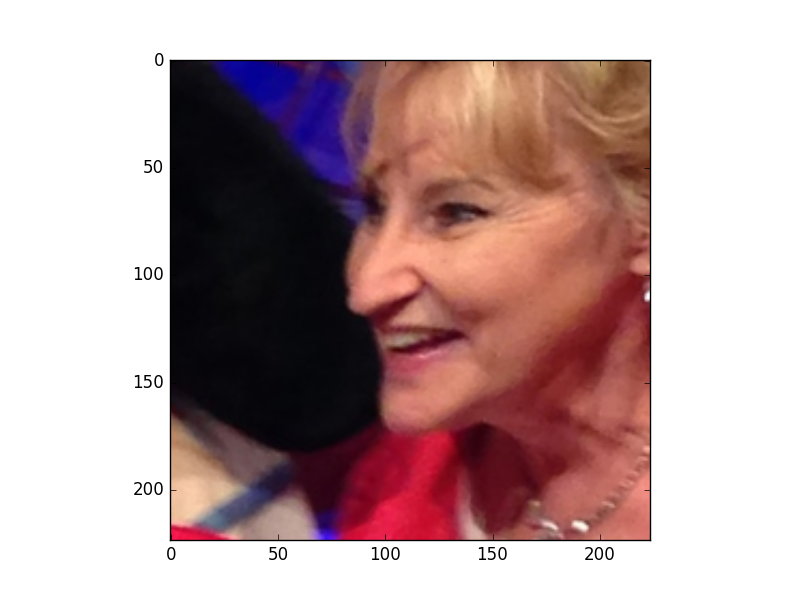}
        \end{minipage}        
        \includegraphics[width=0.15\textwidth]{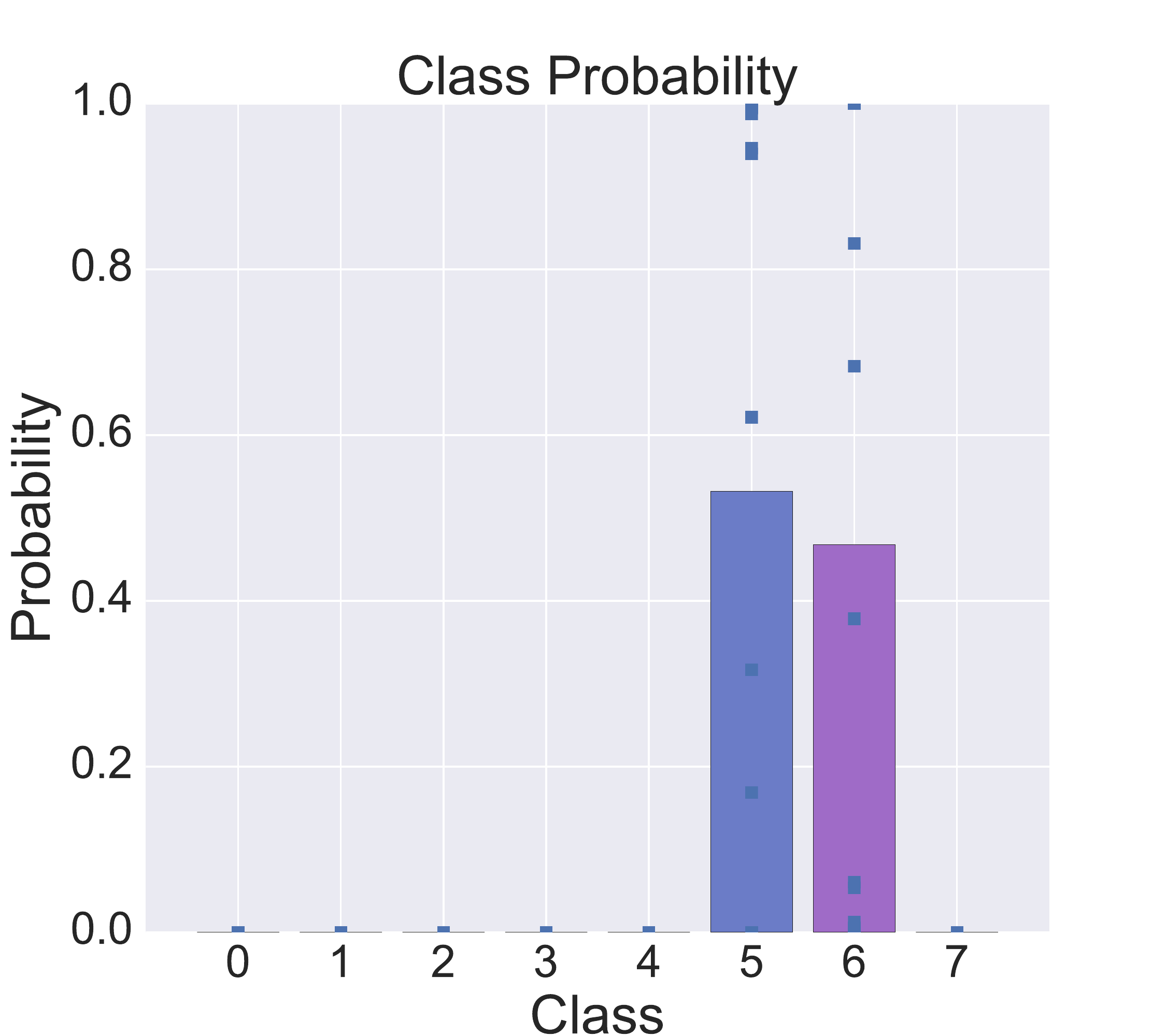}
        \includegraphics[width=0.15\textwidth]{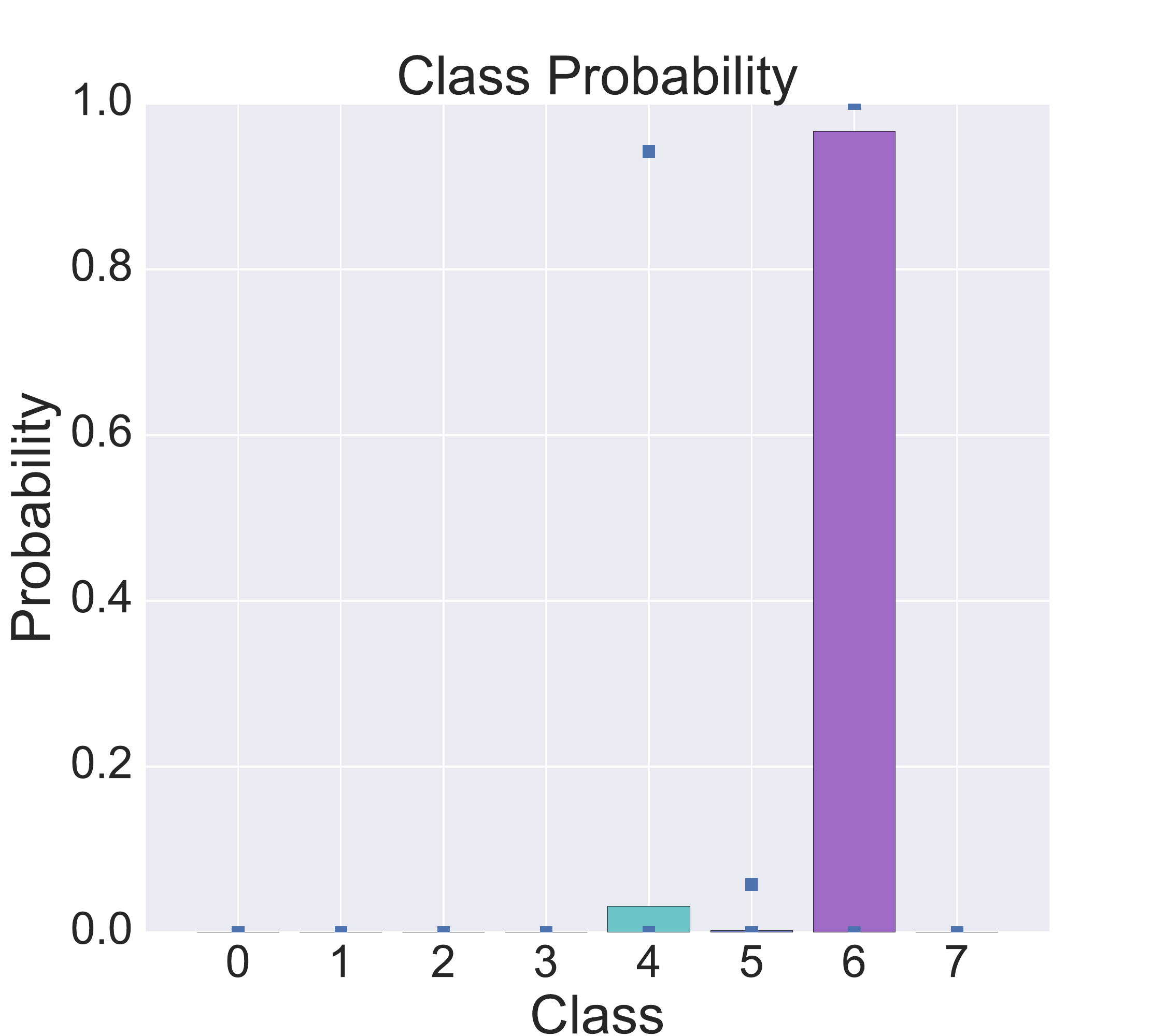}
        \includegraphics[width=0.15\textwidth]{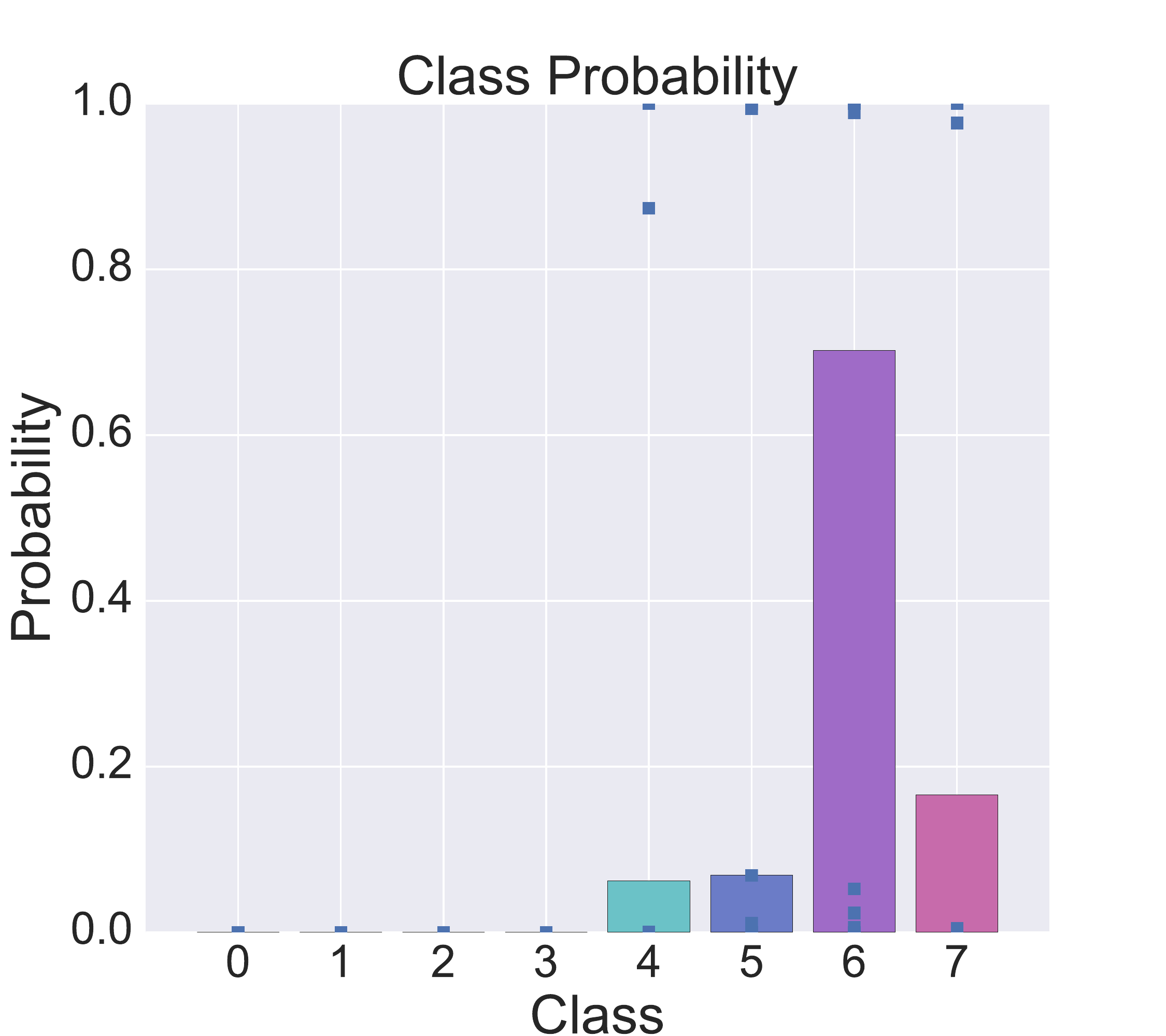}
        \includegraphics[width=0.15\textwidth]{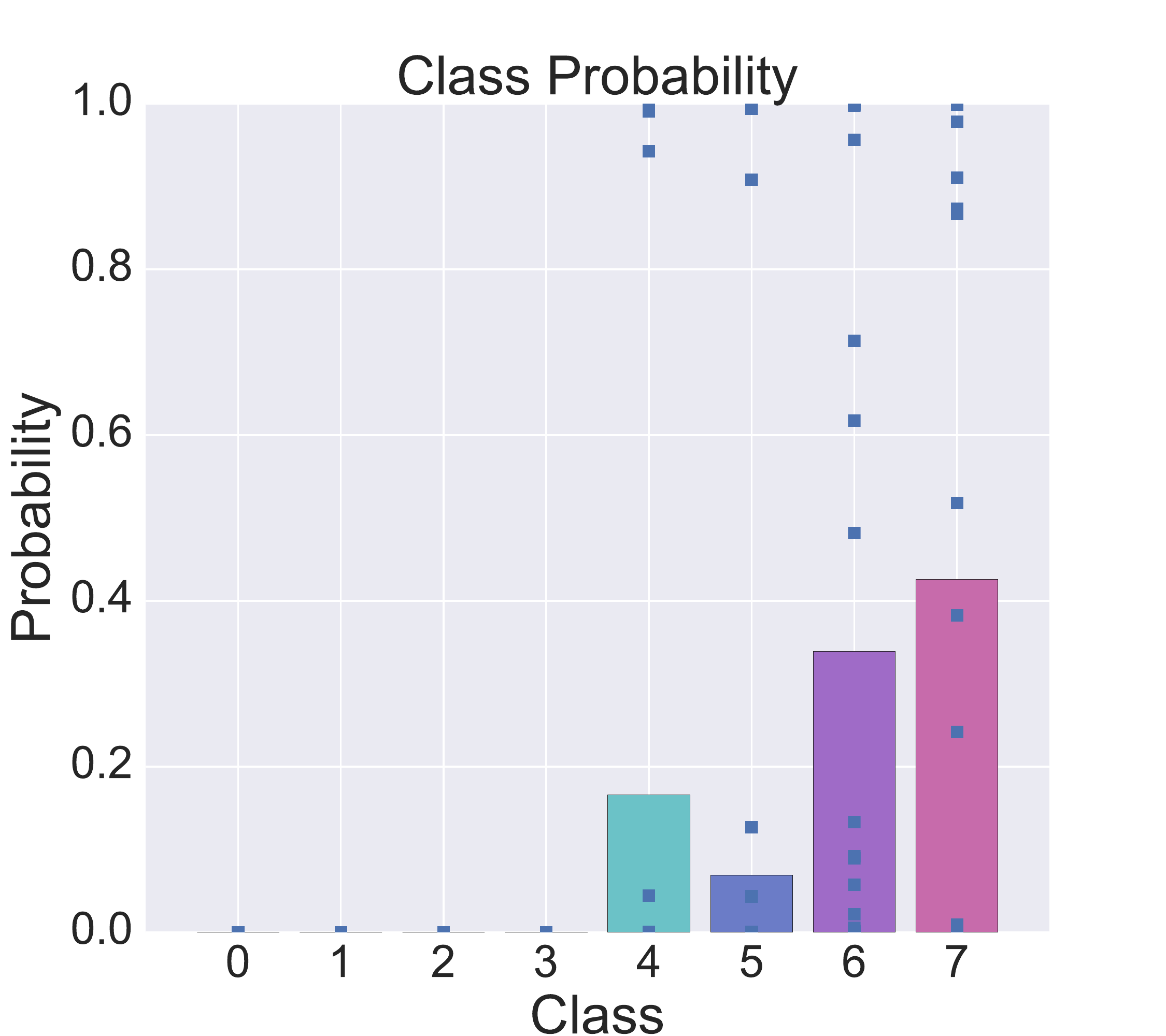}
        \includegraphics[width=0.15\textwidth]{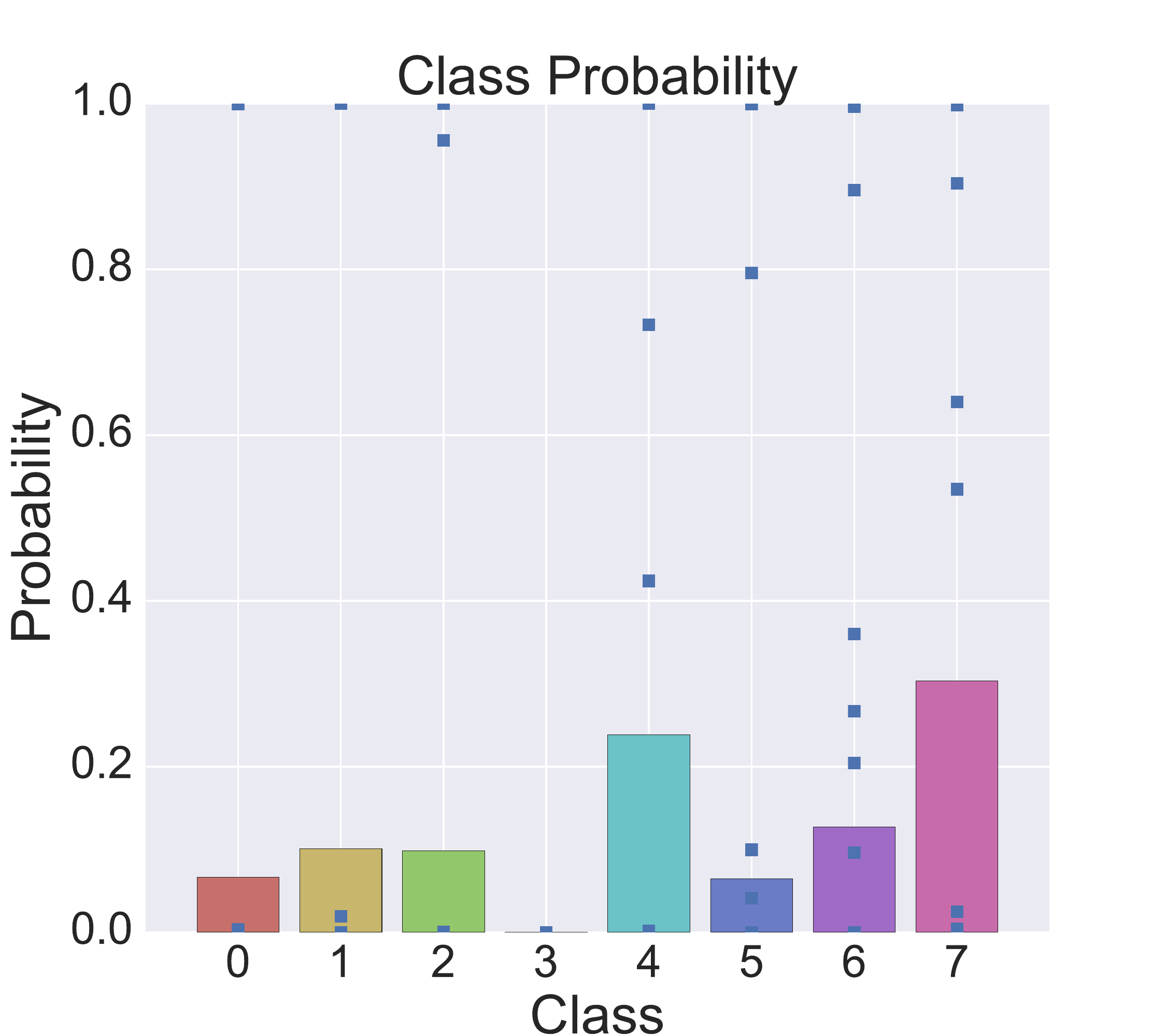}
    \end{minipage}

    \caption{Predictive distribution for randomly selected samples from the ADIENCE dataset.}
    \label{fig:comparison}
\end{figure*}

In the ADIENCE data set there are many confusing examples which are usually neighboring classes. One example for each class is randomly sampled and the predictive distributions for each model are evaluated. These results are shown in Figure \ref{fig:comparison}. In general, SGHMC and SGLD produce over-confident probabilities,assigning high values to incorrect labels. In contrast, D-SGHMC reduces that confidence, increasing uncertainty in the class predictions.

In particular, high confidence estimates for both SGHMC and SGLD on one example of class $4$ can be seen in Figure \ref{fig:comparison}. On the other hand, the proposed method allows to improve the uncertainty estimates between the neighboring classes, which significantly improves the classification results.

This situation is also replicated on one example of class $7$, where SGHMC and SGLD achieves over-confident misclassification (class $1$). The proposed method allows to alleviate the misclassification error and return uncertainty estimates for neighboring classes. The predictive distribution improves the classification results and provides the improved uncertainty estimates.


\clearpage 	
\section{Conclusion}

In many decision-making systems, the estimated uncertainty plays a crucial role in the final classification. Over-confident estimates can arise from noisy examples or when leading with classes that were not part of the training data set. The fundamental objective of achieving improved uncertainty estimates is to encourage decision-makers to question their decisions. In this way, compared to other state-of-the-art MCMC methods for large scale and high dimensional classification problems, the methodology presented in this paper improves the uncertainty estimated.

The proposed method is based on HMC and Dropout regularization. The experiments demonstrated that the method is capable of generating an approximation to the posterior distribution. In addition, the resulting predictive distributions also alleviate the misclassification error in difficult examples. However, it has not yet been proven that generated stochastic dynamics preserve the volume in its entirety. Future work will perform comparisons with other state-of-the-art variational methods. Moreover, the relationship between the proposed approach to approximate Bayesian model averaging can be also another line of research. 


%
\section*{Compliance with Ethical Standards}

\textbf{Funding} This work was supported by two CONICYT grants. Project CONICYT/FONDECYT Robust Multi-Target Tracking using Discrete Visual Features code 11140598. Project CONICYT/PFCHA  Beca Magister Nacional 2017 code 22170718. \textbf{Ethical approval} This article does not contain any studies with human participants or animals performed by any of the authors.

\bibliography{d_hmc}

\begin{thebibliography}{39}
\providecommand{\natexlab}[1]{#1}
\providecommand{\url}[1]{\texttt{#1}}
\expandafter\ifx\csname urlstyle\endcsname\relax
  \providecommand{\doi}[1]{doi: #1}\else
  \providecommand{\doi}{doi: \begingroup \urlstyle{rm}\Url}\fi

\bibitem[Afshar and Domke(2015)]{afshar2015reflection}
H.~M. Afshar and J.~Domke.
\newblock Reflection, refraction, and {Hamiltonian Monte Carlo}.
\newblock In \emph{Advances in Neural Information Processing Systems}, pages
  3007--3015, 2015.

\bibitem[Baldi and Sadowski(2014)]{baldi2014dropout}
P.~Baldi and P.~Sadowski.
\newblock The dropout learning algorithm.
\newblock \emph{Artificial intelligence}, 210:\penalty0 78--122, 2014.

\bibitem[Baldi and Sadowski(2013)]{baldi2013understanding}
P.~Baldi and P.~J. Sadowski.
\newblock Understanding dropout.
\newblock In \emph{Advances in neural information processing systems}, pages
  2814--2822, 2013.

\bibitem[Bardenet et~al.(2014)Bardenet, Doucet, and
  Holmes]{bardenet2014towards}
R.~Bardenet, A.~Doucet, and C.~Holmes.
\newblock Towards scaling up {Markov chain Monte Carlo}: an adaptive
  subsampling approach.
\newblock In \emph{International Conference on Machine Learning}, pages
  405--413, 2014.

\bibitem[Beskos et~al.(2013)Beskos, Pillai, Roberts, Sanz-Serna, Stuart,
  et~al.]{beskos2013optimal}
A.~Beskos, N.~Pillai, G.~Roberts, J.-M. Sanz-Serna, A.~Stuart, et~al.
\newblock Optimal tuning of the hybrid monte carlo algorithm.
\newblock \emph{Bernoulli}, 19\penalty0 (5A):\penalty0 1501--1534, 2013.

\bibitem[Bishop(2007)]{bishop2007pattern}
C.~M. Bishop.
\newblock Pattern recognition and machine learning (information science and
  statistics), 1st edn. 2006. corr. 2nd printing edn.
\newblock \emph{Springer, New York}, 2007.

\bibitem[Carpenter et~al.(2017)Carpenter, Gelman, Hoffman, Lee, Goodrich,
  Betancourt, Brubaker, Guo, Li, and Riddell]{carpenter2017stan}
B.~Carpenter, A.~Gelman, M.~D. Hoffman, D.~Lee, B.~Goodrich, M.~Betancourt,
  M.~Brubaker, J.~Guo, P.~Li, and A.~Riddell.
\newblock Stan: A probabilistic programming language.
\newblock \emph{Journal of Statistical Software}, 76\penalty0 (1), 2017.

\bibitem[Chaari et~al.(2016)Chaari, Tourneret, Chaux, and
  Batatia]{chaari2016hamiltonian}
L.~Chaari, J.-Y. Tourneret, C.~Chaux, and H.~Batatia.
\newblock A hamiltonian monte carlo method for non-smooth energy sampling.
\newblock \emph{IEEE Transactions on Signal Processing}, 64\penalty0
  (21):\penalty0 5585--5594, 2016.

\bibitem[Chen et~al.(2014)Chen, Fox, and Guestrin]{chen2014stochastic}
T.~Chen, E.~Fox, and C.~Guestrin.
\newblock Stochastic gradient hamiltonian monte carlo.
\newblock In \emph{International Conference on Machine Learning}, pages
  1683--1691, 2014.

\bibitem[Eidinger et~al.(2014)Eidinger, Enbar, and Hassner]{eidinger2014age}
E.~Eidinger, R.~Enbar, and T.~Hassner.
\newblock Age and gender estimation of unfiltered faces.
\newblock \emph{IEEE Transactions on Information Forensics and Security},
  9\penalty0 (12):\penalty0 2170--2179, 2014.

\bibitem[Gal and Ghahramani(2016)]{gal2016dropout}
Y.~Gal and Z.~Ghahramani.
\newblock Dropout as a bayesian approximation: Representing model uncertainty
  in deep learning.
\newblock In \emph{international conference on machine learning}, pages
  1050--1059, 2016.

\bibitem[Gelman et~al.(2014)Gelman, Carlin, Stern, Dunson, Vehtari, and
  Rubin]{gelman2014bayesian}
A.~Gelman, J.~B. Carlin, H.~S. Stern, D.~B. Dunson, A.~Vehtari, and D.~B.
  Rubin.
\newblock \emph{Bayesian data analysis}, volume~2.
\newblock CRC press Boca Raton, FL, 2014.

\bibitem[Girolami and Calderhead(2011)]{girolami2011riemann}
M.~Girolami and B.~Calderhead.
\newblock Riemann manifold langevin and hamiltonian monte carlo methods.
\newblock \emph{Journal of the Royal Statistical Society: Series B (Statistical
  Methodology)}, 73\penalty0 (2):\penalty0 123--214, 2011.

\bibitem[Griewank and Walther(2008)]{Griewank2008}
A.~Griewank and A.~Walther.
\newblock \emph{Evaluating Derivatives: Principles and Techniques of
  Algorithmic Differentiation}.
\newblock Society for Industrial and Applied Mathematics, Philadelphia, PA,
  USA, second edition, 2008.
\newblock ISBN 0898716594, 9780898716597.

\bibitem[Guyon et~al.(2005)Guyon, Gunn, Ben-Hur, and Dror]{guyon2005result}
I.~Guyon, S.~Gunn, A.~Ben-Hur, and G.~Dror.
\newblock Result analysis of the nips 2003 feature selection challenge.
\newblock In \emph{Advances in neural information processing systems}, pages
  545--552, 2005.

\bibitem[Hoffman(2017)]{hoffman2017learning}
M.~D. Hoffman.
\newblock Learning deep latent gaussian models with markov chain monte carlo.
\newblock In \emph{International Conference on Machine Learning}, pages
  1510--1519, 2017.

\bibitem[Hoffman and Gelman(2014)]{hoffman2014no}
M.~D. Hoffman and A.~Gelman.
\newblock The no-u-turn sampler: adaptively setting path lengths in hamiltonian
  monte carlo.
\newblock \emph{Journal of Machine Learning Research}, 15\penalty0
  (1):\penalty0 1593--1623, 2014.

\bibitem[Kingma et~al.(2015)Kingma, Salimans, and Welling]{Kingma2015}
D.~P. Kingma, T.~Salimans, and M.~Welling.
\newblock Variational dropout and the local reparameterization trick.
\newblock In C.~Cortes, N.~D. Lawrence, D.~D. Lee, M.~Sugiyama, and R.~Garnett,
  editors, \emph{Advances in Neural Information Processing Systems 28}, pages
  2575--2583. Curran Associates, Inc., 2015.

\bibitem[LeCun et~al.(1998)LeCun, Bottou, Bengio, and
  Haffner]{lecun1998gradient}
Y.~LeCun, L.~Bottou, Y.~Bengio, and P.~Haffner.
\newblock Gradient-based learning applied to document recognition.
\newblock \emph{Proceedings of the IEEE}, 86\penalty0 (11):\penalty0
  2278--2324, 1998.

\bibitem[Leibig et~al.(2017)Leibig, Allken, Ayhan, Berens, and
  Wahl]{leibig2017leveraging}
C.~Leibig, V.~Allken, M.~S. Ayhan, P.~Berens, and S.~Wahl.
\newblock Leveraging uncertainty information from deep neural networks for
  disease detection.
\newblock \emph{Scientific reports}, 7\penalty0 (1):\penalty0 17816, 2017.

\bibitem[Levi and Hassner(2015)]{levi2015age}
G.~Levi and T.~Hassner.
\newblock Age and gender classification using convolutional neural networks.
\newblock In \emph{Proceedings of the IEEE Conference on Computer Vision and
  Pattern Recognition Workshops}, pages 34--42, 2015.

\bibitem[Li and Hoiem(2018)]{Li2018}
Z.~Li and D.~Hoiem.
\newblock Learning without forgetting.
\newblock \emph{IEEE Transactions on Pattern Analysis and Machine
  Intelligence}, pages 1--1, 2018.
\newblock ISSN 0162-8828.
\newblock \doi{10.1109/TPAMI.2017.2773081}.

\bibitem[Miceli~Barone et~al.(2017)Miceli~Barone, Haddow, Germann, and
  Sennrich]{micelibarone2017}
A.~V. Miceli~Barone, B.~Haddow, U.~Germann, and R.~Sennrich.
\newblock Regularization techniques for fine-tuning in neural machine
  translation.
\newblock In \emph{Proceedings of the 2017 Conference on Empirical Methods in
  Natural Language Processing}, pages 1489--1494, Copenhagen, Denmark,
  September 2017. Association for Computational Linguistics.

\bibitem[Neal(2012)]{neal2012bayesian}
R.~M. Neal.
\newblock \emph{Bayesian learning for neural networks}, volume 118.
\newblock Springer Science \& Business Media, 2012.

\bibitem[Neal et~al.(2011)]{neal2011mcmc}
R.~M. Neal et~al.
\newblock {MCMC} using hamiltonian dynamics.
\newblock \emph{Handbook of Markov Chain Monte Carlo}, 2\penalty0 (11), 2011.

\bibitem[Nishimura et~al.(2017)Nishimura, Dunson, and
  Lu]{nishimura2017discontinuous}
A.~Nishimura, D.~Dunson, and J.~Lu.
\newblock Discontinuous hamiltonian monte carlo for models with discrete
  parameters and discontinuous likelihoods.
\newblock \emph{arXiv preprint arXiv:1705.08510}, 2017.

\bibitem[Pakman and Paninski(2013)]{pakman2013auxiliary}
A.~Pakman and L.~Paninski.
\newblock Auxiliary-variable exact hamiltonian monte carlo samplers for binary
  distributions.
\newblock In \emph{Advances in neural information processing systems}, pages
  2490--2498, 2013.

\bibitem[Parkhi et~al.(2015)Parkhi, Vedaldi, and Zisserman]{Parkhi15}
O.~M. Parkhi, A.~Vedaldi, and A.~Zisserman.
\newblock Deep face recognition.
\newblock In \emph{British Machine Vision Conference}, 2015.

\bibitem[Pearl(1988)]{Pearl:1988:PRI:52121}
J.~Pearl.
\newblock \emph{Probabilistic Reasoning in Intelligent Systems: Networks of
  Plausible Inference}.
\newblock Morgan Kaufmann Publishers Inc., San Francisco, CA, USA, 1988.
\newblock ISBN 0-934613-73-7.

\bibitem[Pereyra(2016)]{pereyra2016proximal}
M.~Pereyra.
\newblock Proximal markov chain monte carlo algorithms.
\newblock \emph{Statistics and Computing}, 26\penalty0 (4):\penalty0 745--760,
  2016.

\bibitem[Prince(2012)]{prince2012computer}
S.~J. Prince.
\newblock \emph{Computer vision: models, learning, and inference}.
\newblock Cambridge University Press, 2012.

\bibitem[Roberts and Rosenthal(1998)]{roberts1998optimal}
G.~O. Roberts and J.~S. Rosenthal.
\newblock Optimal scaling of discrete approximations to langevin diffusions.
\newblock \emph{Journal of the Royal Statistical Society: Series B (Statistical
  Methodology)}, 60\penalty0 (1):\penalty0 255--268, 1998.

\bibitem[Roberts and Stramer(2002)]{roberts2002langevin}
G.~O. Roberts and O.~Stramer.
\newblock Langevin diffusions and metropolis-hastings algorithms.
\newblock \emph{Methodology and computing in applied probability}, 4\penalty0
  (4):\penalty0 337--357, 2002.

\bibitem[Srivastava et~al.(2014)Srivastava, Hinton, Krizhevsky, Sutskever, and
  Salakhutdinov]{srivastava2014dropout}
N.~Srivastava, G.~Hinton, A.~Krizhevsky, I.~Sutskever, and R.~Salakhutdinov.
\newblock Dropout: A simple way to prevent neural networks from overfitting.
\newblock \emph{The Journal of Machine Learning Research}, 15\penalty0
  (1):\penalty0 1929--1958, 2014.

\bibitem[Tran et~al.(2017)Tran, Hoffman, Saurous, Brevdo, Murphy, and
  Blei]{tran2017deep}
D.~Tran, M.~D. Hoffman, R.~A. Saurous, E.~Brevdo, K.~Murphy, and D.~M. Blei.
\newblock Deep probabilistic programming.
\newblock In \emph{International Conference on Learning Representations}, 2017.

\bibitem[Wan et~al.(2013)Wan, Zeiler, Zhang, Cun, and
  Fergus]{wan2013regularization}
L.~Wan, M.~Zeiler, S.~Zhang, Y.~L. Cun, and R.~Fergus.
\newblock Regularization of neural networks using dropconnect.
\newblock In \emph{Proceedings of the 30th international conference on machine
  learning (ICML-13)}, pages 1058--1066, 2013.

\bibitem[Wang et~al.(2013)Wang, Mohamed, and Freitas]{wang2013adaptive}
Z.~Wang, S.~Mohamed, and N.~Freitas.
\newblock Adaptive hamiltonian and riemann manifold monte carlo.
\newblock In \emph{International Conference on Machine Learning}, pages
  1462--1470, 2013.

\bibitem[Warde-Farley et~al.(2013)Warde-Farley, Goodfellow, Courville, and
  Bengio]{warde2013empirical}
D.~Warde-Farley, I.~J. Goodfellow, A.~Courville, and Y.~Bengio.
\newblock An empirical analysis of dropout in piecewise linear networks.
\newblock \emph{arXiv preprint arXiv:1312.6197}, 2013.

\bibitem[Welling and Teh(2011)]{welling2011bayesian}
M.~Welling and Y.~W. Teh.
\newblock Bayesian learning via stochastic gradient langevin dynamics.
\newblock In \emph{Proceedings of the 28th International Conference on Machine
  Learning (ICML-11)}, pages 681--688, 2011.

\end{thebibliography}

\end{document}